\documentclass[11pt,a4paper]{article}
\usepackage{rawrag_arxiv}

\usepackage{amsmath,amsfonts,bm}

\def\eqref#1{equation~\ref{#1}}

\def\1{\bm{1}}

\DeclareMathAlphabet{\mathsfit}{\encodingdefault}{\sfdefault}{m}{sl}
\SetMathAlphabet{\mathsfit}{bold}{\encodingdefault}{\sfdefault}{bx}{n}

\usepackage{booktabs}
\usepackage{hhline}
\usepackage{longtable}
\usepackage{multirow}
\usepackage{placeins}
\usepackage{colortbl}
\usepackage{graphicx}
\usepackage{capt-of}
\usepackage{microtype}
\usepackage{hyperref}
\hypersetup{
  colorlinks=true,
  allcolors=arxivlink,
  pdftitle={Is Complex Agentic Memory Always Needed? Naive RAG Is Stronger Than We Thought},
  pdfauthor={Zhengye Bao, Shengtao Zhang, Haokai Zhang, Shiqi Lou, Muning Wen, Minnan Luo}
}
\usepackage{url}
\usepackage[most]{tcolorbox}
\renewcommand{\bottomfraction}{0.85}
\renewcommand{\floatpagefraction}{0.85}
\titlespacing*{\section}{0pt}{12pt plus 2pt minus 2pt}{4pt}
\titlespacing*{\subsection}{0pt}{9pt plus 2pt minus 2pt}{2pt}
\titlespacing*{\subsubsection}{0pt}{7pt plus 2pt minus 1pt}{2pt}
\titlespacing*{\paragraph}{0pt}{5pt plus 1pt minus 1pt}{0.7em}
\renewcommand{\bibfont}{\small}

\tcbset{
  findingboxstyle/.style={
    width=\linewidth,
    top=7pt,
    bottom=2pt,
    left=2pt,
    right=2pt,
    colback=blue!6!white,
    colframe=black,
    colbacktitle=black,
    coltitle=white,
    fonttitle=\bfseries,
    enhanced,
    attach boxed title to top left={yshift=-0.1in,xshift=0.1in},
    boxed title style={boxrule=0pt,colframe=white},
  }
}
\newtcolorbox{findingbox}[1]{findingboxstyle,title=#1}
\newtcblisting{promptbox}[1]{
  enhanced,breakable,listing only,
  title={#1},fonttitle=\small\bfseries,
  colback=gray!3,colframe=gray!55,colbacktitle=gray!12,coltitle=black,
  boxrule=0.4pt,arc=1pt,left=5pt,right=5pt,top=3pt,bottom=3pt,
  before skip=4pt,after skip=4pt,
  listing options={
    basicstyle=\footnotesize\rmfamily,
    columns=fullflexible,keepspaces=true,showstringspaces=false,
    breaklines=true,breakatwhitespace=true,
    literate={→}{{$\rightarrow$}}1
  }
}

\title{Is Complex Agentic Memory Always Needed? \\Naive RAG Is Stronger Than We Thought}
\author{%
  Zhengye Bao\textsuperscript{1}\enspace
  Shengtao Zhang\textsuperscript{2,1,*}\enspace
  Haokai Zhang\textsuperscript{3,1,*}\enspace
  Shiqi Lou\textsuperscript{4}\enspace
  Muning Wen\textsuperscript{2,\dag}\enspace
  Minnan Luo\textsuperscript{1,\dag}%
}
\date{}
\fancypagestyle{arxivfirstpage}{%
  \fancyhead[L]{%
    \hspace*{12pt}%
    \raisebox{4pt}{\includegraphics[height=22pt]{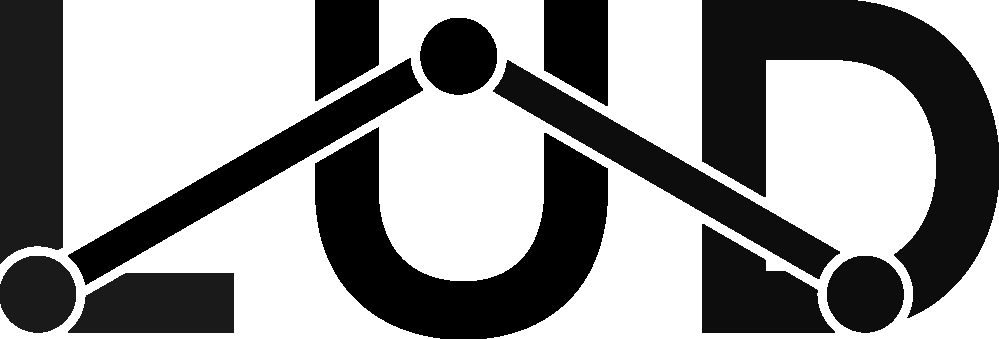}}%
  }%
  \fancyhead[R]{%
    \includegraphics[height=30pt]{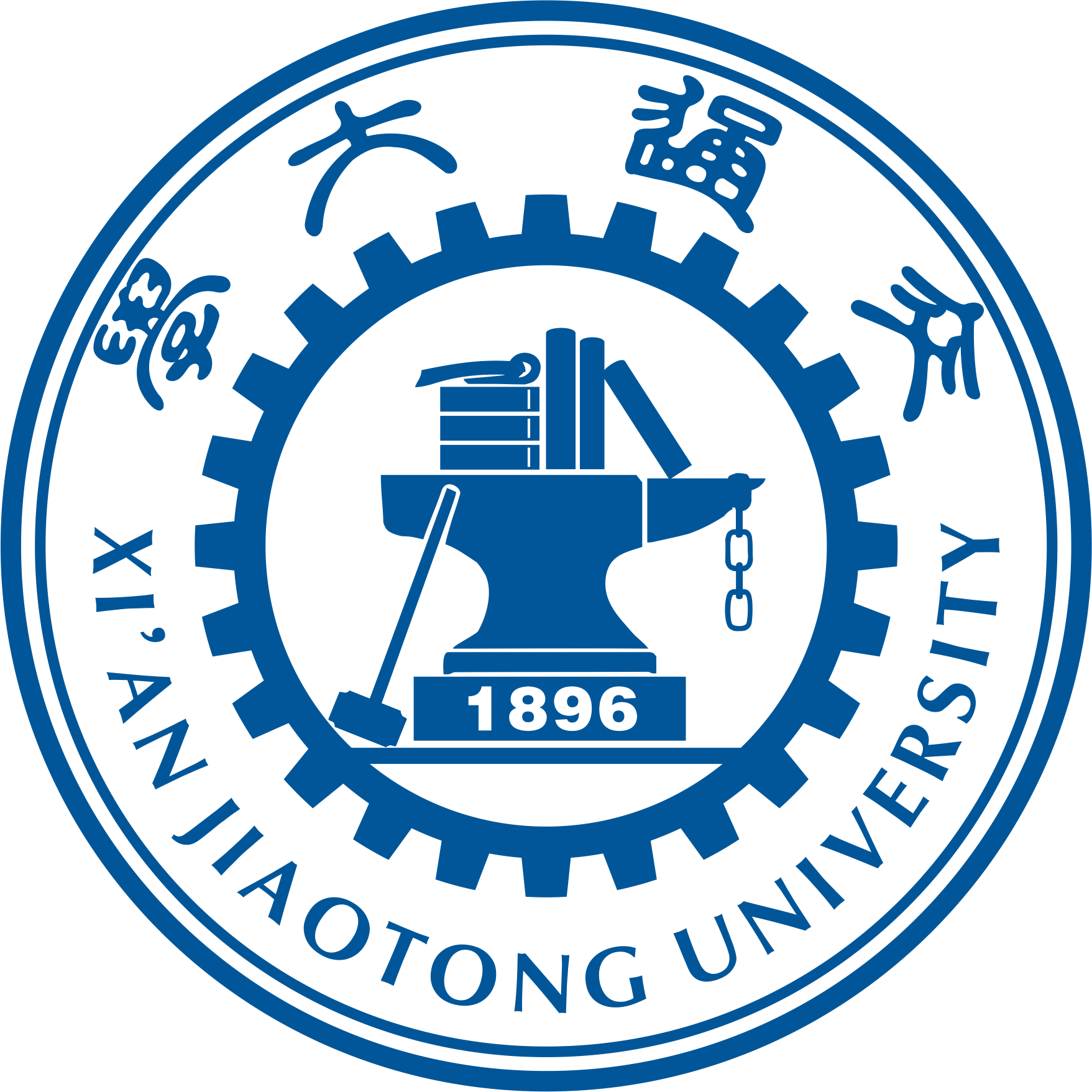}\hspace{10pt}%
    \includegraphics[height=30pt]{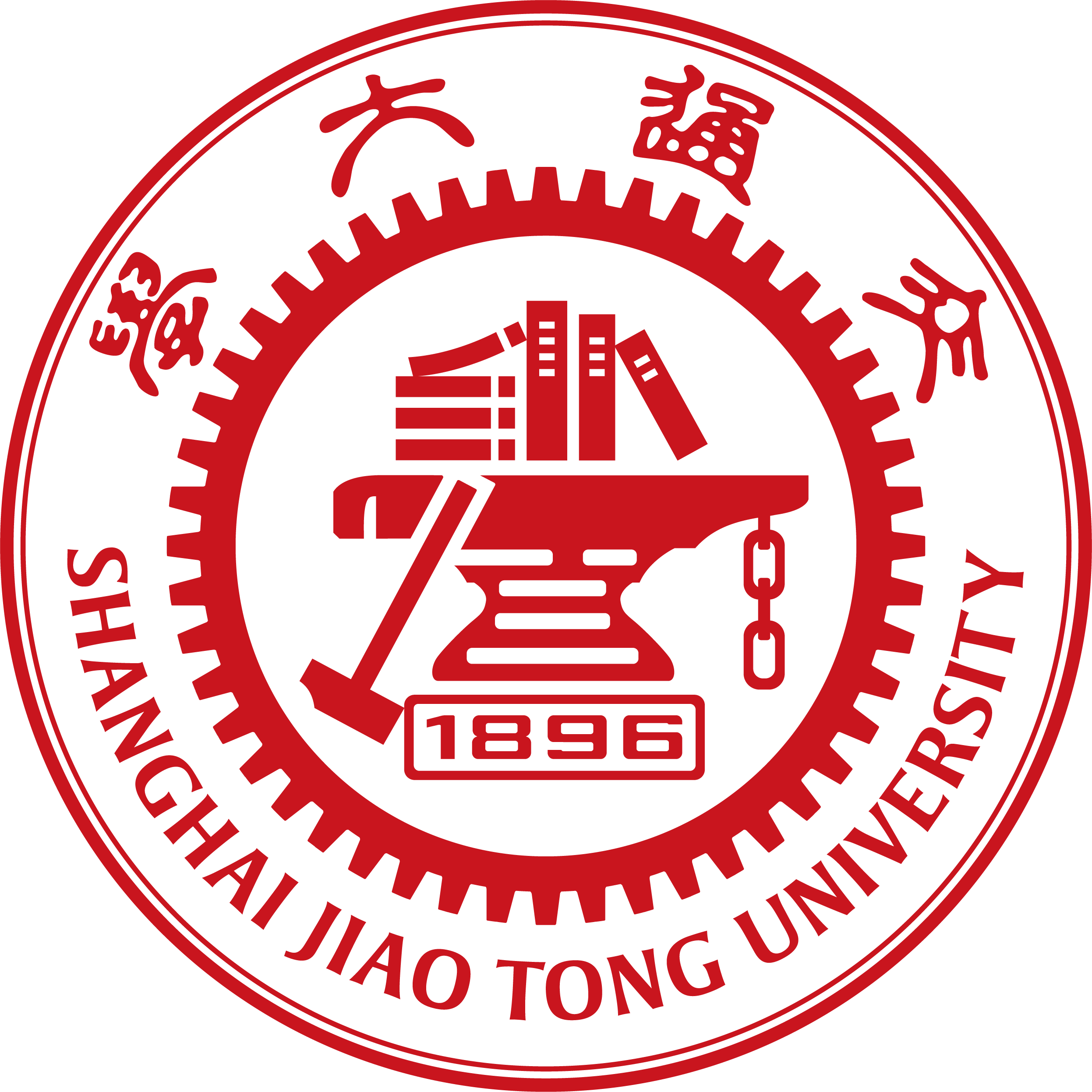}\hspace{10pt}%
    \includegraphics[height=30pt]{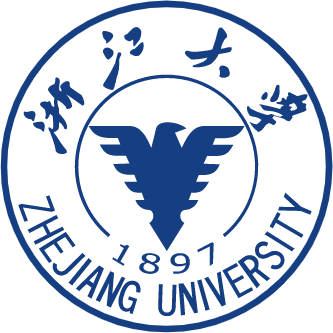}\hspace{10pt}%
    \includegraphics[height=30pt]{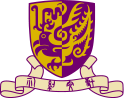}%
    \hspace*{12pt}%
  }%
  \renewcommand{\headrulewidth}{1pt}%
  \renewcommand{\footrulewidth}{0pt}%
}
\makeatletter
\renewcommand{\maketitle}{%
  \thispagestyle{arxivfirstpage}%
  \par\begingroup
    \centering
    \setlength{\parskip}{0pt}%
    \vspace*{-24pt}%
    {\normalfont\bfseries\fontsize{16}{20}\selectfont\@title\par}%
    \vskip8pt
    {\normalfont\fontsize{10}{12}\selectfont\@author\par}%
    \vskip4pt
    {\normalfont\fontsize{10}{12}\selectfont\url{https://github.com/Litmeb/RAW-RAG}\par}%
    \vskip4pt
  \endgroup
  \@thanks
  \begingroup
    \renewcommand{\thefootnote}{}%
    \renewcommand{\@makefntext}[1]{\noindent##1}%
    \footnotetext[0]{%
      \textsuperscript{1}Luo Lab Undergraduate Division (LUD), Xi'an Jiaotong University;
      \textsuperscript{2}Shanghai Jiao Tong University;
      \textsuperscript{3}Zhejiang University;
      \textsuperscript{4}The Chinese University of Hong Kong.
      \textsuperscript{*}Equal contribution.
      \textsuperscript{\dag}Corresponding authors:
      Minnan Luo\textless\href{mailto:minnluo@xjtu.edu.cn}{minnluo@xjtu.edu.cn}\textgreater;
      Muning Wen\textless\href{mailto:muningwen@sjtu.edu.cn}{muningwen@sjtu.edu.cn}\textgreater.
    }%
  \endgroup
}
\makeatother
\newcommand{\rawrag}{\textsc{Raw-RAG}}
\definecolor{rawraghighlight}{HTML}{FFE4CC}
\newcommand{\best}[1]{\textbf{#1}}
\newcommand{\second}[1]{\underline{#1}}
\newcommand{\modelname}[1]{\raisebox{-0.18ex}[\height][\depth]{\textit{#1}}}
\newcommand{\methodrule}{\specialrule{0.4pt}{\aboverulesep}{\belowrulesep}}
\newcommand{\diagnostictablestyle}{%
  \setlength{\heavyrulewidth}{0.9pt}%
  \setlength{\lightrulewidth}{0.4pt}%
  \setlength{\aboverulesep}{2pt}%
  \setlength{\belowrulesep}{2pt}%
}
\newcommand{\paircell}[3]{%
  \renewcommand{\arraystretch}{0.9}%
  \begin{tabular}[c]{@{}c@{}}
    #1\\[-1pt]{\scriptsize #2}\\[-1pt]{\scriptsize $#3$}
  \end{tabular}%
}
\begin{document}
\maketitle
\suppressfloats[t]

\begin{abstract}
LLM memory is widely used in long-term conversation understanding. Agentic memory systems often use mutation, structural organization, learned retrieval strategies, or other complex operations. Is this complexity always necessary? We study \rawrag{}, a naive RAG baseline using dense retrieval and reranking over raw history to supply evidence to a reader LLM. Across LoCoMo, LongMemEval, MEME, and MemoryAgentBench, \rawrag{} ranks first among the compared memory systems in six of seven benchmark--model settings, outperforming the strongest compared agentic memory system by 0.13--5.53 points. We investigate why raw-history retrieval remains competitive. Evidence analysis shows that its higher coverage of necessary evidence can offset lower utilization. Useful evidence includes both answer values and the relations and dependency paths needed to derive them. Even complete access to transformed memory stores leaves an accuracy gap relative to raw evidence, highlighting the importance of preserving raw information. These findings challenge the necessity of complex agentic memory for the evaluated conversation tasks and establish naive RAG as an essential baseline.
\end{abstract}

\begingroup
\setlength{\skip\footins}{6pt}
\par\vfill
\noindent\begin{minipage}{\linewidth}
  \centering
  \captionsetup{type=figure,belowskip=0pt}
  \includegraphics[width=0.95\linewidth,height=0.43\textheight,keepaspectratio]{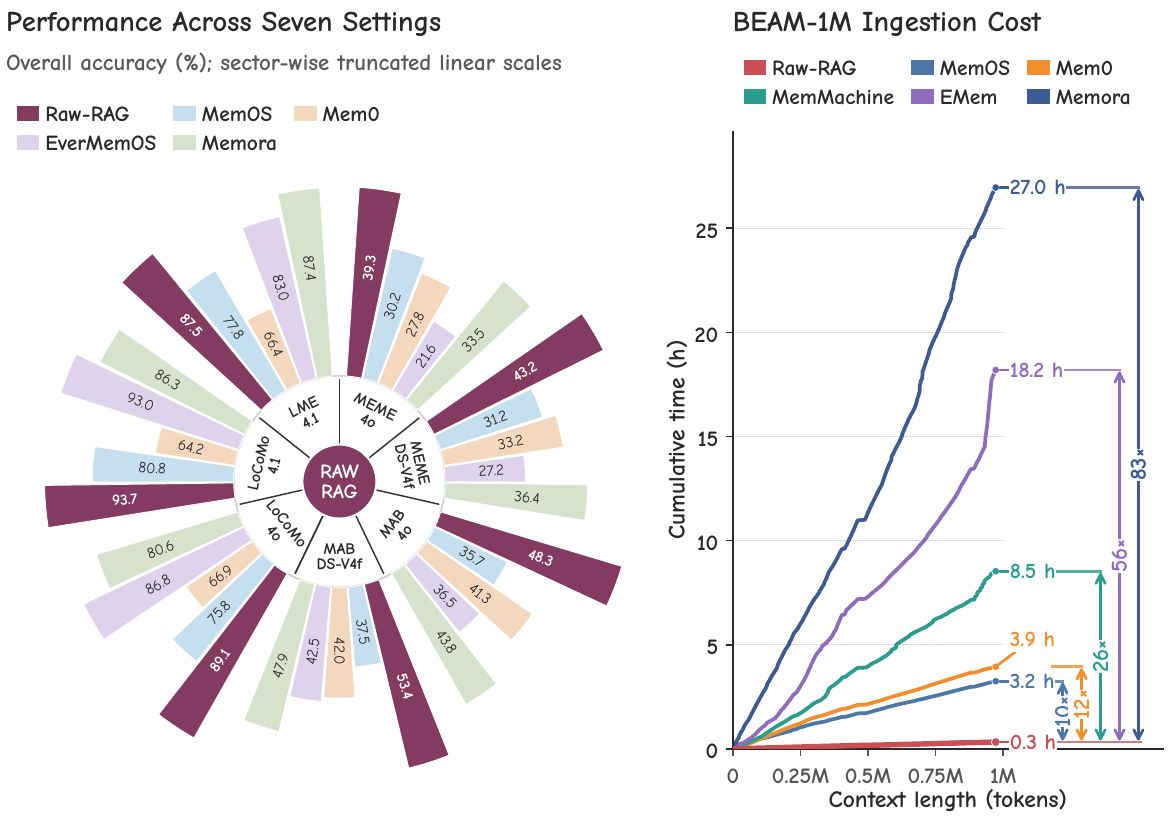}
  \caption{Strong accuracy with low ingestion cost.
  Left: accuracy across seven benchmark--model settings, with sector-wise
  truncated linear scales. Right: cumulative sequential ingestion time on
  BEAM-1M.}
  \label{fig:teaser-results}
\end{minipage}\par
\newpage
\endgroup
\begin{figure*}[!t]
  \centering
  \includegraphics[width=\textwidth]{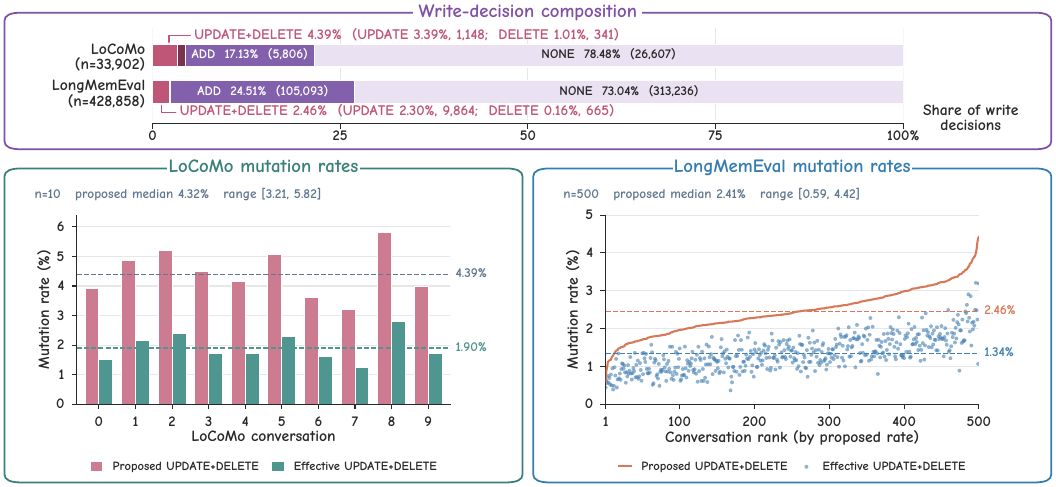}
  \caption{Mem0 proposes few mutations, and fewer take effect.
  \textbf{Top}: write-action distributions. \textbf{Bottom}: per-conversation proposed and
  effective mutation rates on LoCoMo (left) and LongMemEval (right).
  Dashed lines: decision-weighted means. Rates are relative to all emitted
  decisions.}
  \label{fig:mem0-actions}
\end{figure*}

\section{Introduction}

Large language models (LLMs) are increasingly used in long-term interactions,
where they must draw on information accumulated across many conversations
and tasks \citep{park2023generative,zhong2024memorybank}.
The resulting histories can far exceed a model's context window, yet relevant
past information must remain accessible for future responses.
To address this challenge, recent research has developed a range of agentic
memory systems that store, organize, and retrieve information beyond the
active context \citep{packer2023memgpt,chhikara2025mem0,xu2025amem,hu2026evermemos}.
Memory management includes extraction, summarization, mutation, consolidation,
clustering, graph construction, and community discovery, while retrieval may
involve iterative search or trainable retrieval strategies
\citep{chhikara2025mem0,zhong2024memorybank,liu2026simplemem,xu2025amem,hu2026evermemos,edge2024graphrag,xia2026memora}.
These developments raise a basic question: \textbf{when is the added complexity of
agentic memory systems}\footnote{
Here, we use \emph{agentic memory system} as an umbrella term for memory systems encompassing agentic content processing, structural organization, and retrieval control beyond a fixed raw-history retrieval pipeline.}
\textbf{necessary?} To investigate, we start with a simple
alternative: retrieval-augmented generation (RAG) over raw history.

RAG is often considered insufficient for long-term memory
\citep{zhang2024memorysurvey,sumers2023coala,packer2023memgpt}. Two common objections to raw-history retrieval motivate our investigation.
The first is conceptual: RAG is commonly
associated with static corpora, whereas long-term interactions involve
superseded facts, deletions, and dependent state changes. Can retrieving history
suffice without explicitly maintaining current state? The second is empirical:
prior evaluations report RAG trailing agentic memory systems
\citep{chhikara2025mem0,dai2026recmem,xia2026memora}.

These concerns motivate a closer examination of memory management.
We first ask how often memory tasks involve dynamic changes, how
complex those changes are, and whether they exceed what reader LLMs can
handle directly from retrieved evidence.
Two preliminary experiments examine explicit state maintenance.
We first audit Mem0 on all
LoCoMo conversations and all 500 LongMemEval examples. UPDATE and DELETE
together account for only 4.39\% of the 33,902 write decisions on LoCoMo
and 2.46\% of the 428,858 write decisions on LongMemEval
(Figure~\ref{fig:mem0-actions}); many proposed updates are duplicates
or malformed. Effective mutations, defined as persisted text-changing updates
or deletions, account for only 1.90\% and 1.34\% of all decisions.
Replacing UPDATE with append-only ADD and DELETE with NONE raises
LoCoMo accuracy from 67.36\% to 69.42\%, without reducing any category score
(Table~\ref{tab:no-update}). Explicit mutation provides no accuracy gain in this
configuration.
We then test whether raw evidence is sufficient for dynamic conversation on MEME,
which includes exact recall,
aggregation, tracking, deletion, cascade, and invalidation tasks. Readers receive
the evidence available at the queried phase in chronological order, eliminating
retrieval errors but still requiring them to resolve state changes. DeepSeek-V4-Flash reaches
82.32\% on cascades and 91.54\% on invalidation (Table~\ref{tab:meme-oracle}).
Performance varies substantially across readers, but these results show
that difficult transitions can be resolved from sufficient raw evidence.
Together, these diagnostics suggest that capable readers can often resolve
changing states directly from raw evidence.
To examine the empirical concern, we review 14 prior works and identify
coarse retrieval units, restricted
retrieval depth, or absent reranking in several evaluated baselines
(Appendix~\ref{app:prior-rag-audit}). These limitations leave open whether the
reported gaps reflect a need for agentic memory systems or weaknesses in the
chosen retrieval baselines.

\begin{table*}[!t]
  \begin{minipage}[b]{0.48\textwidth}
    \caption{LoCoMo accuracy for Mem0~\citep{chhikara2025mem0} under the full mutation policy and
    an ADD/NONE-only policy. Category 5 adversarial questions are excluded.
    $\Delta$: ADD/NONE minus full.}
    \label{tab:no-update}
    \vspace{0.5em}
    \centering\scriptsize
    \setlength{\tabcolsep}{2.5pt}
    \diagnostictablestyle
    \begin{tabular}[b]{l|cc>{\columncolor{gray!15}}c}
      \toprule
      Category & Full policy & ADD/NONE only & $\Delta$ \\
      \midrule
      Multi-hop & \second{63.24} & \best{64.66} & +1.42 \\
      Temporal & \second{49.74} & \best{53.06} & +3.32 \\
      Open-domain & \second{53.47} & \best{60.42} & +6.94 \\
      Single-hop & \second{77.05} & \best{78.28} & +1.23 \\
      \midrule
      Overall & \second{67.36} & \best{69.42} & +2.06 \\
      \bottomrule
    \end{tabular}
  \end{minipage}\hfill
  \begin{minipage}[b]{0.48\textwidth}
    \caption{MEME phase-visible oracle accuracy (\%). Overall is the six-task
    macro-average.}
    \label{tab:meme-oracle}
    \vspace{0.5em}
    \centering\scriptsize
    \setlength{\tabcolsep}{1.35pt}
    \diagnostictablestyle
    \begin{tabular}[b]{l|ccc}
      \toprule
      Task & GPT-4o-mini & DeepSeek-V4-Flash & Qwen3.5-35B-A3B \\
      \midrule
      ER & 98.00 & 99.00 & 97.00 \\
      Agg & 47.00 & 63.00 & 56.00 \\
      Tr & 81.00 & 92.00 & 89.00 \\
      Del & 39.00 & 86.00 & 46.00 \\
      Cas & 1.83 & 82.32 & 39.63 \\
      Abs & 36.15 & 91.54 & 93.85 \\
      \midrule
      \textbf{Overall} & \textbf{50.50} & \textbf{85.48} & \textbf{70.25} \\
      \bottomrule
    \end{tabular}%
  \end{minipage}
\end{table*}

We introduce \rawrag{} as a diagnostic baseline: naive RAG over raw history,
using only dense retrieval and cross-encoder reranking to select evidence
for a reader \citep{lewis2020rag,nogueira2019bert}.
Across four benchmarks, this simple baseline
ranks first among the compared memory systems in six of seven
benchmark--model settings.

Our contributions are summarized as follows:
\begin{itemize}
  \setlength{\itemsep}{0pt}
  \item We challenge the presumed necessity of explicit memory mutation:
  capable readers can resolve dynamic states from sufficient raw history,
  and disabling mutation improves accuracy in the tested configuration.
  \item We establish \rawrag{} as a strong reference baseline, ranking first
  in six of seven settings across four benchmarks, showing that naive RAG
  matches or exceeds the compared agentic memory systems in most settings.
  \item We identify evidence access and relational completeness as key
  bottlenecks: retrieving answer values does not ensure sufficient supporting
  paths, and complete-store access does not close the gap to raw evidence.
\end{itemize}

\FloatBarrier
\section{Related Work}

\subsection{Agentic memory systems}
External-memory architectures organize persistent information through
write, management, and read operations
\citep{sumers2023coala,zhang2024memorysurvey,wang2024llmagentsurvey}.
Early approaches summarize or reflect on experience streams
\citep{park2023generative,shinn2023reflexion,wang2023voyager,
zhong2024memorybank,wang2024recursively,liu2023thinkinmemory},
or provide tiered stores and read--write interfaces
\citep{modarressi2023retllm,hu2023chatdb,packer2023memgpt,
liang2023scm}.
Recent approaches extract and mutate facts or maintain evolving graphs
\citep{chhikara2025mem0,xu2025amem,rasmussen2025zep,
nan2025nemori},
consolidate hierarchical or recurrent stores
\citep{kang2025memoryos,li2025memos,hu2026evermemos,nevamind2025memu,
dai2026recmem,ma2025nemori,wuzhu2026agentzeromemory},
or construct selective event-level and abstract representations
\citep{zhou2025simple,liu2026simplemem,ma2025nemori,xia2026memora,
fang2025lightmem,hu2026xmemory,wang2026cuemem}.
Hybrid and lightweight approaches retain source episodes alongside
derived memories or selectively maintain events and preferences
\citep{wang2026memmachine,liao2026leanmem,sun2025pamu,
pan2025secom,kim2025premem,wu2025sgmem,xux2026com}.
Concurrent work studies ReFind, which preserves raw chat logs and gives an
agent iterative lexical search with session, temporal, context-expansion, and
deduplication controls \citep{li2026refind}. We evaluate a
fixed dense-retrieval and reranking baseline, isolating evidence access
from query-time search control.
We compare these systems with naive RAG for long-term conversation
understanding and diagnose their evidence-access limitations.

\begin{figure*}[!t]
  \centering
  \includegraphics[width=\textwidth,pagebox=cropbox,clip]{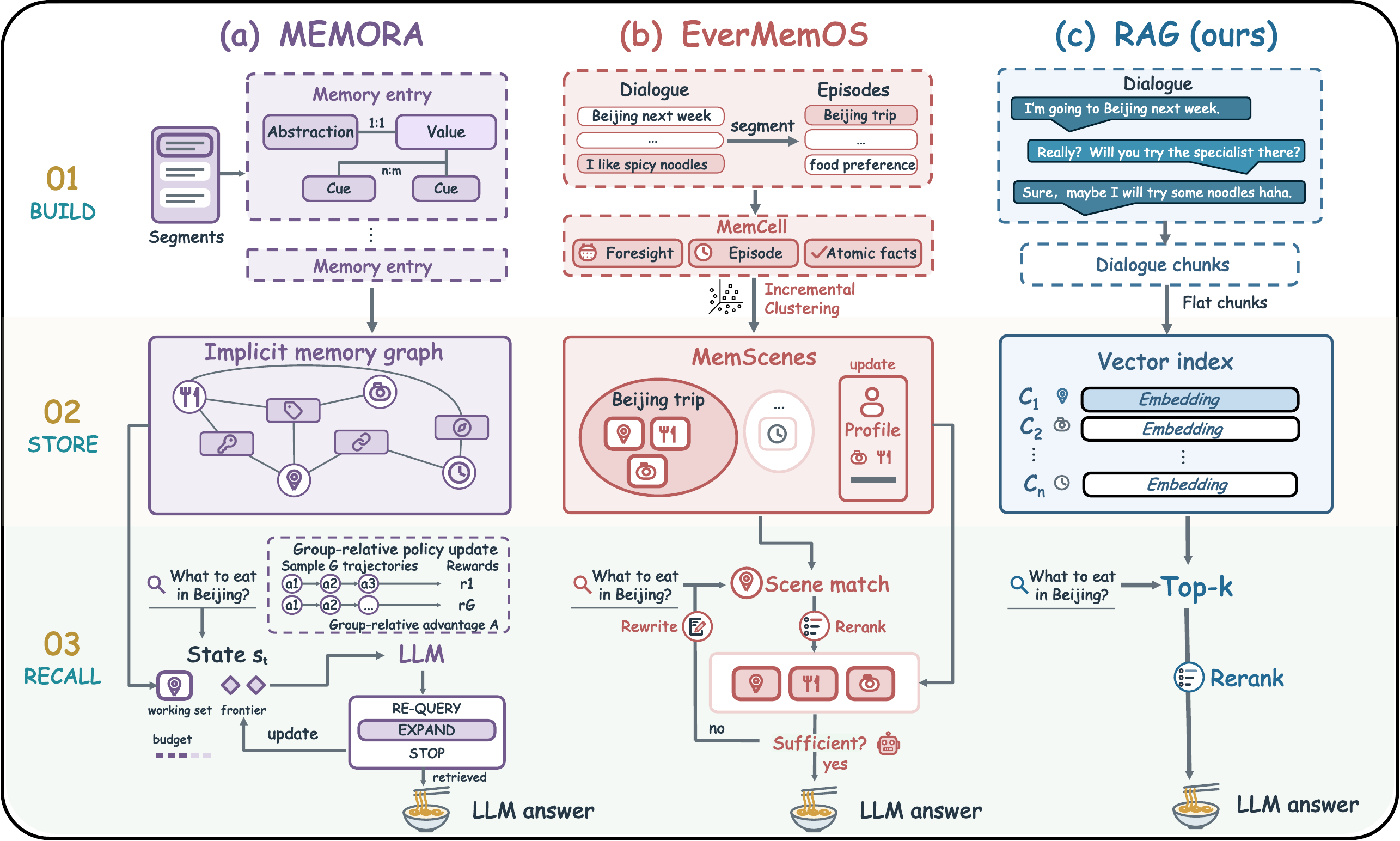}
  \caption{Pipelines of agentic memory systems (EverMemOS and Memora) and
  naive RAG (\rawrag{}) across build, store, and recall stages.}
  \label{fig:raw-rag-pipeline}
\end{figure*}

\subsection{Retrieval-augmented generation}
Retrieval-augmented language models combine external non-parametric memory
with a generator
\citep{lewis2020rag,izacard2021fid,borgeaud2022retro}.
Lexical and dense retrievers select candidates; late-interaction models and
cross-encoders refine ranking
\citep{robertson2009probabilistic,reimers2019sentencebert,khattab2020colbert,nogueira2019bert}.
Structure-aware variants use summaries, graphs, or hierarchies
\citep{sarthi2024raptor,edge2024graphrag,gutierrez2024hipporag,chen2023memwalker,
hu2026xmemory}.
Adaptive variants learn when to retrieve and assess evidence or answers
\citep{asai2024selfrag}.
Long-context and memory-QA studies emphasize evidence selection, granularity,
index keys, and temporal cues
\citep{liu2024lostinthemiddle,maharana2024locomo,wu2025longmemeval,
jiang2025personamemv2,jung2026meme,hu2026memoryagentbench,
tavakoli2026beam}.
We use standard dense retrieval and reranking to assess what agentic
memory systems add to long-term conversation understanding.

\begingroup
\setlength{\textfloatsep}{10pt plus 2pt minus 2pt}

\begin{table}[!t]
  \caption{LoCoMo and LongMemEval accuracy by question type (\%).
  For LongMemEval,
  S-U., S-A., and S-P. denote single-session user, assistant, and preference;
  M-S. denotes multi-session; KU. knowledge update; and TR. temporal
  reasoning. Bold: column maxima excluding Full-context.}
  \label{tab:main}
  \centering\small
  \setlength{\tabcolsep}{2.5pt}
  \renewcommand{\arraystretch}{1.05}
  \resizebox{\textwidth}{!}{%
  \begin{tabular}{l*{4}{c}>{\columncolor{yellow!15}}c
                       *{4}{c}>{\columncolor{yellow!15}}c
                       *{6}{c}>{\columncolor{yellow!15}}c}
    \toprule
    \multirow{3}{*}[-\dimexpr\aboverulesep+\belowrulesep+\cmidrulewidth\relax]{Method}
    & \multicolumn{10}{c}{LoCoMo} & \multicolumn{7}{c}{LongMemEval} \\
    \cmidrule(lr){2-11}\cmidrule(lr){12-18}
    & \multicolumn{5}{>{\columncolor{gray!15}}c}{\modelname{GPT-4o-mini}}
    & \multicolumn{5}{>{\columncolor{gray!15}}c}{\modelname{GPT-4.1-mini}}
    & \multicolumn{7}{>{\columncolor{gray!15}}c}{\modelname{GPT-4.1-mini}} \\
    \cmidrule(lr){2-6}\cmidrule(lr){7-11}\cmidrule(lr){12-18}
    & Single & Multi & Temp. & Open & Overall
    & Single & Multi & Temp. & Open & Overall
    & S-U. & S-A. & S-P. & M-S. & KU. & TR. & Overall \\
    \midrule
    \textit{Full-context}
    & 83.00 & 66.80 & 56.20 & 48.60 & 72.30
    & 86.90 & 77.20 & 74.20 & 56.60 & 80.60
    & 85.70 & 98.20 & 16.70 & 51.10 & 76.90 & 60.20 & 65.60 \\
    \specialrule{\lightrulewidth}{\aboverulesep}{\belowrulesep}
    MemU~\citep{nevamind2025memu}
    & 72.77 & 62.41 & 33.96 & 46.88 & 61.15
    & 74.91 & 72.34 & 43.61 & 54.17 & 66.67
    & 67.14 & 19.64 & 76.67 & 42.10 & 41.02 & 17.29 & 38.40 \\
    Mem0~\citep{chhikara2025mem0}
    & 67.13 & 51.15 & 55.51 & 72.93 & 66.88
    & 68.97 & 61.70 & 58.26 & 50.00 & 64.20
    & 82.86 & 26.78 & 90.00 & 63.15 & 66.67 & 72.18 & 66.40 \\
    A-MEM~\citep{xu2025amem}
    & 66.83 & 52.13 & 61.99 & 30.21 & 60.84
    & 73.25 & 59.93 & 72.90 & 42.71 & 68.83
    & 95.71 & \best{100.00} & 63.33 & 61.65 & 82.05 & 52.63 & 71.60 \\
    MemoryOS~\citep{kang2025memoryos}
    & 74.55 & 58.87 & 44.24 & 46.88 & 63.64
    & 77.05 & 66.31 & 47.66 & 55.21 & 67.60
    & 94.29 & 89.29 & \best{100.00} & 67.67 & 80.77 & 54.89 & 74.40 \\
    Nemori~\citep{nan2025nemori}
    & 82.10 & 65.30 & 71.00 & 44.80 & 74.40
    & 84.90 & 75.10 & 77.60 & 51.00 & 79.40
    & 90.00 & 92.90 & 86.70 & 55.60 & 79.50 & 72.20 & 74.60 \\
    RecMem~\citep{dai2026recmem}
    & 79.79 & 72.70 & 57.32 & 58.33 & 72.47
    & 86.92 & 78.37 & 75.39 & 57.29 & 81.10
    & 95.71 & 96.43 & 86.67 & 65.41 & 75.64 & 68.42 & 76.80 \\
    MemOS~\citep{li2025memos}
    & 81.09 & 67.49 & 75.18 & 55.90 & 75.80
    & 85.37 & 79.43 & 75.08 & 64.58 & 80.76
    & 95.71 & 67.86 & \second{96.67} & 70.67 & 74.26 & 77.44 & 77.80 \\
    SimpleMem~\citep{liu2026simplemem}
    & 86.01 & 74.94 & 63.76 & 68.40 & 78.25
    & 88.15 & 79.55 & 78.82 & \second{76.04} & 83.87
    & 85.71 & 75.00 & 76.67 & 60.92 & 79.48 & \second{83.46} & 76.87 \\
    EMem~\citep{zhou2025simple}
    & 83.00 & 70.20 & 77.10 & 60.20 & 78.00
    & 89.70 & 79.00 & 80.00 & 65.20 & 84.20
    & 93.80 & 82.10 & 46.70 & \best{82.10} & \second{95.40} & 80.60 & 83.00 \\
    MemMachine~\citep{wang2026memmachine}
    & \second{93.94} & 84.04 & 80.69 & \best{73.96} & \second{88.12}
    & 95.12 & 88.30 & \best{91.59} & 71.88 & 91.69
    & 92.86 & \best{100.00} & 55.56 & 66.67 & 86.32 & 58.65 & 74.33 \\
    Memora~\citep{xia2026memora}
    & 87.08 & 69.50 & 79.54 & 60.07 & 80.61
    & 91.80 & 78.70 & 86.60 & 59.40 & 86.30
    & \second{98.60} & 78.60 & 83.30 & \second{78.20} & \best{97.40} & \best{89.50} & \second{87.40} \\
    EverMemOS~\citep{hu2026evermemos}
    & 91.08 & \best{86.17} & \second{81.93} & 66.67 & 86.76
    & \second{96.67} & \second{91.84} & 89.72 & \second{76.04} & \second{93.05}
    & 97.14 & 85.71 & 93.33 & 73.68 & 89.74 & 77.44 & 83.00 \\
    \methodrule
    \cellcolor{rawraghighlight}\textbf{\rawrag}
    & \best{94.57} & \second{85.93} & \best{82.04} & \second{73.61} & \best{89.07}
    & \best{96.79} & \best{92.32} & \second{90.86} & \best{80.21} & \best{93.70}
    & \best{100.00} & \second{98.21} & \best{100.00} & 77.69 & 91.03 & 81.45 & \best{87.53} \\
    \bottomrule
  \end{tabular}%
  }
\end{table}

\section{The RAW-RAG}
\label{sec:raw-rag}

Our motivation is to test whether agentic memory systems improve
long-term conversation understanding beyond retrieval over raw history.
\rawrag{} (Figure~\ref{fig:raw-rag-pipeline}(c)) provides a simple reference,
leaving evidence interpretation and state reconstruction to the reader.

\rawrag{} indexes source history without
LLM-based extraction, summarization, graph construction, or mutation.
Each conversation is
serialized with message timestamps and image captions, split into 256-token
windows with 32-token overlap, and embedded in a flat index. These embeddings
support semantic matching while the original window text remains available
as reader evidence.

At query time, the question is embedded and cosine similarity retrieves the
128 closest windows. A cross-encoder jointly scores each question--window
pair for relevance and selects the top 15.
The selected evidence is then supplied with the question to the
reader, which resolves temporal order and relevant state changes when
generating the answer. Top-15 limits the number of selected items.
Appendix~\ref{app:method} specifies the full pipeline.

\section{RAG Is Stronger Than We Thought}
\label{sec:results}

This section presents our experiments evaluating \rawrag{} against agentic
memory systems, including benchmark comparisons, ablations, and efficiency.

\subsection{Benchmarks and baselines}

We evaluate memory-augmented conversation on four benchmarks, namely LoCoMo (10 conversations; 1,540 non-adversarial questions on
single-hop, multi-hop, temporal, and open-domain reasoning)
\citep{maharana2024locomo}, LongMemEval (500 examples on recall, temporal
reasoning, preferences, and updates) \citep{wu2025longmemeval}, MEME
(100 episodes; 1,188 phase-specific questions on dynamic memory updates and complex memory dependency understanding)
\citep{jung2026meme}, and MemoryAgentBench (2,071 questions from 14 sources on accurate 
retrieval, test-time learning, long-range understanding, and conflict resolution)
\citep{hu2026memoryagentbench}.

Baselines include EverMemOS~\citep{hu2026evermemos},
MemOS~\citep{li2025memos}, EMem~\citep{zhou2025simple},
SimpleMem~\citep{liu2026simplemem}, MemU~\citep{nevamind2025memu},
MemMachine~\citep{wang2026memmachine}, Mem0~\citep{chhikara2025mem0},
Memora~\citep{xia2026memora}, Nemori~\citep{nan2025nemori},
MemoryOS~\citep{kang2025memoryos}, A-MEM~\citep{xu2025amem},
and RecMem~\citep{dai2026recmem},
alongside Full-context.
On LoCoMo and LongMemEval, we evaluate \rawrag{} under settings fully aligned
with the corresponding comparison settings, as detailed in
Appendix~\ref{app:results}.
On MEME and MemoryAgentBench, all systems follow the same evaluation procedure.
Comparisons under aligned evaluation settings
and their scope are presented in Appendix~\ref{app:results}.
Appendices~\ref{app:benchmark-details} and \ref{app:baseline-details} detail
benchmark tasks and baseline settings.
Appendix~\ref{app:reader-context-budget} reports evidence lengths and
5,000-token controls, where \rawrag{} remains competitive.

\subsection{Main results}

\begin{table}[!t]
  \caption{MEME and MemoryAgentBench scores by subset (\%).
  Bold: reader-wise column maxima excluding Full-context.
  Overall: six-task macro-average (MEME) or mean of four capability macros
  (MemoryAgentBench). Dashes: Full-context results unavailable due to
  context-length limits.}
  \label{tab:main-other}
  \centering\scriptsize
  \setlength{\tabcolsep}{2.5pt}
  \renewcommand{\arraystretch}{1.05}
  \resizebox{0.94\textwidth}{!}{%
  \begin{tabular}{l*{6}{c}>{\columncolor{yellow!15}}c
                       *{4}{c}>{\columncolor{yellow!15}}c}
    \toprule
    \multirow{2}{*}[-\dimexpr(\aboverulesep+\belowrulesep+\cmidrulewidth)/2\relax]{Method}
    & \multicolumn{7}{c}{MEME} & \multicolumn{5}{c}{MemoryAgentBench} \\
    \cmidrule(lr){2-8}\cmidrule(lr){9-13}
    & ER. & Agg. & Tr. & Del. & Cas. & Abs. & Overall
    & AR. & TTL. & LRU. & CR. & Overall \\
    \methodrule
    \multicolumn{13}{>{\columncolor{gray!15}}c}{\modelname{GPT-4o-mini}} \\
    \methodrule
    \textit{Full-context}
    & 93.00 & 10.00 & 44.00 & 16.00 & 3.66 & 9.23 & 29.31
    & -- & -- & -- & -- & -- \\
    \specialrule{\lightrulewidth}{\aboverulesep}{\belowrulesep}
    MemU~\citep{nevamind2025memu}
    & 52.00 & 6.00 & 31.00 & 19.00 & 1.22 & 10.00 & 19.87
    & 52.30 & 9.12 & 22.54 & 20.50 & 26.12 \\
    EverMemOS~\citep{hu2026evermemos}
    & 44.00 & 3.00 & 50.00 & 20.00 & 2.44 & 10.00 & 21.57
    & 43.38 & \second{48.30} & 24.77 & 29.50 & 36.49 \\
    MemOS~\citep{li2025memos}
    & 78.00 & \second{19.00} & 49.00 & 21.00 & 3.66 & 10.77 & 30.24
    & 52.03 & 45.73 & \best{33.13} & 12.00 & 35.72 \\
    Mem0~\citep{chhikara2025mem0}
    & 70.00 & 5.00 & 61.00 & 14.00 & \best{7.32} & 9.23 & 27.76
    & 57.90 & 46.75 & 20.47 & \second{40.00} & 41.28 \\
    EMem~\citep{zhou2025simple}
    & 48.00 & \best{36.00} & 63.00 & \best{37.00} & \second{6.71} & \best{12.31} & 33.84
    & 57.60 & \best{49.07} & 17.11 & 30.00 & 38.44 \\
    MemMachine~\citep{wang2026memmachine}
    & \best{98.00} & 9.00 & \second{73.00} & \second{29.00} & 4.88 & 8.46 & \second{37.06}
    & 58.12 & 25.12 & 31.38 & 27.50 & 35.53 \\
    Memora~\citep{xia2026memora}
    & 93.00 & 11.00 & 58.00 & 23.00 & 6.10 & 10.00 & 33.52
    & \best{68.78} & 47.77 & 26.82 & 32.00 & \second{43.84} \\
    \methodrule
    \cellcolor{rawraghighlight}\textbf{\rawrag}
    & \second{97.00} & \second{19.00} & \best{78.00} & 26.00 & 4.27 & \second{11.54} & \best{39.30}
    & \second{67.85} & 47.68 & \second{31.74} & \best{46.00} & \best{48.32} \\
    \methodrule
    \multicolumn{13}{>{\columncolor{gray!15}}c}{\modelname{DeepSeek-V4-Flash}} \\
    \methodrule
    \textit{Full-context}
    & 97.00 & 59.00 & 88.00 & 63.00 & 48.17 & 23.85 & 63.17
    & -- & -- & -- & -- & -- \\
    \specialrule{\lightrulewidth}{\aboverulesep}{\belowrulesep}
    MemU~\citep{nevamind2025memu}
    & 50.00 & 9.00 & 23.00 & 17.00 & 3.05 & 10.77 & 18.80
    & 62.12 & 7.13 & 22.82 & 21.00 & 28.27 \\
    EverMemOS~\citep{hu2026evermemos}
    & 44.00 & 4.00 & 71.00 & 22.00 & 5.49 & 16.92 & 27.24
    & 52.85 & \best{51.02} & 32.29 & 34.00 & 42.54 \\
    MemOS~\citep{li2025memos}
    & 77.00 & \second{29.00} & 33.00 & 24.00 & 6.71 & 17.69 & 31.23
    & 56.68 & 49.27 & 32.44 & 11.50 & 37.47 \\
    Mem0~\citep{chhikara2025mem0}
    & 85.00 & 13.00 & 57.00 & 22.00 & 4.27 & 17.69 & 33.16
    & 62.92 & 48.70 & 21.19 & \second{35.00} & 41.95 \\
    EMem~\citep{zhou2025simple}
    & 48.00 & \best{36.00} & \best{87.00} & \second{38.00} & 6.10 & 15.38 & 38.41
    & 63.62 & 49.72 & 16.53 & 28.50 & 39.59 \\
    MemMachine~\citep{wang2026memmachine}
    & \best{100.00} & 11.00 & \second{85.00} & \best{48.00} & \second{7.32} & 16.92 & \best{44.71}
    & 70.08 & 27.02 & \best{44.09} & 27.00 & 42.05 \\
    Memora~\citep{xia2026memora}
    & \second{92.00} & 2.00 & 57.00 & \second{38.00} & \best{9.15} & \best{20.00} & 36.36
    & \second{75.05} & \second{50.72} & 31.76 & 34.00 & \second{47.88} \\
    \methodrule
    \cellcolor{rawraghighlight}\textbf{\rawrag}
    & \best{100.00} & 21.00 & \best{87.00} & 29.00 & 3.66 & \second{18.46} & \second{43.19}
    & \best{75.78} & 50.39 & \second{33.48} & \best{54.00} & \best{53.41} \\
    \bottomrule
  \end{tabular}%
  }
\end{table}

Tables~\ref{tab:main} and~\ref{tab:main-other} summarize the main results;
all evaluation details are provided in Appendix~\ref{app:results}.
\rawrag{} achieves the highest score among the compared memory systems in six
of seven benchmark--model settings, excluding Full-context. It leads LoCoMo
by 0.65--0.95 points, LongMemEval by 0.13 points, GPT-4o-mini MEME by
2.24 points, and MemoryAgentBench by 4.48--5.53 points. MemMachine leads
DeepSeek-V4-Flash MEME among the memory systems by 1.52 points, while
Full-context leads \rawrag{} by 19.98 points in that setting. \rawrag{}
is competitive across these tasks despite its simpler ingestion
procedure.

\subsection{Ablation studies and efficiency}
\label{sec:ablation}

We organize the analysis around three research questions: why \rawrag{}
uses its retrieval design, whether additional memory management improves
accuracy, and how efficiently it ingests, organizes and retrieves long histories.

\begin{table}[!t]
  \begin{minipage}[t]{\textwidth}
    \vspace{0pt}
    \caption{MemoryAgentBench \rawrag{} ablations.
    Gray: default; blue: retrieval variants; orange: memory-management additions.
    $a/b$: chunk size/overlap in tokens.}
    \label{tab:ablation}
    \vspace{0.5em}
    \centering\scriptsize
    {\setlength{\aboverulesep}{1.5pt}\setlength{\belowrulesep}{1.5pt}%
    \renewcommand{\arraystretch}{1.15}%
    \resizebox{0.95\linewidth}{!}{%
    \begin{tabular}{>{\columncolor{white}[\tabcolsep][\tabcolsep]}l|
                    >{\columncolor{white}[\dimexpr\tabcolsep-1.5pt\relax][\tabcolsep]}c*{6}{c}}
      \toprule
      Setting & AR. & TTL. & LRU. & CR. & Total & $\Delta$ & 95\% CI \\
      \midrule
      \rowcolor{gray!15}
      \rawrag & 67.85 & 47.68 & 31.74 & 46.00 & 48.32 & -- & -- \\
      \midrule
      \cellcolor{blue!15}$-$ reranker & 59.13 & 44.29 & 29.66 & 44.50 & 44.40 & -3.92 & [-5.80,-2.06] \\
      \cellcolor{blue!15}128/16 & 70.50 & 46.89 & 28.93 & 46.00 & 48.08 & -0.24 & [-1.54,+1.05] \\
      \cellcolor{blue!15}512/64 & 65.68 & 49.27 & 32.49 & 46.00 & 48.36 & +0.04 & [-1.15,+1.25] \\
      \cellcolor{blue!15}1024/128 & 66.28 & 48.08 & 31.79 & 46.00 & 48.04 & -0.28 & [-1.76,+1.19] \\
      \cellcolor{blue!15}$+$ display-turn dedup & 68.27 & 37.69 & 31.75 & 46.00 & 45.93 & -2.39 & [-3.53,-1.28] \\
      \cellcolor{orange!18}$+$ Atomic facts & 57.25 & 15.37 & 21.17 & 21.00 & 28.70 & -19.62 & [-21.99,-17.21] \\
      \cellcolor{orange!18}$+$ Raw summary & 67.53 & 45.89 & 29.62 & 27.00 & 42.51 & -5.81 & [-7.70,-3.90] \\
      \cellcolor{orange!18}$+$ Graph-expanded & 65.42 & 46.49 & 31.74 & 26.50 & 42.54 & -5.78 & [-7.50,-4.07] \\
      \bottomrule
    \end{tabular}}}
  \end{minipage}
  \par\medskip
  \begin{minipage}[t]{0.48\textwidth}
    \vspace{0pt}
    \caption{Ingestion and retrieval latency on 10 MEME episodes and
    110 questions (seconds). Total: Add-to-ready plus Search, without
    answer generation or evaluation.}
    \label{tab:latency}
    \vspace{0.5em}
    \centering\scriptsize
    \setlength{\tabcolsep}{2.2pt}
    \resizebox{\linewidth}{!}{%
    \begin{tabular}{l|cc>{\columncolor{gray!15}}c}
      \toprule
      System & Add$\rightarrow$ready & Search & Total \\
      \midrule
      EverMemOS~\citep{hu2026evermemos} & 8217.69 & 1083.79 & 9301.48 \\
      Mem0~\citep{chhikara2025mem0} & 4025.23 & 117.72 & 4142.95 \\
      MemMachine~\citep{wang2026memmachine} & 3104.35 & 411.39 & 3515.74 \\
      MemOS~\citep{li2025memos} & 2591.59 & 177.97 & 2769.56 \\
      EMem~\citep{zhou2025simple} & 1082.37 & 479.59 & \second{1561.97} \\
      \midrule
      \cellcolor{rawraghighlight}\textbf{\rawrag} & 50.32 & 255.59 & \best{305.91} \\
      \bottomrule
    \end{tabular}}
  \end{minipage}\hfill
  \begin{minipage}[t]{0.48\textwidth}
    \vspace{0pt}
    \centering
    \includegraphics[width=\linewidth]{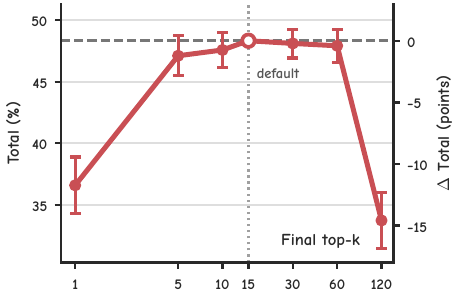}
    \par\vspace{-0.85em}
    \captionof{figure}{Final top-$k$ sensitivity: Total and change from default
    $k=15$ on MemoryAgentBench, with paired 95\% intervals.}
    \label{fig:topk-ablation}
  \end{minipage}
\end{table}

\paragraph{RQ1: Why is \rawrag{} designed this way?}
We first test the roles of reranking, retrieval depth, and chunk size.
Table~\ref{tab:ablation} and Figure~\ref{fig:topk-ablation} isolate these
choices on MemoryAgentBench using
20,000 paired, source-stratified bootstrap resamples and Holm correction.
Removing reranking lowers Total by 3.92 points ($p_{\mathrm{Holm}}=0.0012$),
while returning only one item lowers it by 11.73 points. Final-$k$
values of 5, 10, 30, and 60 stay within 1.3 points of the baseline.
Top-120 loses 14.57 points, driven by
long-range understanding and conflict resolution. Chunk sizes from 128 to 1,024
do not significantly change Total. These results support reranking and
moderate retrieval depth, without requiring the exact default chunk size or $k$.

\paragraph{RQ2: Can additional memory management improve \rawrag{}?}
We evaluate three memory-management operations: atomic-fact extraction,
summary augmentation, and graph-based retrieval expansion.
These variants reuse the chunking, embedding model, and reranker, while
changing the indexed representation or retrieval candidate set.
None improves \rawrag{} (48.32\%) on the
same 14-source MemoryAgentBench Total (Table~\ref{tab:ablation}, orange rows).
The exact construction and retrieval procedures are given in the
representation and retrieval ablation appendix (Appendix~\ref{app:agentic-mutation-ablations}),
with the extraction prompts in the prompt appendix
(Appendix~\ref{app:prompt-agentic-mutation-ablations}).
Atomic-fact extraction reduces test-time learning from 47.68\% to 15.37\%
and Total by 19.62 points, consistent with loss or disruption of usable
demonstrations (Appendix~\ref{app:additional-limitations}).
Summary augmentation largely preserves Accurate Retrieval performance but loses 5.81 points
overall, chiefly on conflict resolution. Graph expansion ties \rawrag{} on
long-range understanding yet remains 5.78 points lower overall.
Thus, these three memory-management additions reduce overall accuracy
in this setting.

\paragraph{RQ3: How efficient is \rawrag{}?}
We finally examine the computational cost of these additions and compare
system-level latency and scaling. On the 11 commonly measured non-conflict
MemoryAgentBench sources, the three additions require 21.6--109.3M
LLM tokens and 2.9--8.8 hours for Add$+$Search, versus
12 minutes for \rawrag{} (Table~\ref{tab:reverse-ablation}).
We then compare systems on the same 10-episode, 110-question MEME
slice, using one external conversation/search worker and each system's native
internal parallelism. \rawrag{} finishes Add-to-ready plus Search in 305.91
seconds (Table~\ref{tab:latency}), 5.1$\times$ faster than the next
system and 30.4$\times$ faster than EverMemOS. For scaling, we stream the
same 1,710-message BEAM-1M
conversation approaching one million tokens~\citep{tavakoli2026beam}
into each system with one external worker
(Figure~\ref{fig:teaser-results}, right panel). At the endpoint, cumulative Add time
ranges from 0.33 hours for \rawrag{} to 26.95 hours for Memora; the next-fastest
completed run, MemOS, takes 3.25 hours. Both studies exclude answer generation
and evaluation; BEAM-1M includes six completed sequential runs.
Appendix~\ref{app:reproducibility} details execution conditions.
On these workloads, \rawrag{} combines competitive accuracy
with lower ingestion and combined ingestion--retrieval costs.

\section{Why Is \rawrag{} Enough in Most Cases?}
\label{sec:mechanism}

\subsection{\rawrag{} offsets lower utilization with high coverage}

\begin{table*}[!t]
  \centering
  \begin{minipage}[t]{0.48\textwidth}
    \centering\small
    \caption{MemoryAgentBench semantic evidence coverage and conditional reader
    utilization (\%).}
    \label{tab:mab-audit}
    \vspace{0.5em}
    \begin{tabular}{l|cc}
      \toprule
      System & Coverage & Utilization \\
      \midrule
      MemU~\citep{nevamind2025memu} & 19.07 & \second{89.37} \\
      MemMachine~\citep{wang2026memmachine} & 21.54 & 84.30 \\
      EverMemOS~\citep{hu2026evermemos} & 25.11 & 80.00 \\
      MemOS~\citep{li2025memos} & 34.48 & 85.85 \\
      EMem~\citep{zhou2025simple} & 34.72 & \best{89.71} \\
      Mem0~\citep{chhikara2025mem0} & \second{40.03} & 82.39 \\
      \midrule
      \cellcolor{rawraghighlight}\textbf{\rawrag} & \best{45.29} & 84.43 \\
      \bottomrule
    \end{tabular}
  \end{minipage}\hfill
  \begin{minipage}[t]{0.48\textwidth}
    \centering\small
    \caption{MEME coverage and utilization from deterministic checks (\%).}
    \label{tab:meme-audit}
    \vspace{0.5em}
    \begin{tabular}{l|cc}
      \toprule
      System & Coverage & Utilization \\
      \midrule
      MemU~\citep{nevamind2025memu} & 35.69 & \second{84.20} \\
      EverMemOS~\citep{hu2026evermemos} & 47.64 & 76.86 \\
      EMem~\citep{zhou2025simple} & 50.08 & \best{86.55} \\
      MemOS~\citep{li2025memos} & 62.12 & 66.67 \\
      Mem0~\citep{chhikara2025mem0} & 63.22 & 65.51 \\
      MemMachine~\citep{wang2026memmachine} & \second{70.37} & 82.66 \\
      \midrule
      \cellcolor{rawraghighlight}\textbf{\rawrag} & \best{76.35} & 73.76 \\
      \bottomrule
    \end{tabular}
  \end{minipage}
\end{table*}

Does \rawrag{} perform well because it supplies better evidence,
or because readers use that evidence more effectively? This distinction
matters: improving the reader cannot reliably compensate for missing
evidence, while broader retrieval may not resolve reasoning errors.

We measure \emph{coverage}, the fraction of questions with sufficient
retrieved evidence, and \emph{utilization}, answer accuracy conditional
on sufficient evidence. Coverage is assessed by an LLM judge on
MemoryAgentBench and task-specific deterministic checks on MEME
(Appendix~\ref{app:coverage-definitions}). Conditioning on coverage
separates evidence availability from subsequent answer generation,
although the system-specific question sets preclude
a controlled comparison of readers.

Tables~\ref{tab:mab-audit} and~\ref{tab:meme-audit} show that \rawrag{}
achieves the highest coverage on both benchmarks: 45.29\% and 76.35\%.
On MemoryAgentBench, coverage spans 19.07--45.29\%, whereas utilization
spans only 80.00--89.71\%. MEME shows the same pattern: coverage varies
by 40.66 points versus 21.04 for utilization. Thus, systems differ
more in supplying sufficient evidence than in answering correctly
once it is supplied. \rawrag{} leads in coverage, helping offset its
less-than-best conditional utilization and explaining its strong performance.

\begin{findingbox}{Finding 1}
Evidence access is a major bottleneck: \rawrag{} supplies sufficient
evidence more often, helping offset lower conditional reader utilization.
\end{findingbox}

\subsection{Raw history preserves the paths needed for reasoning}

Raw history can preserve not only answer values but also the information
needed to derive them. Useful retrieval therefore includes the order and
dependencies that make one value current.

On MEME tracking and cascade after-state questions, an LLM jointly
assesses \emph{result coverage} (required answer values) and
\emph{path coverage} (those values plus chronological order or linked
dependency rules and upstream triggers;
Appendix~\ref{app:prompt-result-path}).
Table~\ref{tab:path} reports $P(\mathrm{path}\mid\mathrm{result})$:
the fraction of result-covered questions whose retrieved contexts
also contain a complete supporting path. This conditional rate
measures how often successful value retrieval also preserves
the relations and paths needed to use those values.

\rawrag{} achieves the highest conditional path coverage (62.34\%),
followed by MemMachine (56.70\%); MemU reaches only 9.62\%.
Systems that retrieve answer values can still differ substantially
in preserving their supporting relations.

\begin{findingbox}{Finding 2}
\rawrag{} is most useful when it preserves both answer values and the
temporal or dependency relations needed to derive them.
\end{findingbox}

\begin{table}[!t]
  \begin{minipage}[t]{0.47\textwidth}
    \vspace{0pt}
    \centering\scriptsize\setlength{\tabcolsep}{2.4pt}
    \caption{MEME result and executable-path coverage judged by an LLM
    on tracking and cascade after-state questions (\%).
    The last column gives path coverage conditional on result coverage.}
    \label{tab:path}
    \diagnostictablestyle
    \resizebox{\linewidth}{!}{%
    \begin{tabular}{l|cc>{\columncolor{gray!15}}c}
      \toprule
      System & Result & Path & $P(\mathrm{path}\mid\mathrm{result})$ \\
      \midrule
      MemU~\citep{nevamind2025memu} & 59.09 & 5.68 & 9.62 \\
      MemOS~\citep{li2025memos} & 88.26 & 21.21 & 24.03 \\
      Mem0~\citep{chhikara2025mem0} & 81.06 & 38.64 & 47.66 \\
      EMem~\citep{zhou2025simple} & 76.52 & 37.88 & 49.50 \\
      EverMemOS~\citep{hu2026evermemos} & 78.03 & 39.02 & 50.00 \\
      MemMachine~\citep{wang2026memmachine} & 84.85 & 48.11 & \second{56.70} \\
      \midrule
      \cellcolor{rawraghighlight}\textbf{\rawrag} & 90.53 & 56.44 & \best{62.34} \\
      \bottomrule
    \end{tabular}}
  \end{minipage}\hfill
  \begin{minipage}[t]{0.50\textwidth}
    \vspace{0pt}
    \centering
    \includegraphics[width=\linewidth]{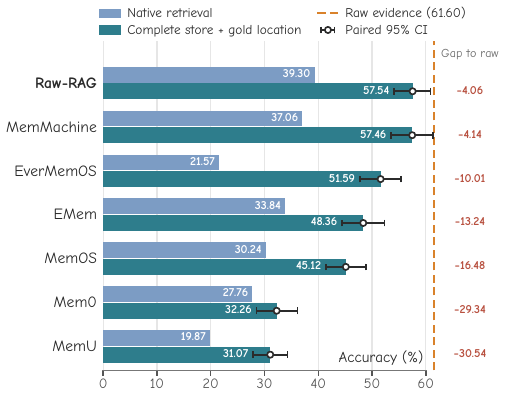}
    \captionof{figure}{GPT-4o-mini MEME: native retrieval vs.\ complete store,
    with gap to raw.}
    \label{fig:complete-store-loss}
  \end{minipage}
\end{table}

\subsection{Preserving raw evidence supports downstream recovery}

The preceding analyses identify missing evidence and incomplete dependency
paths in retrieved contexts, but do not establish where these gaps arise.
Relevant information may remain in memory but escape retrieval, or it may
have been discarded or distorted by ingestion-time processing, such as
replacing source passages with extracted facts or compressed summaries.
We therefore ask: \emph{can improved access to
the stored memories recover performance, or do these ingestion-time
operations leave information gaps that later retrieval cannot repair?}

We compare three conditions on MEME.
\emph{Native} uses each system's ordinary retrieval and readout.
The \emph{complete-store oracle} uses gold-informed selection over all
exported phase-visible memories.
The \emph{raw-evidence control} selects original annotated support sessions
using the same gold information. Both controls share a value-free task
policy and a 5,000-token packing budget.

Figure~\ref{fig:complete-store-loss} shows that the complete-store oracle
improves all seven systems, but the recovered performance varies across
stored representations. \rawrag{} and MemMachine, which retain raw
conversational content in their evaluated stores, reach 57.54\% and
57.46\%, respectively, compared with 61.60\% for raw evidence.
These results are consistent with a role for information preservation during
ingestion in downstream recovery.
The two source-preserving systems recover performance closest to the
raw-evidence control, whereas access to the complete transformed stores
leaves substantially larger deficits. Removing the packing ceiling leaves
all seven stores 7.58--29.38 points below raw evidence, with all paired
95\% intervals for these gaps excluding zero.
However, the residual gap is not a pure
measure of information loss during ingestion: even \rawrag{} falls
short, and evidence assembly and positional effects such as \emph{lost in
the middle} \citep{liu2024lostinthemiddle} may also contribute.
Appendix~\ref{app:complete-store} specifies the controls;
Appendix~\ref{app:complete-store-results} reports the full results.

\begin{findingbox}{Finding 3}
Preserving information during ingestion matters for downstream recovery:
information discarded or distorted during ingestion is difficult to
recover through retrieval alone.
\end{findingbox}

\section{Limitations and Conclusion}

\subsection{Limitations and future work}
Our evidence is limited to offline conversation on public benchmarks; it does not establish
performance for interactive, multimodal, or broader
operational settings. More discussions are in Appendix~\ref{app:additional-limitations};
raw-history approaches also raise storage and privacy concerns. Future work should evaluate a broader range of models on longer histories while accounting for explicit privacy and deletion constraints. It should also identify when structured or compressed memories offer advantages over raw retrieval, considering trade-offs among accuracy, latency and computational cost.
State reconstruction also depends on reader capability. Future evaluations should
test whether explicit state maintenance benefits weaker readers or reduces
repeated reasoning across queries.

\subsection{Conclusion}
Is agentic memory management always needed for long-term conversation
understanding? Across four
benchmarks, \rawrag{}, a naive baseline using dense retrieval and reranking
over raw history, ranks first among the compared memory systems in six of seven
benchmark--model settings. Our diagnostic analyses identify evidence access
as the larger observed bottleneck: useful evidence includes not only answer
values but also supporting relations and dependency paths. Even complete
access to transformed stores does not close the gap to raw evidence.
These findings challenge the need for complex agentic memory systems
in the evaluated conversation tasks. Improved agentic memory systems should
demonstrate gains over a well-specified naive RAG baseline and establish what the added
complexity contributes.

\endgroup

\section*{Acknowledgments}
This work is supported by the Fundamental and Interdisciplinary Disciplines
Breakthrough Plan of the Ministry of Education of China (No. JYB2025XDXM101),
the National Natural Science Foundation of China (No. 62272374, No. 62192781),
the Natural Science Foundation of Shaanxi Province (No. 2024JC-JCQN-62),
the State Key Laboratory of Communication Content Cognition under Grant
No. A202502, and the Key Research and Development Project in Shaanxi Province
(No. 2023GXLH-024).

\bibliography{iclr2027_conference}

\begin{thebibliography}{59}
\providecommand{\natexlab}[1]{#1}
\providecommand{\url}[1]{\texttt{#1}}
\expandafter\ifx\csname urlstyle\endcsname\relax
  \providecommand{\doi}[1]{doi: #1}\else
  \providecommand{\doi}{doi: \begingroup \urlstyle{rm}\Url}\fi

\bibitem[Asai et~al.(2024)Asai, Wu, Wang, Sil, and Hajishirzi]{asai2024selfrag}
Akari Asai, Zeqiu Wu, Yizhong Wang, Avirup Sil, and Hannaneh Hajishirzi.
\newblock {Self-RAG}: Learning to retrieve, generate, and critique through
  self-reflection.
\newblock In \emph{International Conference on Learning Representations}, 2024.

\bibitem[Borgeaud et~al.(2022)Borgeaud, Mensch, Hoffmann, Cai, Rutherford,
  Millican, van~den Driessche, Lespiau, Damoc, Clark, de~Las~Casas, Guy,
  Menick, Ring, Hennigan, Huang, Maggiore, Jones, Cassirer, Brock, Paganini,
  Irving, Vinyals, Osindero, Simonyan, Rae, Elsen, and
  Sifre]{borgeaud2022retro}
Sebastian Borgeaud, Arthur Mensch, Jordan Hoffmann, Trevor Cai, Eliza
  Rutherford, Katie Millican, George van~den Driessche, Jean-Baptiste Lespiau,
  Bogdan Damoc, Aidan Clark, Diego de~Las~Casas, Aurelia Guy, Jacob Menick,
  Roman Ring, Tom Hennigan, Saffron Huang, Loren Maggiore, Chris Jones, Albin
  Cassirer, Andy Brock, Michela Paganini, Geoffrey Irving, Oriol Vinyals, Simon
  Osindero, Karen Simonyan, Jack~W. Rae, Erich Elsen, and Laurent Sifre.
\newblock Improving language models by retrieving from trillions of tokens.
\newblock In \emph{Proceedings of the 39th International Conference on Machine
  Learning}, volume 162, pp.\  2206--2240, 2022.

\bibitem[Chen et~al.(2023)Chen, Pasunuru, Weston, and
  Celikyilmaz]{chen2023memwalker}
Howard Chen, Ramakanth Pasunuru, Jason Weston, and Asli Celikyilmaz.
\newblock Walking down the memory maze: Beyond context limit through
  interactive reading.
\newblock \emph{arXiv preprint arXiv:2310.05029}, 2023.

\bibitem[Chhikara et~al.(2025)Chhikara, Khant, Aryan, Singh, and
  Yadav]{chhikara2025mem0}
Prateek Chhikara, Dev Khant, Saket Aryan, Taranjeet Singh, and Deshraj Yadav.
\newblock {Mem0}: Building production-ready {AI} agents with scalable long-term
  memory.
\newblock \emph{arXiv preprint arXiv:2504.19413}, 2025.

\bibitem[Dai et~al.(2026)Dai, Deng, Guan, Tian, Yao, Yan, and
  Cheng]{dai2026recmem}
Zijie Dai, Shiyuan Deng, Sheng Guan, Yizhou Tian, Xin Yao, Xiao Yan, and James
  Cheng.
\newblock {RecMem}: Recurrence-based memory consolidation for efficient and
  effective long-running {LLM} agents.
\newblock In \emph{Findings of the Association for Computational Linguistics:
  ACL 2026}, pp.\  32353--32376, 2026.
\newblock \doi{10.18653/v1/2026.findings-acl.1619}.
\newblock URL \url{https://aclanthology.org/2026.findings-acl.1619/}.

\bibitem[Edge et~al.(2024)Edge, Trinh, Cheng, Bradley, Chao, Mody, Truitt,
  Metropolitansky, Ness, and Larson]{edge2024graphrag}
Darren Edge, Ha~Trinh, Newman Cheng, Joshua Bradley, Alex Chao, Apurva Mody,
  Steven Truitt, Dasha Metropolitansky, Robert~Osazuwa Ness, and Jonathan
  Larson.
\newblock From local to global: A {Graph RAG} approach to query-focused
  summarization.
\newblock \emph{arXiv preprint arXiv:2404.16130}, 2024.

\bibitem[Fang et~al.(2026)Fang, Deng, Xu, Jiang, Tang, Xu, Deng, Yao, Wang,
  Qiao, Chen, and Zhang]{fang2025lightmem}
Jizhan Fang, Xinle Deng, Haoming Xu, Ziyan Jiang, Yuqi Tang, Ziwen Xu, Shumin
  Deng, Yunzhi Yao, Mengru Wang, Shuofei Qiao, Huajun Chen, and Ningyu Zhang.
\newblock {LightMem}: Lightweight and efficient memory-augmented generation.
\newblock In \emph{International Conference on Learning Representations}, 2026.
\newblock URL
  \url{https://proceedings.iclr.cc/paper_files/paper/2026/hash/a05b72653ec5b473732129829ae04195-Abstract-Conference.html}.

\bibitem[Guti{\'e}rrez et~al.(2024)Guti{\'e}rrez, Shu, Gu, Yasunaga, and
  Su]{gutierrez2024hipporag}
Bernal~Jim{\'e}nez Guti{\'e}rrez, Yiheng Shu, Yu~Gu, Michihiro Yasunaga, and
  Yu~Su.
\newblock {HippoRAG}: Neurobiologically inspired long-term memory for large
  language models.
\newblock In \emph{Advances in Neural Information Processing Systems}, 2024.

\bibitem[Hu et~al.(2023)Hu, Fu, Du, Luo, Zhao, and Zhao]{hu2023chatdb}
Chenxu Hu, Jie Fu, Chenzhuang Du, Simian Luo, Junbo Zhao, and Hang Zhao.
\newblock {ChatDB}: Augmenting {LLM}s with databases as their symbolic memory.
\newblock \emph{arXiv preprint arXiv:2306.03901}, 2023.

\bibitem[Hu et~al.(2026{\natexlab{a}})Hu, Gao, Zhou, Xu, Bai, Li, Zhang, Li,
  Zhang, Bing, and Deng]{hu2026evermemos}
Chuanrui Hu, Xingze Gao, Zuyi Zhou, Dannong Xu, Yi~Bai, Xintong Li, Hui Zhang,
  Tong Li, Chong Zhang, Lidong Bing, and Yafeng Deng.
\newblock {EverMemOS}: A self-organizing memory operating system for structured
  long-horizon reasoning.
\newblock In \emph{Proceedings of the 64th Annual Meeting of the Association
  for Computational Linguistics (Volume 1: Long Papers)}, pp.\  45836--45853,
  2026{\natexlab{a}}.
\newblock \doi{10.18653/v1/2026.acl-long.2125}.
\newblock URL \url{https://aclanthology.org/2026.acl-long.2125/}.

\bibitem[Hu et~al.(2026{\natexlab{b}})Hu, Wang, and
  McAuley]{hu2026memoryagentbench}
Yuanzhe Hu, Yu~Wang, and Julian McAuley.
\newblock Evaluating memory in {LLM} agents via incremental multi-turn
  interactions.
\newblock In \emph{International Conference on Learning Representations},
  2026{\natexlab{b}}.
\newblock URL \url{https://openreview.net/forum?id=DT7JyQC3MR}.

\bibitem[Hu et~al.(2026{\natexlab{c}})Hu, Zhu, Zhao, Liang, Yan, He, and
  Gui]{hu2026xmemory}
Zhanghao Hu, Qinglin Zhu, Runcong Zhao, Di~Liang, Hanqi Yan, Yulan He, and Lin
  Gui.
\newblock Beyond {RAG} for agent memory: Retrieval by decoupling and
  aggregation.
\newblock \emph{arXiv preprint arXiv:2602.02007}, 2026{\natexlab{c}}.

\bibitem[Izacard \& Grave(2021)Izacard and Grave]{izacard2021fid}
Gautier Izacard and Edouard Grave.
\newblock Leveraging passage retrieval with generative models for open domain
  question answering.
\newblock In \emph{Proceedings of the 16th Conference of the European Chapter
  of the Association for Computational Linguistics: Main Volume}, pp.\
  874--880, 2021.
\newblock \doi{10.18653/v1/2021.eacl-main.74}.

\bibitem[Jiang et~al.(2025)Jiang, Yuan, Shen, Hao, Xu, Chen, Liu, Vijjini, He,
  Yu, Poovendran, Wornell, Ungar, Roth, Chen, and
  Taylor]{jiang2025personamemv2}
Bowen Jiang, Yuan Yuan, Maohao Shen, Zhuoqun Hao, Zhangchen Xu, Zichen Chen,
  Ziyi Liu, Anvesh~Rao Vijjini, Jiashu He, Hanchao Yu, Radha Poovendran,
  Gregory Wornell, Lyle Ungar, Dan Roth, Sihao Chen, and Camillo~Jose Taylor.
\newblock {PersonaMem-v2}: Towards personalized intelligence via learning
  implicit user personas and agentic memory.
\newblock \emph{arXiv preprint arXiv:2512.06688}, 2025.

\bibitem[Jung et~al.(2026)Jung, Rubinstein, Uselis, Yun, and Oh]{jung2026meme}
Seokwon Jung, Alexander Rubinstein, Arnas Uselis, Sangdoo Yun, and Seong~Joon
  Oh.
\newblock {MEME}: Multi-entity \& evolving memory evaluation.
\newblock \emph{arXiv preprint arXiv:2605.12477}, 2026.
\newblock Code and data: \url{https://github.com/SeokwonJung-Jay/MEME-public}.

\bibitem[Kang et~al.(2025)Kang, Ji, Zhao, and Bai]{kang2025memoryos}
Jiazheng Kang, Mingming Ji, Zhe Zhao, and Ting Bai.
\newblock Memory {OS} of {AI} agent.
\newblock In \emph{Proceedings of the 2025 Conference on Empirical Methods in
  Natural Language Processing}, pp.\  25961--25970, 2025.
\newblock \doi{10.18653/v1/2025.emnlp-main.1318}.
\newblock URL \url{https://aclanthology.org/2025.emnlp-main.1318/}.

\bibitem[Khattab \& Zaharia(2020)Khattab and Zaharia]{khattab2020colbert}
Omar Khattab and Matei Zaharia.
\newblock {ColBERT}: Efficient and effective passage search via contextualized
  late interaction over {BERT}.
\newblock In \emph{Proceedings of the 43rd International ACM SIGIR Conference
  on Research and Development in Information Retrieval}, 2020.

\bibitem[Kim et~al.(2025)Kim, Lee, Kim, Kim, and Cho]{kim2025premem}
Sangyeop Kim, Yohan Lee, Sanghwa Kim, Hyunjong Kim, and Sungzoon Cho.
\newblock Pre-storage reasoning for episodic memory: Shifting inference burden
  to memory for personalized dialogue.
\newblock In \emph{Findings of the Association for Computational Linguistics:
  EMNLP 2025}, pp.\  22096--22113, 2025.
\newblock \doi{10.18653/v1/2025.findings-emnlp.1204}.
\newblock URL \url{https://aclanthology.org/2025.findings-emnlp.1204/}.

\bibitem[Lewis et~al.(2020)Lewis, Perez, Piktus, Petroni, Karpukhin, Goyal,
  K{\"u}ttler, Lewis, Yih, Rockt{\"a}schel, Riedel, and Kiela]{lewis2020rag}
Patrick Lewis, Ethan Perez, Aleksandra Piktus, Fabio Petroni, Vladimir
  Karpukhin, Naman Goyal, Heinrich K{\"u}ttler, Mike Lewis, Wen-tau Yih, Tim
  Rockt{\"a}schel, Sebastian Riedel, and Douwe Kiela.
\newblock Retrieval-augmented generation for knowledge-intensive {NLP} tasks.
\newblock In \emph{Advances in Neural Information Processing Systems},
  volume~33, 2020.

\bibitem[Li et~al.(2026)Li, Zhang, Xu, Du, Fu, and Chen]{li2026refind}
Ruizhe Li, Licheng Zhang, Benfeng Xu, Mingxuan Du, Zheren Fu, and Weidong Chen.
\newblock When your agent opens the chat app: Agent-controlled search over raw
  chat logs rivals structured memory.
\newblock \emph{arXiv preprint arXiv:2608.12888}, 2026.

\bibitem[Li et~al.(2025)Li, Song, Wang, Niu, Chen, Yang, Xi, Lai, Zhao, Wang,
  Ren, Lin, Huo, Chen, Chen, Li, Yin, Yu, Tang, Yang, Xu, and
  Xiong]{li2025memos}
Zhiyu Li, Shichao Song, Hanyu Wang, Simin Niu, Ding Chen, Jiawei Yang, Chenyang
  Xi, Huayi Lai, Jihao Zhao, Yezhaohui Wang, Junpeng Ren, Zehao Lin, Jiahao
  Huo, Tianyi Chen, Kai Chen, Kehang Li, Zhiqiang Yin, Qingchen Yu, Bo~Tang,
  Hongkang Yang, Zhi-Qin~John Xu, and Feiyu Xiong.
\newblock {MemOS}: An operating system for memory-augmented generation ({MAG})
  in large language models.
\newblock \emph{arXiv preprint arXiv:2505.22101}, 2025.

\bibitem[Liao et~al.(2026)Liao, Wu, Hou, Liu, Wu, and Wang]{liao2026leanmem}
Yuxin Liao, Le~Wu, Min Hou, Hao Liu, Han Wu, and Zishu Wang.
\newblock Leanmem: Simple and efficient long-term memory for {LLM} agents.
\newblock \emph{arXiv preprint arXiv:2608.03463}, 2026.

\bibitem[Liu et~al.(2026)Liu, Su, Xia, Han, Zheng, Xie, Ding, and
  Yao]{liu2026simplemem}
Jiaqi Liu, Yaofeng Su, Peng Xia, Siwei Han, Zeyu Zheng, Cihang Xie, Mingyu
  Ding, and Huaxiu Yao.
\newblock {SimpleMem}: Efficient lifelong memory for {LLM} agents.
\newblock \emph{arXiv preprint arXiv:2601.02553}, 2026.

\bibitem[Liu et~al.(2023)Liu, Yang, Shen, Hu, Zhang, Gu, and
  Zhang]{liu2023thinkinmemory}
Lei Liu, Xiaoyan Yang, Yue Shen, Binbin Hu, Zhiqiang Zhang, Jinjie Gu, and
  Guannan Zhang.
\newblock Think-in-memory: Recalling and post-thinking enable {LLM}s with
  long-term memory.
\newblock \emph{arXiv preprint arXiv:2311.08719}, 2023.

\bibitem[Liu et~al.(2024)Liu, Lin, Hewitt, Paranjape, Bevilacqua, Petroni, and
  Liang]{liu2024lostinthemiddle}
Nelson~F. Liu, Kevin Lin, John Hewitt, Ashwin Paranjape, Michele Bevilacqua,
  Fabio Petroni, and Percy Liang.
\newblock Lost in the middle: How language models use long contexts.
\newblock \emph{Transactions of the Association for Computational Linguistics},
  12:\penalty0 157--173, 2024.

\bibitem[Ma et~al.(2026)Ma, Nan, and Wu]{ma2025nemori}
Wenquan Ma, Jiayan Nan, and WenLong Wu.
\newblock What deserves memory: Adaptive memory distillation for {LLM} agents.
\newblock In \emph{Proceedings of the 64th Annual Meeting of the Association
  for Computational Linguistics (Volume 1: Long Papers)}, pp.\  34789--34812,
  2026.

\bibitem[Maharana et~al.(2024)Maharana, Lee, Tulyakov, Bansal, Barbieri, and
  Fang]{maharana2024locomo}
Adyasha Maharana, Dong-Ho Lee, Sergey Tulyakov, Mohit Bansal, Francesco
  Barbieri, and Yuwei Fang.
\newblock Evaluating very long-term conversational memory of {LLM} agents.
\newblock In \emph{Proceedings of the 62nd Annual Meeting of the Association
  for Computational Linguistics (Volume 1: Long Papers)}, pp.\  13851--13870,
  2024.
\newblock \doi{10.18653/v1/2024.acl-long.747}.

\bibitem[{Memobase Team}(2025)]{memobase2025locomo}
{Memobase Team}.
\newblock {LoCoMo} benchmark for various memory backends.
\newblock
  \url{https://github.com/memodb-io/memobase/tree/main/docs/experiments/locomo-benchmark},
  2025.
\newblock Public evaluation configuration and results; accessed September 26,
  2026.

\bibitem[Modarressi et~al.(2023)Modarressi, Imani, Fayyaz, and
  Sch{\"u}tze]{modarressi2023retllm}
Ali Modarressi, Ayyoob Imani, Mohsen Fayyaz, and Hinrich Sch{\"u}tze.
\newblock {RET-LLM}: Towards a general read-write memory for large language
  models.
\newblock \emph{arXiv preprint arXiv:2305.14322}, 2023.

\bibitem[Nan et~al.(2025)Nan, Ma, Wu, and Chen]{nan2025nemori}
Jiayan Nan, Wenquan Ma, Wenlong Wu, and Yize Chen.
\newblock {Nemori}: Self-organizing agent memory inspired by cognitive science.
\newblock \emph{arXiv preprint arXiv:2508.03341v1}, 2025.

\bibitem[{NevaMind-AI}(2025)]{nevamind2025memu}
{NevaMind-AI}.
\newblock {memU}.
\newblock \url{https://github.com/NevaMind-AI/memU}, 2025.
\newblock Accessed September 26, 2026.

\bibitem[Nogueira \& Cho(2019)Nogueira and Cho]{nogueira2019bert}
Rodrigo Nogueira and Kyunghyun Cho.
\newblock Passage re-ranking with {BERT}.
\newblock \emph{arXiv preprint arXiv:1901.04085}, 2019.

\bibitem[Packer et~al.(2023)Packer, Wooders, Lin, Fang, Patil, Stoica, and
  Gonzalez]{packer2023memgpt}
Charles Packer, Sarah Wooders, Kevin Lin, Vivian Fang, Shishir~G. Patil, Ion
  Stoica, and Joseph~E. Gonzalez.
\newblock {MemGPT}: Towards {LLM}s as operating systems.
\newblock \emph{arXiv preprint arXiv:2310.08560}, 2023.

\bibitem[Pan et~al.(2025)Pan, Wu, Jiang, Luo, Cheng, Li, Yang, Lin, Zhao, Qiu,
  and Gao]{pan2025secom}
Zhuoshi Pan, Qianhui Wu, Huiqiang Jiang, Xufang Luo, Hao Cheng, Dongsheng Li,
  Yuqing Yang, Chin-Yew Lin, H.~Vicky Zhao, Lili Qiu, and Jianfeng Gao.
\newblock {SeCom}: On memory construction and retrieval for personalized
  conversational agents.
\newblock In \emph{International Conference on Learning Representations}, 2025.
\newblock URL
  \url{https://proceedings.iclr.cc/paper_files/paper/2025/hash/e56f394bbd4f0ec81393d767caa5a31b-Abstract-Conference.html}.

\bibitem[Park et~al.(2023)Park, O'Brien, Cai, Morris, Liang, and
  Bernstein]{park2023generative}
Joon~Sung Park, Joseph~C. O'Brien, Carrie~J. Cai, Meredith~Ringel Morris, Percy
  Liang, and Michael~S. Bernstein.
\newblock Generative agents: Interactive simulacra of human behavior.
\newblock In \emph{Proceedings of the 36th Annual ACM Symposium on User
  Interface Software and Technology}, 2023.

\bibitem[Rasmussen et~al.(2025)Rasmussen, Paliychuk, Beauvais, Ryan, and
  Chalef]{rasmussen2025zep}
Preston Rasmussen, Pavlo Paliychuk, Travis Beauvais, Jack Ryan, and Daniel
  Chalef.
\newblock Zep: A temporal knowledge graph architecture for agent memory.
\newblock \emph{arXiv preprint arXiv:2501.13956}, 2025.

\bibitem[Reimers \& Gurevych(2019)Reimers and
  Gurevych]{reimers2019sentencebert}
Nils Reimers and Iryna Gurevych.
\newblock Sentence-{BERT}: Sentence embeddings using siamese {BERT}-networks.
\newblock In \emph{Proceedings of the 2019 Conference on Empirical Methods in
  Natural Language Processing and the 9th International Joint Conference on
  Natural Language Processing (EMNLP-IJCNLP)}, pp.\  3982--3992, 2019.
\newblock \doi{10.18653/v1/D19-1410}.

\bibitem[Robertson \& Zaragoza(2009)Robertson and
  Zaragoza]{robertson2009probabilistic}
Stephen Robertson and Hugo Zaragoza.
\newblock The probabilistic relevance framework: {BM25} and beyond.
\newblock \emph{Foundations and Trends in Information Retrieval}, 3\penalty0
  (4):\penalty0 333--389, 2009.

\bibitem[Sarthi et~al.(2024)Sarthi, Abdullah, Tuli, Khanna, Goldie, and
  Manning]{sarthi2024raptor}
Parth Sarthi, Salman Abdullah, Aditi Tuli, Shubh Khanna, Anna Goldie, and
  Christopher~D. Manning.
\newblock {RAPTOR}: Recursive abstractive processing for tree-organized
  retrieval.
\newblock In \emph{International Conference on Learning Representations}, 2024.

\bibitem[Shinn et~al.(2023)Shinn, Cassano, Gopinath, Narasimhan, and
  Yao]{shinn2023reflexion}
Noah Shinn, Federico Cassano, Ashwin Gopinath, Karthik Narasimhan, and Shunyu
  Yao.
\newblock Reflexion: Language agents with verbal reinforcement learning.
\newblock In \emph{Advances in Neural Information Processing Systems}, 2023.

\bibitem[Sumers et~al.(2024)Sumers, Yao, Narasimhan, and
  Griffiths]{sumers2023coala}
Theodore~R. Sumers, Shunyu Yao, Karthik Narasimhan, and Thomas~L. Griffiths.
\newblock Cognitive architectures for language agents.
\newblock \emph{Transactions on Machine Learning Research}, 2024.

\bibitem[Sun et~al.(2025)Sun, Zhang, and Zeng]{sun2025pamu}
Haoran Sun, Zekun Zhang, and Shaoning Zeng.
\newblock Preference-aware memory update for long-term {LLM} agents.
\newblock \emph{arXiv preprint arXiv:2510.09720}, 2025.

\bibitem[Tavakoli et~al.(2026)Tavakoli, Salemi, Ye, Abdalla, Zamani, and
  Mitchell]{tavakoli2026beam}
Mohammad Tavakoli, Alireza Salemi, Carrie Ye, Mohamed Abdalla, Hamed Zamani,
  and J.~Ross Mitchell.
\newblock Beyond a million tokens: Benchmarking and enhancing long-term memory
  in {LLM}s.
\newblock In \emph{International Conference on Learning Representations}, 2026.

\bibitem[Wang et~al.(2026{\natexlab{a}})Wang, Liang, Yang, Huang, Wu, Wu, Ma,
  and Li]{liang2023scm}
Bing Wang, Xinnian Liang, Jian Yang, Hui Huang, Zhenhe Wu, ShuangZhi Wu, Zejun
  Ma, and Zhoujun Li.
\newblock {SCM}: Enhancing large language model with self-controlled memory
  framework.
\newblock In \emph{Database Systems for Advanced Applications: 30th
  International Conference, DASFAA 2025, Singapore, Singapore, May 26--29,
  2025, Proceedings, Part VI}, volume 15991 of \emph{Lecture Notes in Computer
  Science}, pp.\  188--203, 2026{\natexlab{a}}.
\newblock \doi{10.1007/978-981-95-4158-4_12}.

\bibitem[Wang et~al.(2026{\natexlab{b}})Wang, Li, Guan, Shi, Lu, and
  Jiang]{wang2026cuemem}
Changjian Wang, Rongzhen Li, Weili Guan, Shuming Shi, Quan Lu, and Ning Jiang.
\newblock Cuemem: Cue-guided context reconstruction for long-term
  conversational memory.
\newblock \emph{arXiv preprint arXiv:2609.12354}, 2026{\natexlab{b}}.

\bibitem[Wang et~al.(2024{\natexlab{a}})Wang, Xie, Jiang, Mandlekar, Xiao, Zhu,
  Fan, and Anandkumar]{wang2023voyager}
Guanzhi Wang, Yuqi Xie, Yunfan Jiang, Ajay Mandlekar, Chaowei Xiao, Yuke Zhu,
  Linxi Fan, and Anima Anandkumar.
\newblock {Voyager}: An open-ended embodied agent with large language models.
\newblock \emph{Transactions on Machine Learning Research}, 2024{\natexlab{a}}.

\bibitem[Wang et~al.(2024{\natexlab{b}})Wang, Ma, Feng, Zhang, Yang, Zhang,
  Chen, Tang, Chen, Lin, Zhao, Wei, and Wen]{wang2024llmagentsurvey}
Lei Wang, Chen Ma, Xueyang Feng, Zeyu Zhang, Hao Yang, Jingsen Zhang, Zhiyuan
  Chen, Jiakai Tang, Xu~Chen, Yankai Lin, Wayne~Xin Zhao, Zhewei Wei, and
  Ji-Rong Wen.
\newblock A survey on large language model based autonomous agents.
\newblock \emph{Frontiers of Computer Science}, 18\penalty0 (6):\penalty0
  186345, 2024{\natexlab{b}}.

\bibitem[Wang et~al.(2025)Wang, Fu, Cao, Wang, Tian, and
  Ding]{wang2024recursively}
Qingyue Wang, Yanhe Fu, Yanan Cao, Shuai Wang, Zhiliang Tian, and Liang Ding.
\newblock Recursively summarizing enables long-term dialogue memory in large
  language models.
\newblock \emph{Neurocomputing}, 639:\penalty0 130193, 2025.
\newblock \doi{10.1016/j.neucom.2025.130193}.

\bibitem[Wang et~al.(2026{\natexlab{c}})Wang, Yu, Love, Zhang, Wong, Scargall,
  and Fan]{wang2026memmachine}
Shu Wang, Edwin Yu, Oscar Love, Tom Zhang, Tom Wong, Steve Scargall, and
  Charles Fan.
\newblock {MemMachine}: A ground-truth-preserving memory system for
  personalized {AI} agents.
\newblock \emph{arXiv preprint arXiv:2604.04853}, 2026{\natexlab{c}}.

\bibitem[Wu et~al.(2025{\natexlab{a}})Wu, Wang, Yu, Zhang, Chang, and
  Yu]{wu2025longmemeval}
Di~Wu, Hongwei Wang, Wenhao Yu, Yuwei Zhang, Kai-Wei Chang, and Dong Yu.
\newblock {LongMemEval}: Benchmarking chat assistants on long-term interactive
  memory.
\newblock In \emph{International Conference on Learning Representations},
  2025{\natexlab{a}}.

\bibitem[Wu \& Zhu(2026)Wu and Zhu]{wuzhu2026agentzeromemory}
Ming Wu and Pengyuan Zhu.
\newblock Agent zero memory: Provenance-aware long-term memory for {LLM}
  agents.
\newblock \emph{arXiv preprint arXiv:2608.29606}, 2026.

\bibitem[Wu et~al.(2025{\natexlab{b}})Wu, Zhang, Liang, and Liu]{wu2025sgmem}
Yaxiong Wu, Yongyue Zhang, Sheng Liang, and Yong Liu.
\newblock {SGMem}: Sentence graph memory for long-term conversational agents.
\newblock \emph{arXiv preprint arXiv:2509.21212}, 2025{\natexlab{b}}.

\bibitem[Xia et~al.(2026)Xia, Zhang, Dixit, Harimurugan, Wang, R{\"u}hle, Sim,
  Bansal, and Rajmohan]{xia2026memora}
Menglin Xia, Xuchao Zhang, Shantanu Dixit, Paramaguru Harimurugan, Rujia Wang,
  Victor R{\"u}hle, Robert Sim, Chetan Bansal, and Saravan Rajmohan.
\newblock Memora: A harmonic memory representation balancing abstraction and
  specificity.
\newblock \emph{arXiv preprint arXiv:2602.03315}, 2026.

\bibitem[Xu et~al.(2025)Xu, Liang, Mei, Gao, Tan, and Zhang]{xu2025amem}
Wujiang Xu, Zujie Liang, Kai Mei, Hang Gao, Juntao Tan, and Yongfeng Zhang.
\newblock {A-Mem}: Agentic memory for {LLM} agents.
\newblock In \emph{Advances in Neural Information Processing Systems}, 2025.

\bibitem[Xu et~al.(2026)Xu, Xu, Tian, Huang, Chen, Li, and Shen]{xux2026com}
Xiucheng Xu, Bingbing Xu, Xueyun Tian, Zihe Huang, Rongxin Chen, Yunfan Li, and
  Huawei Shen.
\newblock Chain-of-memory: Lightweight memory construction with dynamic
  evolution for {LLM} agents.
\newblock In \emph{Proceedings of the 64th Annual Meeting of the Association
  for Computational Linguistics (Volume 1: Long Papers)}, pp.\  11618--11631,
  2026.
\newblock \doi{10.18653/v1/2026.acl-long.534}.

\bibitem[Zhang et~al.(2025{\natexlab{a}})Zhang, Li, Long, Zhang, Lin, Yang,
  Xie, Yang, Liu, Lin, Huang, and Zhou]{zhang2025qwen3embedding}
Yanzhao Zhang, Mingxin Li, Dingkun Long, Xin Zhang, Huan Lin, Baosong Yang,
  Pengjun Xie, An~Yang, Dayiheng Liu, Junyang Lin, Fei Huang, and Jingren Zhou.
\newblock {Qwen3 Embedding}: Advancing text embedding and reranking through
  foundation models.
\newblock \emph{arXiv preprint arXiv:2506.05176}, 2025{\natexlab{a}}.
\newblock Reranker model card:
  \url{https://huggingface.co/Qwen/Qwen3-Reranker-4B}.

\bibitem[Zhang et~al.(2025{\natexlab{b}})Zhang, Dai, Bo, Ma, Li, Chen, Zhu,
  Dong, and Wen]{zhang2024memorysurvey}
Zeyu Zhang, Quanyu Dai, Xiaohe Bo, Chen Ma, Rui Li, Xu~Chen, Jieming Zhu,
  Zhenhua Dong, and Ji-Rong Wen.
\newblock A survey on the memory mechanism of large language model-based
  agents.
\newblock \emph{ACM Transactions on Information Systems}, 43\penalty0
  (6):\penalty0 155:1--155:47, 2025{\natexlab{b}}.
\newblock \doi{10.1145/3748302}.

\bibitem[Zhong et~al.(2024)Zhong, Guo, Gao, Ye, and Wang]{zhong2024memorybank}
Wanjun Zhong, Lianghong Guo, Qiqi Gao, He~Ye, and Yanlin Wang.
\newblock {MemoryBank}: Enhancing large language models with long-term memory.
\newblock In \emph{Proceedings of the AAAI Conference on Artificial
  Intelligence}, volume~38, pp.\  19724--19731, 2024.

\bibitem[Zhou \& Han(2025)Zhou and Han]{zhou2025simple}
Sizhe Zhou and Jiawei Han.
\newblock A simple yet strong baseline for long-term conversational memory of
  {LLM} agents.
\newblock \emph{arXiv preprint arXiv:2511.17208}, 2025.

\end{thebibliography}
\bibliographystyle{iclr2027_conference}

\appendix
\clearpage
\setcounter{topnumber}{3}
\setcounter{bottomnumber}{2}
\setcounter{totalnumber}{5}
\renewcommand{\topfraction}{0.95}
\renewcommand{\bottomfraction}{0.85}
\renewcommand{\textfraction}{0.05}
\renewcommand{\floatpagefraction}{0.95}
\setlength{\belowcaptionskip}{0.5em}
\makeatletter
\let\appendix@addtocontents\addtocontents
\renewcommand{\addtocontents}[2]{%
  \def\appendix@extension{#1}\def\appendix@toc{toc}%
  \ifx\appendix@extension\appendix@toc
    \appendix@addtocontents{apptoc}{#2}%
  \else
    \appendix@addtocontents{#1}{#2}%
  \fi
}
\begingroup
\setcounter{tocdepth}{3}
\hypersetup{linkcolor=red!65!black,linktoc=section}
\parindent=0pt
\pdfbookmark[0]{Appendix}{appendix-overview}
\section*{Appendix}
\section*{Table of Contents}
\nobreak\vspace{0.2em}
\hrule height 0.5pt
\vspace{0.6em}
\small
\setlength{\parskip}{0pt}
\renewcommand{\@pnumwidth}{2em}
\renewcommand{\@tocrmarg}{3em}
\renewcommand{\l@section}[2]{%
  \addpenalty{-\@highpenalty}%
  \vskip 0.8em plus 0.2em
  \begingroup
    \bfseries
    \renewcommand{\@dotsep}{10000}%
    \@dottedtocline{1}{1.5em}{1.8em}{#1}{#2}%
  \endgroup
  \nopagebreak[4]%
}
\renewcommand{\l@subsection}{\@dottedtocline{2}{3.3em}{2.6em}}
\renewcommand{\l@subsubsection}{\@dottedtocline{3}{5.9em}{3.4em}}
\@starttoc{apptoc}
\par\vspace{0.8em}
\hrule height 0.5pt
\endgroup
\makeatother

\section{Experimental Settings and Complete Results}
\label{app:results}
\label{app:meme-mab-results}

We supplement the main-table comparisons with evaluations of \rawrag{}
under settings fully aligned with the corresponding prior studies.
Each comparison first specifies its evaluation setting and then presents
the corresponding results. Published baseline scores are retained, and
\rawrag{} results are measured in this study.
All scores use a 0--100 scale unless stated otherwise; ``--'' means
unavailable. Full-context uses no embeddings.

The remaining sections report retrieval and memory-management ablations,
context-budget controls, diagnostic analyses with uncertainty estimates,
and efficiency measurements.

\subsection{LoCoMo results}
\label{app:locomo-results}

\paragraph{Evaluation and score averaging.}
For \rawrag{}, judge accuracy is the mean of the binary correctness judgments
across questions and judge repetitions.
LoCoMo excludes category 5 and contains 1,540 questions: single-hop
(category 4; 841), multi-hop (category 1; 282), temporal (category 2; 321),
and open-domain (category 3; 96). The \rawrag{} configurations
use 256-token windows, 32-token overlap, 128 dense candidates, and 15 reranked
chunks expanded to complete turns.
Appendix~\ref{app:prompt-locomo-judge} specifies the inherited judge
instructions, response parsing, failure scoring, and the limits of
retrospective format-failure accounting.

\subsubsection{Text-embedding-3-small setting}
\label{app:locomo-openai}

This setting uses \texttt{text-embedding-3-small}, with separate
GPT-4o-mini and GPT-4.1-mini reader blocks. \rawrag{} uses Cohere
Rerank-v3.5 and follows the corresponding evaluation setting for history
access, answer generation, and scoring. Each fixed answer
receives three GPT-4o-mini judgments; category and overall accuracy are
question-weighted means of these binary judgments.
Table~\ref{tab:locomo-openai-protocol} reports the matched comparison,
with published baseline scores from \citet{dai2026recmem},
our evaluations of SimpleMem~\citep{liu2026simplemem},
and the MemMachine and Memora results from Table~\ref{tab:main}.
SimpleMem, MemMachine, and GPT-4o-mini Memora are evaluated in this study;
GPT-4.1-mini Memora uses the published policy-retriever
result~\citep{xia2026memora}.

\begin{table}[!htbp]
  \caption{LoCoMo accuracy (\%) under the text-embedding-3-small
  evaluation setting, scored by a large language model (LLM) judge.
  Single, Multi, and Open denote single-hop, multi-hop, and open-domain
  questions. Overall is question-weighted; adversarial questions are
  excluded. Bold marks the highest score in each column within each
  reader block.}
  \label{tab:locomo-openai-protocol}
  \centering\small
  \setlength{\tabcolsep}{6pt}
  \begin{tabular}{lcccc>{\columncolor{yellow!15}}c}
    \toprule
    Method & Single & Multi & Temporal & Open & Overall \\
    \midrule
    \multicolumn{6}{>{\columncolor{gray!15}}c}{\modelname{GPT-4o-mini}} \\
    \methodrule
    MemoryOS~\citep{kang2025memoryos} & 74.55 & 58.87 & 44.24 & 46.88 & 63.64 \\
    Mem0~\citep{chhikara2025mem0} & 65.52 & 52.84 & 52.02 & 37.50 & 58.64 \\
    A-MEM~\citep{xu2025amem} & 66.83 & 52.13 & 61.99 & 30.21 & 60.84 \\
    RecMem~\citep{dai2026recmem} & 79.79 & 72.70 & 57.32 & 58.33 & 72.47 \\
    SimpleMem~\citep{liu2026simplemem} & 86.01 & 74.94 & 63.76 & 68.40 & 78.25 \\
    MemMachine~\citep{wang2026memmachine} & \second{93.94} & \second{84.04} & \second{80.69} & \best{73.96} & \second{88.12} \\
    Memora~\citep{xia2026memora} & 87.08 & 69.50 & 79.54 & 60.07 & 80.61 \\
    \methodrule
    \cellcolor{rawraghighlight}\textbf{\rawrag{}} & \best{94.57} & \best{85.93} & \best{82.04} & \second{73.61} & \best{89.07} \\
    \midrule
    \multicolumn{6}{>{\columncolor{gray!15}}c}{\modelname{GPT-4.1-mini}} \\
    \methodrule
    MemoryOS~\citep{kang2025memoryos} & 77.05 & 66.31 & 47.66 & 55.21 & 67.60 \\
    Mem0~\citep{chhikara2025mem0} & 66.23 & 58.16 & 63.86 & 44.79 & 62.92 \\
    A-MEM~\citep{xu2025amem} & 73.25 & 59.93 & 72.90 & 42.71 & 68.83 \\
    RecMem~\citep{dai2026recmem} & 86.92 & 78.37 & 75.39 & 57.29 & 81.10 \\
    SimpleMem~\citep{liu2026simplemem} & 88.15 & 79.55 & 78.82 & \second{76.04} & 83.87 \\
    MemMachine~\citep{wang2026memmachine} & \second{95.12} & \second{88.30} & \best{91.59} & 71.88 & \second{91.69} \\
    Memora~\citep{xia2026memora} & 91.80 & 78.70 & 86.60 & 59.40 & 86.30 \\
    \methodrule
    \cellcolor{rawraghighlight}\textbf{\rawrag{}} & \best{96.79} & \best{92.32} & \second{90.86} & \best{80.21} & \best{93.70} \\
    \bottomrule
  \end{tabular}
\end{table}
\subsubsection{Qwen3 retrieval setting}
\label{app:locomo-qwen}

This setting uses Qwen3-Embedding-4B and Qwen3-Reranker-4B, with
GPT-4o-mini and GPT-4.1-mini reader blocks over all 1,540 questions.
\rawrag{} follows the EverMemOS evaluation setting~\citep{hu2026evermemos},
including history access, answer generation, grading
prompts, and score averaging. The judge is GPT-4o-mini-2024-07-18 at
temperature zero, called three times per fixed answer; accuracy is the
mean of the three question-weighted accuracies.
Table~\ref{tab:locomo-qwen-protocol} presents the corresponding results,
with baseline scores from \citet{hu2026evermemos}.

\begin{table}[!htbp]
  \caption{LoCoMo LLM-judge accuracy (\%) with Qwen3-Embedding-4B and
  Qwen3-Reranker-4B. Single, Multi, and Open denote single-hop, multi-hop,
  and open-domain questions. Overall is question-weighted; adversarial
  questions are excluded. Bold marks column maxima within each reader
  block, including ties at the displayed precision.}
  \label{tab:locomo-qwen-protocol}
  \centering\small
  \setlength{\tabcolsep}{6pt}
  \begin{tabular}{lcccc>{\columncolor{yellow!15}}c}
    \toprule
    Method & Single & Multi & Temporal & Open & Overall \\
    \midrule
    \multicolumn{6}{>{\columncolor{gray!15}}c}{\modelname{GPT-4o-mini}} \\
    \methodrule
    MemoryOS~\citep{kang2025memoryos} & 62.43 & 56.50 & 37.18 & 40.28 & 54.70 \\
    Mem0~\citep{chhikara2025mem0} & 66.71 & 58.16 & 55.45 & 40.62 & 61.00 \\
    MemU~\citep{nevamind2025memu} & 72.77 & 62.41 & 33.96 & 46.88 & 61.15 \\
    MemOS~\citep{li2025memos} & 81.45 & 69.15 & 72.27 & 60.42 & 75.87 \\
    Zep~\citep{rasmussen2025zep} & 88.11 & 71.99 & 74.45 & \second{66.67} & 81.06 \\
    EverMemOS~\citep{hu2026evermemos} & \second{91.08} & \second{86.17} & \best{81.93} & \second{66.67} & \second{86.76} \\
    \methodrule
    \cellcolor{rawraghighlight}\textbf{\rawrag{}} & \best{95.28} & \best{86.88} & \second{79.96} & \best{72.57} & \best{89.13} \\
    \midrule
    \multicolumn{6}{>{\columncolor{gray!15}}c}{\modelname{GPT-4.1-mini}} \\
    \methodrule
    MemoryOS~\citep{kang2025memoryos} & 67.30 & 59.34 & 42.26 & 59.03 & 60.11 \\
    Mem0~\citep{chhikara2025mem0} & 68.97 & 61.70 & 58.26 & 50.00 & 64.20 \\
    MemU~\citep{nevamind2025memu} & 74.91 & 72.34 & 43.61 & 54.17 & 66.67 \\
    MemOS~\citep{li2025memos} & 85.37 & 79.43 & 75.08 & 64.58 & 80.76 \\
    Zep~\citep{rasmussen2025zep} & \second{90.84} & 81.91 & 77.26 & 75.00 & 85.22 \\
    EverMemOS~\citep{hu2026evermemos} & \best{96.67} & \second{91.84} & \second{89.72} & \second{76.04} & \second{93.05} \\
    \methodrule
    \cellcolor{rawraghighlight}\textbf{\rawrag{}} & \best{96.67} & \best{94.68} & \best{92.94} & \best{80.90} & \best{94.55} \\
    \bottomrule
  \end{tabular}
\end{table}
\subsubsection{Short-answer evaluation}
\label{app:locomo-mem0-like}

This comparison aligns \rawrag{} with the short-output evaluation
setting used for the Mem0 and Nemori comparisons.
We report LLM-judge accuracy ($J$), token F1, and BLEU-1 ($B_1$) for the
resulting answers.
We keep the results published by Mem0~\citep{chhikara2025mem0} separate from
Nemori's version-1 re-evaluations~\citep{nan2025nemori}: even the same named baseline
has different reported scores. Table~\ref{tab:locomo-mem0-overall} gives
overall results, and Tables~\ref{tab:locomo-mem0-categories}
and~\ref{tab:locomo-nemori-categories} give the category-level metrics.
Our evaluation averages three binary judge decisions per answer.

For \rawrag{}, $J$, F1, and BLEU-1 evaluate the same generated
answers. The lexical metrics follow
Nemori's definitions. F1 uses sets of
lowercased tokens after replacing periods, commas, exclamation marks, and
question marks with spaces. BLEU-1 uses lowercased word tokenization,
unigram weights, and sentence-level scoring with method-1 smoothing; scores
are averaged over questions.
The \rawrag{} ``minimal'' rows use GPT-4o-mini and GPT-4.1-mini with the
same shortest-answer instruction and retrieved evidence, obtained using
text-embedding-3-small and Cohere Rerank-v3.5.
The instruction asks for only the entity, date, number, or list needed to
answer the question, without explanation or background.

\paragraph{Published overall scores.}
Mem0's published GPT-4o-mini category scores, in Single / Multi /
Temporal / Open order, are 67.13 / 51.15 / 55.51 / 72.93.
Weighting these by 841 / 282 / 321 / 96 questions gives 62.14,
rather than the published overall score of 66.88. Question-level
judgments are unavailable, so the tables preserve both the published
category scores and the published aggregate without recomputing either.

\begin{table}[!htbp]
  \caption{Overall LoCoMo scores under the short-answer evaluation setting
  (0--100). Metrics are LLM-judge accuracy ($J$), token F1, and unigram
  BLEU ($B_1$). Sources M, N, and L denote the Mem0~\citep{chhikara2025mem0} paper, the Nemori~\citep{nan2025nemori}
  paper, and this study. RAG (256/2) uses 256-token chunks and top-2
  retrieval; Mem0$^g$~\citep{chhikara2025mem0} is the graph variant. ``Minimal'' requests only the
  shortest sufficient answer. Dashes indicate unavailable scores; bold
  marks column maxima per reader, excluding Full-context.}
  \label{tab:locomo-mem0-overall}
  \centering\small
  \setlength{\tabcolsep}{6pt}
  \renewcommand{\arraystretch}{1.08}
  \begin{tabular*}{\linewidth}{@{\extracolsep{\fill}}>{\columncolor{white}[0pt][\tabcolsep]}llccc@{}}
    \toprule
    Method & Source & $J$ & F1 & $B_1$ \\
    \midrule
    \multicolumn{5}{>{\columncolor{gray!15}}c}{\modelname{GPT-4o-mini}} \\
    \methodrule
    \textit{Full-context} & M & 72.90 & -- & -- \\
    \methodrule
    A-MEM~\citep{xu2025amem} & M & 48.38 & -- & -- \\
    LangMem & M & 58.10 & -- & -- \\
    Zep~\citep{rasmussen2025zep} & M & 65.99 & -- & -- \\
    OpenAI & M & 52.90 & -- & -- \\
    Mem0~\citep{chhikara2025mem0} & M & 66.88 & -- & -- \\
    Mem0$^{g}$~\citep{chhikara2025mem0} & M & 68.44 & -- & -- \\
    \addlinespace
    \textit{Full-context} & N & 72.30 & 46.20 & 37.80 \\
    \methodrule
    LangMem & N & 51.30 & 35.80 & 29.40 \\
    Zep~\citep{rasmussen2025zep} & N & 58.50 & 37.50 & 30.90 \\
    Mem0~\citep{chhikara2025mem0} & N & 61.30 & 41.50 & 34.20 \\
    Nemori~\citep{nan2025nemori} & N & \second{74.40} & \second{49.50} & \second{38.50} \\
    \methodrule
    RAG (256/2) & M & 60.97 & -- & -- \\
    RAG & N & 30.20 & 20.80 & 16.40 \\
    \cellcolor{rawraghighlight}\textbf{\rawrag{}} (minimal) & L & \best{85.78} & \best{52.58} & \best{44.50} \\
    \midrule
    \multicolumn{5}{>{\columncolor{gray!15}}c}{\modelname{GPT-4.1-mini}} \\
    \methodrule
    \textit{Full-context} & N & 80.60 & 53.30 & 45.00 \\
    \methodrule
    LangMem & N & 73.40 & 47.60 & 40.00 \\
    Zep~\citep{rasmussen2025zep} & N & 61.60 & 36.90 & 30.90 \\
    Mem0~\citep{chhikara2025mem0} & N & 66.30 & 43.50 & 36.50 \\
    Nemori~\citep{nan2025nemori} & N & \second{79.40} & \best{53.40} & \best{45.60} \\
    \methodrule
    RAG & N & 32.90 & 23.50 & 19.20 \\
    \cellcolor{rawraghighlight}\textbf{\rawrag{}} (minimal) & L & \best{90.87} & \second{49.57} & \second{41.35} \\
    \bottomrule
  \end{tabular*}
\end{table}
\begin{table}[!htbp]
  \caption{LoCoMo short-answer scores by question type with GPT-4o-mini
  (0--100). Metrics are LLM-judge accuracy ($J$), token F1, and unigram
  BLEU ($B_1$). Single, Multi, and Open mean single-hop, multi-hop, and
  open-domain. Src.\ identifies the source: M, Mem0~\citep{chhikara2025mem0} paper; N, Nemori~\citep{nan2025nemori} paper;
  L, this study. Mem0$^g$~\citep{chhikara2025mem0} is the graph variant; ``minimal'' requests the
  shortest sufficient answer. Bold marks column maxima excluding
  Full-context.}
  \label{tab:locomo-mem0-categories}
  \centering\footnotesize
  \setlength{\tabcolsep}{2pt}
  \renewcommand{\arraystretch}{1.12}
  \begin{tabular*}{\linewidth}{@{\extracolsep{\fill}}ll*{4}{ccc}@{}}
    \toprule
    \multirow{2}{*}[-\dimexpr(\aboverulesep+\belowrulesep+\cmidrulewidth)/2\relax]{Method}
      & \multirow{2}{*}[-\dimexpr(\aboverulesep+\belowrulesep+\cmidrulewidth)/2\relax]{Src.}
      & \multicolumn{3}{c}{Single} & \multicolumn{3}{c}{Multi}
      & \multicolumn{3}{c}{Temporal} & \multicolumn{3}{c}{Open} \\
    \cmidrule(lr){3-5}\cmidrule(lr){6-8}\cmidrule(lr){9-11}\cmidrule(lr){12-14}
      & & $J$ & F1 & $B_1$ & $J$ & F1 & $B_1$
      & $J$ & F1 & $B_1$ & $J$ & F1 & $B_1$ \\
    \midrule
    \multicolumn{14}{>{\columncolor{gray!15}}c}{\modelname{GPT-4o-mini}} \\
    \methodrule
    A-MEM~\citep{xu2025amem} (rerun) & M & 39.79 & 20.76 & 14.90 & 18.85 & 9.22 & 8.81 & 49.91 & 35.40 & 31.08 & 54.05 & 33.34 & 27.58 \\
    LangMem & M & 62.23 & 35.51 & 26.86 & 47.92 & 26.04 & 22.32 & 23.43 & 30.75 & 25.84 & 71.12 & 40.91 & 33.63 \\
    Zep~\citep{rasmussen2025zep} & M & 61.70 & 35.74 & 23.30 & 41.35 & 19.37 & 14.82 & 49.31 & 42.00 & 34.53 & \best{76.60} & \best{49.56} & \second{38.92} \\
    OpenAI & M & 63.79 & 34.30 & 23.72 & 42.92 & 20.09 & 15.42 & 21.71 & 14.04 & 11.25 & 62.29 & 39.31 & 31.16 \\
    Mem0~\citep{chhikara2025mem0} & M & 67.13 & 38.72 & 27.13 & 51.15 & 28.64 & 21.58 & 55.51 & 48.93 & 40.51 & 72.93 & 47.65 & 38.72 \\
    Mem0$^{g}$~\citep{chhikara2025mem0} & M & 65.71 & 38.09 & 26.03 & 47.19 & 24.32 & 18.82 & 58.13 & 51.55 & 40.28 & \second{75.71} & \second{49.27} & \best{40.30} \\
    \midrule
    \textit{Full-context} & N & 83.00 & 53.10 & 44.70 & 66.80 & 35.40 & 26.10 & 56.20 & 44.10 & 36.10 & 48.60 & 24.50 & 17.20 \\
    \methodrule
    LangMem & N & 61.40 & 38.80 & 33.10 & 52.40 & 33.50 & 23.90 & 24.90 & 31.90 & 26.20 & 47.60 & 29.40 & 23.50 \\
    Zep~\citep{rasmussen2025zep} & N & 63.20 & 39.70 & 33.70 & 50.50 & 27.50 & 19.30 & 58.90 & 44.80 & 38.10 & 39.60 & 22.90 & 15.70 \\
    Mem0~\citep{chhikara2025mem0} & N & 68.10 & 44.40 & 37.70 & 60.30 & 34.30 & 25.20 & 50.40 & 44.40 & 37.60 & 40.60 & 27.10 & 19.40 \\
    Nemori~\citep{nan2025nemori} & N & \second{82.10} & \second{54.40} & \second{43.20} & \second{65.30} & \second{36.50} & \second{25.60} & \second{71.00} & \second{56.70} & \second{46.60} & 44.80 & 20.80 & 15.10 \\
    \methodrule
    RAG & N & 32.00 & 22.20 & 18.60 & 31.30 & 18.60 & 11.70 & 23.70 & 19.50 & 15.70 & 32.60 & 19.00 & 13.50 \\
    \cellcolor{rawraghighlight}\textbf{\rawrag{}} (minimal) & L & \best{91.84} & \best{55.47} & \best{46.71} & \best{84.40} & \best{44.69} & \best{37.36} & \best{80.06} & \best{61.53} & \best{52.88} & 55.90 & 20.60 & 18.10 \\
    \bottomrule
  \end{tabular*}
\end{table}

\begin{table}[!htbp]
  \caption{LoCoMo short-answer scores by question type with GPT-4.1-mini
  (0--100). Metrics are LLM-judge accuracy ($J$), token F1, and unigram
  BLEU ($B_1$). Single, Multi, and Open mean single-hop, multi-hop, and
  open-domain. Baseline scores are from the Nemori~\citep{nan2025nemori} paper; \rawrag{}
  uses the shortest-sufficient-answer instruction. Bold marks column
  maxima excluding Full-context.}
  \label{tab:locomo-nemori-categories}
  \centering\footnotesize
  \setlength{\tabcolsep}{2pt}
  \renewcommand{\arraystretch}{1.12}
  \begin{tabular*}{\linewidth}{@{\extracolsep{\fill}}l*{4}{ccc}@{}}
    \toprule
    \multirow{2}{*}[-\dimexpr(\aboverulesep+\belowrulesep+\cmidrulewidth)/2\relax]{Method}
      & \multicolumn{3}{c}{Single} & \multicolumn{3}{c}{Multi}
      & \multicolumn{3}{c}{Temporal} & \multicolumn{3}{c}{Open} \\
    \cmidrule(lr){2-4}\cmidrule(lr){5-7}\cmidrule(lr){8-10}\cmidrule(lr){11-13}
      & $J$ & F1 & $B_1$ & $J$ & F1 & $B_1$
      & $J$ & F1 & $B_1$ & $J$ & F1 & $B_1$ \\
    \midrule
    \multicolumn{13}{>{\columncolor{gray!15}}c}{\modelname{GPT-4.1-mini}} \\
    \methodrule
    \textit{Full-context} & 86.90 & 61.40 & 53.40 & 77.20 & 44.20 & 33.70 & 74.20 & 47.50 & 40.00 & 56.60 & 28.40 & 22.20 \\
    \methodrule
    LangMem & 84.50 & 51.00 & 43.60 & 71.00 & 41.50 & \second{32.50} & 50.80 & \second{48.50} & \second{40.90} & \second{59.00} & \best{32.80} & \best{26.40} \\
    Zep~\citep{rasmussen2025zep} & 66.90 & 45.50 & 40.00 & 53.70 & 30.50 & 20.40 & 60.20 & 23.90 & 20.00 & 43.80 & 24.20 & 19.30 \\
    Mem0~\citep{chhikara2025mem0} & 71.40 & 48.60 & 42.00 & 68.20 & 40.10 & 30.30 & 56.90 & 39.20 & 33.20 & 47.90 & 23.70 & 17.70 \\
    Nemori~\citep{nan2025nemori} & \second{84.90} & \best{58.80} & \best{51.50} & \second{75.10} & \second{41.70} & 31.90 & \second{77.60} & \best{57.70} & \best{50.20} & 51.00 & 25.80 & 19.30 \\
    \methodrule
    RAG & 35.90 & 25.80 & 22.00 & 31.70 & 20.10 & 12.80 & 27.40 & 22.30 & 19.10 & 28.80 & 17.90 & 13.90 \\
    \cellcolor{rawraghighlight}\textbf{\rawrag{}} (minimal) & \best{95.80} & \second{56.21} & \second{46.66} & \best{87.12} & \best{42.61} & \best{35.65} & \best{87.12} & 44.33 & 37.88 & \best{71.18} & \second{29.43} & \second{23.21} \\
    \bottomrule
  \end{tabular*}
\end{table}
\subsection{LongMemEval-S results}
\label{app:longmemeval-results}

We evaluate \rawrag{} under the corresponding matched LongMemEval-S
settings with GPT-4.1-mini as the reader. The two full-set retrieval
settings use all 500 questions; the non-abstention setting uses the
470 answerable questions. Overall accuracy is question-weighted.

\subsubsection{Text-embedding-3-small setting}
\label{app:longmemeval-openai}

This setting uses \texttt{text-embedding-3-small} and includes all
500 questions, including abstention items. The category counts in
S-U./S-A./S-P./M-S./KU./TR. order are 70/56/30/133/78/133.
\rawrag{} uses Cohere Rerank-v3.5 and follows the corresponding
history-access, answer-generation, evidence-budget, and scoring settings.
Each fixed answer receives three GPT-4o-mini judgments, averaged
over questions and repetitions.
Table~\ref{tab:longmemeval-openai-protocol} reports MemoryOS, A-MEM,
RecMem, Nemori, MemMachine, Memora, and \rawrag{} in this setting.
Published results are from \citet{dai2026recmem,nan2025nemori,xia2026memora};
Memora uses its published policy retriever.
Full-context is omitted because it performs no embedding-based retrieval.
EMem and EMem-G belong to the 470-question setting in
Appendix~\ref{app:longmemeval-no-abs}.

\begin{table}[!htbp]
  \caption{LongMemEval-S accuracy (\%) with GPT-4.1-mini under the
  text-embedding-3-small evaluation setting, using all 500 questions.
  S-U., S-A., and S-P.\ denote single-session user, assistant, and
  preference questions; M-S., multi-session; KU., knowledge update;
  TR., temporal reasoning. Overall is question-weighted.
  Bold marks column maxima.}
  \label{tab:longmemeval-openai-protocol}
  \centering\small
  \setlength{\tabcolsep}{3pt}
  \begin{tabular*}{\linewidth}{@{\extracolsep{\fill}}lcccccc>{\columncolor{yellow!15}[\tabcolsep][0pt]}c@{}}
    \toprule
    Method & S-U. & S-A. & S-P. & M-S. & KU. & TR. & Overall \\
    \midrule
    \multicolumn{8}{>{\columncolor{gray!15}}c}{\modelname{GPT-4.1-mini}} \\
    \methodrule
    MemoryOS~\citep{kang2025memoryos} & 94.29 & 89.29 & \best{100.00} & 67.67 & 80.77 & 54.89 & 74.40 \\
    A-MEM~\citep{xu2025amem} & 95.71 & \best{100.00} & 63.33 & 61.65 & 82.05 & 52.63 & 71.60 \\
    RecMem~\citep{dai2026recmem} & 95.71 & 96.43 & 86.67 & 65.41 & 75.64 & 68.42 & 76.80 \\
    Nemori~\citep{nan2025nemori} & 90.00 & 92.90 & \second{86.70} & 55.60 & 79.50 & 72.20 & 74.60 \\
    MemMachine~\citep{wang2026memmachine} & 92.86 & \best{100.00} & 55.56 & 66.67 & 86.32 & 58.65 & 74.33 \\
    Memora~\citep{xia2026memora} & \second{98.60} & 78.60 & 83.30 & \best{78.20} & \best{97.40} & \best{89.50} & \second{87.40} \\
    \methodrule
    \cellcolor{rawraghighlight}\textbf{\rawrag{}} & \best{100.00} & \second{98.21} & \best{100.00} & \second{77.69} & \second{91.03} & \second{81.45} & \best{87.53} \\
    \bottomrule
  \end{tabular*}
\end{table}
\subsubsection{Qwen3 retrieval setting}
\label{app:longmemeval-qwen}

This setting uses Qwen3-Embedding-4B and Qwen3-Reranker-4B over all
500 questions. \rawrag{} follows the EverMemOS evaluation
setting~\citep{hu2026evermemos}, including history access, answer generation, and the repeated-judge scoring procedure.
As on LoCoMo, the shared grading prompts are evaluated with three
GPT-4o-mini judgments per fixed answer, and overall accuracy averages
correctness over questions and repetitions.
Table~\ref{tab:longmemeval-qwen-protocol} reports the corresponding
results, retaining the published EverMemOS baseline set.

\begin{table}[!htbp]
  \caption{LongMemEval-S LLM-judge accuracy (\%) with GPT-4.1-mini,
  Qwen3-Embedding-4B, and Qwen3-Reranker-4B, using all 500 questions.
  S-U., S-A., and S-P.\ denote single-session user, assistant, and
  preference questions; M-S., multi-session; KU., knowledge update;
  TR., temporal reasoning. Overall is question-weighted.
  Bold marks column maxima.}
  \label{tab:longmemeval-qwen-protocol}
  \centering\small
  \setlength{\tabcolsep}{3pt}
  \begin{tabular*}{\linewidth}{@{\extracolsep{\fill}}lcccccc>{\columncolor{yellow!15}[\tabcolsep][0pt]}c@{}}
    \toprule
    Method & S-U. & S-A. & S-P. & M-S. & KU. & TR. & Overall \\
    \midrule
    \multicolumn{8}{>{\columncolor{gray!15}}c}{\modelname{GPT-4.1-mini}} \\
    \methodrule
    MemU~\citep{nevamind2025memu} & 67.14 & 19.64 & 76.67 & 42.10 & 41.02 & 17.29 & 38.40 \\
    Zep~\citep{rasmussen2025zep} & 92.90 & 75.00 & 53.30 & 47.40 & 74.40 & 54.10 & 63.80 \\
    Mem0~\citep{chhikara2025mem0} & 82.86 & 26.78 & 90.00 & 63.15 & 66.67 & 72.18 & 66.40 \\
    MemOS~\citep{li2025memos} & 95.71 & 67.86 & \best{96.67} & 70.67 & 74.26 & \second{77.44} & 77.80 \\
    EverMemOS~\citep{hu2026evermemos} & \second{97.14} & \second{85.71} & 93.33 & \second{73.68} & \second{89.74} & \second{77.44} & \second{83.00} \\
    \methodrule
    \cellcolor{rawraghighlight}\textbf{\rawrag{}} & \best{100.00} & \best{96.43} & \second{95.56} & \best{79.20} & \best{92.31} & \best{89.97} & \best{89.93} \\
    \bottomrule
  \end{tabular*}
\end{table}
\subsubsection{Non-abstention evaluation}
\label{app:longmemeval-no-abs}

This setting follows EMem and its graph variant
EMem-G~\citep{zhou2025simple}: it contains 470 non-abstention questions,
with category counts 64/56/30/121/72/127 in the
S-U./S-A./S-P./M-S./KU./TR. order. It removes the 30 annotated
abstention items from the full set: six S-U., twelve M-S.,
six KU., and six TR. questions.
The \rawrag{} comparison uses the corresponding matched evaluation
conditions, including answer generation, judge prompts,
and three GPT-4o-mini judge repetitions.
The textemb3 configuration pairs \texttt{text-embedding-3-small} with Cohere
Rerank-v3.5; Qwen3 pairs Qwen3-Embedding-4B with Qwen3-Reranker-4B.
We reuse the existing GPT-4.1-mini answers and their three binary
GPT-4o-mini judgments, remove only the 30 annotated abstention items,
and recompute question-weighted accuracy. For the remaining question set
$Q_{\mathrm{ans}}$ and correctness judgment $y_{ir}\in\{0,1\}$,
\begin{equation}
  \label{eq:longmemeval-non-abstention-accuracy}
  \mathrm{Acc}_{\mathrm{ans}}
  = \frac{100}{3|Q_{\mathrm{ans}}|}
    \sum_{i\in Q_{\mathrm{ans}}}\sum_{r=1}^{3}y_{ir},
  \qquad |Q_{\mathrm{ans}}|=470.
\end{equation}
This recalculation reuses the existing answers and judgments.
The textemb3 and Qwen3 \rawrag{} configurations yield respectively
1,238/1,410 and 1,277/1,410 correct judgments, or 87.80 and 90.57.
Their full-set scores remain 87.53 and 89.93.
Table~\ref{tab:longmemeval-no-abs} reports the matched non-abstention
comparison. The textemb3-configured \rawrag{} exceeds EMem and EMem-G by
4.80 and 2.90 points; the Qwen3-configured \rawrag{} exceeds them by
7.57 and 5.67 points. The separate 5,000-token textemb3-configured control
in Appendix~\ref{app:reader-context-budget} scores 90.14
(1,271/1,410) on this subset, compared with its full-set 89.20.

\begin{table}[!htbp]
  \caption{LongMemEval-S accuracy (\%) with GPT-4.1-mini on the 470
  non-abstention questions. The textemb3 setting uses
  \texttt{text-embedding-3-small}; Qwen3 uses Qwen3-Embedding-4B.
  EMem-G~\citep{zhou2025simple} is EMem~\citep{zhou2025simple}'s graph variant. S-U., S-A., and S-P.\ denote
  single-session user, assistant, and preference questions; M-S.,
  multi-session; KU., knowledge update; TR., temporal reasoning.
  Overall is question-weighted. Bold marks column maxima.}
  \label{tab:longmemeval-no-abs}
  \centering\small
  \setlength{\tabcolsep}{3pt}
  \begin{tabular*}{\linewidth}{@{\extracolsep{\fill}}lcccccc>{\columncolor{yellow!15}[\tabcolsep][0pt]}c@{}}
    \toprule
    \multicolumn{1}{c}{Method} & S-U. & S-A. & S-P. & M-S. & KU. & TR. & Overall \\
    \midrule
    \multicolumn{8}{>{\columncolor{gray!15}}c}{\modelname{GPT-4.1-mini}} \\
    \methodrule
    EMem~\citep{zhou2025simple} & 93.80 & 82.10 & 46.70 & \second{82.10} & \best{95.40} & 80.60 & 83.00 \\
    EMem-G~\citep{zhou2025simple} & \second{94.80} & 87.50 & 50.00 & \best{82.60} & 94.40 & \second{83.70} & 84.90 \\
    \methodrule
    \cellcolor{rawraghighlight}\textbf{\rawrag{}} (textemb3) & \best{100.00} & \best{98.21} & \best{100.00} & 76.31 & \second{94.44} & 81.36 & \second{87.80} \\
    \cellcolor{rawraghighlight}\textbf{\rawrag{}} (Qwen3) & \best{100.00} & \second{96.43} & \second{95.56} & 78.79 & \second{94.44} & \best{91.08} & \best{90.57} \\
    \bottomrule
  \end{tabular*}
\end{table}

\subsection{PersonaMem-v2 results}
\label{app:personamem-v2}

Table~\ref{tab:personamem-v2} extends the evaluation to
PersonaMem-v2~\citep{jiang2025personamemv2},
covering nine conversational scenarios. The \rawrag{} run uses
GPT-4.1-mini as the answer model and is evaluated by multiple-choice
accuracy on 5,000 questions. We report scenario-level percentages;
its overall score is the question-weighted
accuracy.

\paragraph{Experimental configuration.}
We fully align the evaluation design and retrieval backbones with the
EverMemOS comparison~\citep{hu2026evermemos}: the PersonaMem-v2 text
benchmark with 32k histories, the 5,000-question multiple-choice task,
GPT-4.1-mini as the reader, and exact-match answer-letter accuracy.
Specifically, \rawrag{} uses Qwen3-Embedding-4B for dense retrieval and
Qwen3-Reranker-4B for reranking~\citep{zhang2025qwen3embedding}.
It indexes 256-token windows with
32-token overlap, retrieves 128 candidates, and selects the top 15
reranked windows, expanding them to complete source turns for the reader.
The reader uses temperature 0 and a maximum of 16,384 output tokens.
\rawrag{} uses no persona profile; EverMemOS uses its profile-enabled
memory-management pipeline.
Appendix~\ref{app:query-construction} specifies the retrieval query;
Appendix~\ref{app:prompt-personamem-reader} gives the exact reader input.

\begin{table}[!htbp]
  \caption{PersonaMem-v2 multiple-choice accuracy (\%) with GPT-4.1-mini
  and 32k-token histories. \rawrag{}'s Overall score is question-weighted
  over 5,000 questions.
  Subset abbreviations: CON = Consultation; EP = Email (Personal);
  TR = Translation; EPR = Email (Professional); CW = Creative Writing;
  WP = Writing (Professional); KQ = Knowledge Query; SM = Social Media;
  CH = Chat. Profile indicates use of persona profiles. Bold marks the highest
  accuracy in each scenario and overall.}
  \label{tab:personamem-v2}
  \centering\footnotesize
  \setlength{\tabcolsep}{3pt}
  \renewcommand{\arraystretch}{1.1}
  \resizebox{\linewidth}{!}{%
  \begin{tabular}{l*{11}{c}}
    \toprule
    Method & Profile & CON & EP & TR & EPR & CW & WP & KQ & SM & CH & Overall \\
    \midrule
    \multicolumn{12}{>{\columncolor{gray!15}}c}{\modelname{GPT-4.1-mini}} \\
    \methodrule
    Zep~\citep{rasmussen2025zep} & No
      & 39.51 & 42.51 & 36.92 & 37.59 & 41.22 & 40.54 & \second{63.43} & 32.35 & 44.87 & 43.40 \\
    Mem0~\citep{chhikara2025mem0} & No
      & 43.21 & 41.30 & 43.08 & 42.41 & 42.86 & 34.75 & 59.20 & 38.66 & 40.30 & 43.85 \\
    MemU~\citep{nevamind2025memu} & Yes
      & 37.86 & 33.20 & 38.46 & 32.76 & 35.51 & 35.14 & 56.97 & 34.03 & 34.22 & 38.70 \\
    MemoryOS~\citep{kang2025memoryos} & Yes
      & 35.80 & 36.84 & 40.00 & 35.86 & 35.51 & 35.91 & 57.96 & 35.29 & 36.88 & 40.05 \\
    MemOS~\citep{li2025memos} & Yes
      & \second{48.15} & 49.80 & \best{51.92} & \second{50.00} & 48.16 & \best{48.26} & 61.94 & 46.64 & 44.87 & \second{50.72} \\
    EverMemOS~\citep{hu2026evermemos} & Yes
      & \best{51.03} & \best{53.85} & 50.00 & \best{53.79} & \best{55.10} & 45.56 & \best{63.68} & \best{47.90} & \best{52.09} & \best{53.25} \\
    \cellcolor{rawraghighlight}\textbf{\rawrag{}} & No
      & 46.51 & \second{50.20} & \second{50.09} & 46.52 & \second{48.80} & \second{48.25} & 57.89 & \second{47.84} & \second{45.58} & 49.62 \\
    \bottomrule
  \end{tabular}}
\end{table}
\paragraph{Comparative performance.}
\rawrag{} achieves 49.62\% accuracy (2,481/5,000), ranking third among
seven systems behind MemOS and EverMemOS, without leading any individual
scenario. The comparison measures overall system performance with each
system's native profile configuration; statistical significance remains
untested.

\subsection{Ablation studies}
\label{app:additional-ablations}
\label{app:reader-context-budget}

This section reports retrieval-design and memory-management ablations,
followed by controls on reader-visible evidence. Table~\ref{tab:ablation}
compares chunking and reranking choices with three added memory-management
operations.
We first give the full final-$k$ sweep, then measure actual reader context
and report two distinct 5k controls: a six-system MemoryAgentBench
intervention and \rawrag{}-only interventions on the other three benchmarks.

\subsubsection{Retrieval and memory ablations}

The retrieval-depth sweep uses the same MemoryAgentBench setting and
official capability-level averaging as Table~\ref{tab:ablation}.
Its paired intervals compare each final-$k$ condition with the default
$k=15$; statistical procedures are specified in
Appendix~\ref{app:statistical-methods}.
Totals and differences are computed from unrounded scores and rounded
independently; a displayed difference can therefore differ by 0.01 points
from subtraction of the displayed totals.

\begin{table}[!htbp]
  \caption{Effect of retrieval depth on \rawrag{} with GPT-4o-mini on
  MemoryAgentBench. Final $k$ is the number of reranked chunks selected for
  the reader; Total (\%) averages the four capability scores. Gray marks
  the default $k=15$. $\Delta$ and its paired-bootstrap 95\% confidence
  interval (CI) compare each setting with the default, in percentage points.}
  \label{tab:topk}
  \centering\normalsize
  \setlength{\tabcolsep}{7pt}
  {\setlength{\aboverulesep}{0pt}\setlength{\belowrulesep}{0pt}%
  \renewcommand{\arraystretch}{1.15}%
  \begin{tabular}{cccc}
    \toprule
    Final $k$ & Total & $\Delta$ & 95\% CI \\
    \midrule
    \multicolumn{4}{>{\columncolor{gray!15}}c}{\modelname{GPT-4o-mini}} \\
    \methodrule
    1   & 36.59 & -11.73 & [-14.05,-9.43] \\
    5   & 47.10 & -1.21  & [-2.81,+0.40] \\
    10  & 47.57 & -0.75  & [-2.19,+0.66] \\
    \rowcolor{gray!15}
    15  & \best{48.32} & -- & -- \\
    30  & \second{48.10} & -0.22  & [-1.40,+0.93] \\
    60  & 47.91 & -0.41  & [-1.80,+0.94] \\
    120 & 33.74 & -14.57 & [-16.83,-12.34] \\
    \bottomrule
  \end{tabular}}
\end{table}

Although top-10 and top-60 do not significantly change Total after Holm
correction, top-60 improves Long-Range Understanding by 2.22 points with a
Holm-adjusted $p=0.025$ within that task family, illustrating that one global
context size need not be optimal for every capability.

The memory-management additions in Table~\ref{tab:ablation} use the stated
retrieval configuration and change the named representation or expansion
operation. Evidence length varies across these conditions. Their computation
and latency results are reported in Appendix~\ref{app:efficiency-results}.

\subsubsection{Evidence-length analysis}
\label{app:reader-evidence-lengths}

We measure the evidence actually supplied to the main \rawrag{} readers
before comparing capped conditions. The selected top-$k$ windows can vary
in token count after expansion and formatting.
Native overflow and truncation policies are specified in
Appendix~\ref{app:native-context-policy}.

\paragraph{Measurement scope.}
Table~\ref{tab:raw-reader-lengths} measures the evidence text saved with
the answers underlying the main \rawrag{} results, after display-time
expansion and formatting. Counts include memory headers, timestamps,
separators, and repeated evidence, but exclude the question, fixed
prompt, chat-message framing, and generated answer. For comparability
we count all rows with \texttt{tiktoken}'s \texttt{o200k\_base}
encoding across readers. Index construction uses the GPT-4 encoding for its
256-token windows. Length statistics weight questions equally.
Percentiles use linear interpolation.
LoCoMo excludes adversarial questions. Evidence-length distributions
are identical across the two readers on LoCoMo, MEME, and MemoryAgentBench;
only GPT-4.1-mini is used on LongMemEval.

\begin{table}[!htbp]
  \centering\small
  \caption{Length of the evidence supplied to \rawrag{} readers, in tokens,
  including formatting but excluding the question and fixed prompt.
  $N$ is the number of questions per reader; P95 is the 95th percentile;
  $>5{,}000$ counts questions exceeding 5,000 evidence tokens.}
  \label{tab:raw-reader-lengths}
  \setlength{\tabcolsep}{5pt}
  \begin{tabular}{lcccccc}
    \toprule
    Benchmark & $N$ & Mean & Median & P95 & Maximum & $>5{,}000$ \\
    \midrule
    LoCoMo & 1,540 & 4,778.8 & 4,778.0 & 4,966.1 & 5,177 & 41 \\
    LongMemEval & 500 & 11,355.6 & 11,248.5 & 13,915.3 & 27,375 & 500 \\
    MEME & 1,188 & 11,986.1 & 10,186.0 & 21,329.6 & 42,490 & 1,188 \\
    MemoryAgentBench & 2,071 & 3,915.8 & 4,173.0 & 4,209.5 & 4,262 & 0 \\
    \bottomrule
  \end{tabular}
\end{table}

\paragraph{Expansion overhead.}
The mean lengths of the selected chunk bodies before display
expansion and headers are 3,734.9, 3,784.9, and 3,813.3 tokens,
on LoCoMo, LongMemEval, and MEME, respectively. The larger means in
Table~\ref{tab:raw-reader-lengths} reflect expansion and formatting
of those selected chunks.
MemoryAgentBench has a mean selected-body length of 3,550.5 tokens.
Evidence-assembly rules are defined in Appendix~\ref{app:method}.

\subsubsection{Cross-system budget control}
\label{app:mab-5k-results}

This intervention freezes each system's retrieved evidence and imposes a
5,000-token ceiling on formatted reader evidence, excluding the outer
prompt and question. It keeps a prefix of complete entries, with
token-prefix clipping only when the first entry alone is too long.
The full rule is defined in Appendix~\ref{app:budget-interventions}.
The reader receives at most 5,000 evidence tokens per question.

The completed control covers six systems and all 14 sources
(2,071 questions per system). Mem0 and MemOS use earlier hosted-service
configurations, rather than their locally hosted counterparts in the main
comparison. Of 84 system--source conditions,
44 required new answers and evaluation after clipping; the remaining
40 reused their unchanged answers and judgments. The rerun reader is
GPT-4o-mini-2024-07-18 and the rerun judge is GPT-4o-2024-11-20
where the source metric uses an LLM judge; reused evaluations record
the GPT-4o alias. The historical clipping log records 5,545 changed
system--question contexts, including 1,479 first-entry cuts.
Scores follow the official weighting scheme: TTL averages the five-source
ICL macro with ReDial, and Overall averages the four capabilities.

\begin{table}[!htbp]
  \centering\small
  \caption{Reader-evidence lengths before and after applying a 5,000-token
  cap on MemoryAgentBench (2,071 questions per system). Native mean,
  Median, P95 (95th percentile), and Maximum describe evidence before
  capping; 5k mean describes capped evidence. Cut counts questions whose
  evidence changes. Mem0~\citep{chhikara2025mem0} uses the earlier hosted-service configuration.}
  \label{tab:mab-5k-lengths}
  \setlength{\tabcolsep}{4pt}
  \begin{tabular}{lcccccc}
    \toprule
    System & Native mean & Median & P95 & Maximum & 5k mean & Cut \\
    \midrule
    EMem~\citep{zhou2025simple} & 2,637.7 & 171 & 7,081.0 & 125,646 & 1,417.7 & 511 \\
    Mem0~\citep{chhikara2025mem0} & 3,572.8 & 1,033 & 6,319.0 & 39,485 & 1,961.3 & 500 \\
    EverMemOS~\citep{hu2026evermemos} & 15,394.8 & 9,443 & 50,750.0 & 70,861 & 4,160.6 & 2,011 \\
    MemMachine~\citep{wang2026memmachine} & 26,296.9 & 21,806 & 82,868.5 & 88,967 & 4,479.2 & 2,030 \\
    \methodrule
    \cellcolor{rawraghighlight}\textbf{\rawrag{}} & 3,915.8 & 4,173 & 4,209.5 & 4,262 & 3,915.8 & 0 \\
    \bottomrule
  \end{tabular}
\end{table}

Table~\ref{tab:mab-5k-lengths} summarizes the distribution of uncapped
reader-evidence lengths and the effect of the 5,000-token ceiling.
\rawrag{} stays below the ceiling for every question, so its mean evidence
length remains 3,915.8 tokens. EMem and Mem0 have lower median lengths
but long upper tails: their 95th percentiles exceed 5,000 tokens, and
clipping affects 511 and 500 questions, respectively. Their mean lengths
fall to 1,417.7 and 1,961.3 tokens under the ceiling.
EverMemOS and MemMachine supply substantially longer uncapped evidence,
with means of 15,394.8 and 26,296.9 tokens. Clipping affects nearly all
their questions (2,011 and 2,030 of 2,071), reducing their means to
4,160.6 and 4,479.2 tokens. Thus, the same ceiling leaves \rawrag{}
unchanged, shortens the upper tails of EMem and Mem0, and substantially
reduces evidence length for most EverMemOS and MemMachine questions.

\begin{table}[!htbp]
  \centering\small
  \caption{MemoryAgentBench scores (\%) with GPT-4o-mini and a 5,000-token
  evidence cap. AR denotes Accurate Retrieval; TTL, Test-Time Learning;
  LRU, Long-Range Understanding; CR, Conflict Resolution.
  Overall averages the four capabilities; Native is the corresponding
  score before capping, and $\Delta$ is 5k minus Native in percentage
  points. Mem0~\citep{chhikara2025mem0} and MemOS~\citep{li2025memos} use earlier hosted-service configurations.
  Bold marks the highest capped score in each column.}
  \label{tab:mab-5k-results}
  \setlength{\tabcolsep}{4pt}
  \begin{tabular}{lccccccc}
    \toprule
    System & AR & TTL & LRU & CR & Overall (5k) & Native & $\Delta$ \\
    \midrule
    \multicolumn{8}{>{\columncolor{gray!15}}c}{\modelname{GPT-4o-mini}} \\
    \methodrule
    EMem~\citep{zhou2025simple} & 57.60 & 45.87 & 16.94 & 30.00 & 37.60 & 38.44 & $-0.84$ \\
    Mem0~\citep{chhikara2025mem0} & \second{65.10} & \second{46.25} & 21.13 & \best{46.50}
      & \second{44.74} & 45.27 & $-0.53$ \\
    EverMemOS~\citep{hu2026evermemos} & 42.42 & 41.38 & 22.62 & 28.50
      & 33.73 & 36.49 & $-2.76$ \\
    MemMachine~\citep{wang2026memmachine} & 53.55 & 23.39 & \second{26.13} & 23.50
      & 31.64 & 35.53 & $-3.89$ \\
    MemOS~\citep{li2025memos} & 52.72 & 46.06 & 21.83 & 15.00
      & 33.90 & 34.18 & $-0.28$ \\
    \methodrule
    \cellcolor{rawraghighlight}\textbf{\rawrag{}} & \best{67.85} & \best{47.68} & \best{31.74} & \second{46.00}
      & \best{48.32} & 48.32 & 0.00 \\
    \bottomrule
  \end{tabular}
\end{table}

\paragraph{Interpretation.}
Table~\ref{tab:mab-5k-results} reports the completed control.
\rawrag{} requires no clipping on any MemoryAgentBench source:
its largest evidence input is 4,262 tokens and its Overall remains
48.32\%, versus 44.74\% for the strongest baseline in this control.
Overall scores for all five baselines decrease under the historical
5k intervention, although individual sources can improve.
\rawrag{} thus remains ahead under the 5k ceiling. Clipping can remove
useful items or disrupt their format, and all systems use their original
tuning. This control covers the six listed systems with GPT-4o-mini.
The controls below extend the evidence ceiling to \rawrag{} on the other
three benchmarks.

\subsubsection{Raw-RAG budget sensitivity}
\label{app:rag-5k-results}

These controls use the main GPT-4o-mini LoCoMo and MEME conditions
and the GPT-4.1-mini LongMemEval condition. Selected top-15 chunks and
their ordering are frozen. For over-budget questions, turn expansion is
disabled before any clipping is considered; the complete policy is
specified in Appendix~\ref{app:budget-interventions}.
No other baseline is rerun under this policy.

In the completed experiments, disabling expansion alone suffices
for every over-budget question: all 15 selected chunks remain, with
no additional chunk dropped or clipped. Contexts change for 41 of
1,540 LoCoMo questions and all 500 LongMemEval and 1,188 MEME
questions. All 3,228 answers and their evaluations are rerun,
including LoCoMo's 1,499 unchanged-context questions. The reruns use
the same reader aliases and benchmark prompts as their native
conditions. Each answer is evaluated three times, using
GPT-4o-mini-2024-07-18 for LoCoMo, the GPT-4o-mini alias for
LongMemEval, and GPT-4o-2024-11-20 for MEME. LoCoMo and LongMemEval
average binary judgments across questions and repetitions; MEME
uses its six-task macro-average with the original before/after filtering.
Evidence-token counts include formatting but exclude the outer prompt
and question.

\begin{table}[!htbp]
  \centering\small
  \caption{\rawrag{} results with a 5,000-token evidence cap, using
  GPT-4o-mini on LoCoMo and MEME and GPT-4.1-mini on LongMemEval.
  $N$ counts questions; Changed counts contexts shortened by disabling
  turn expansion, with all selected chunks retained.
  Scores are percentages; $\Delta$ is 5k minus Native in percentage
  points. Bold marks the higher score for each benchmark.}
  \label{tab:rag-other-5k-results}
  \setlength{\tabcolsep}{3pt}
  \begin{tabular}{lccccccc}
    \toprule
    & & & \multicolumn{2}{c}{5k evidence tokens}
      & \multicolumn{3}{c}{Score} \\
    \cmidrule(lr){4-5}\cmidrule(lr){6-8}
    Benchmark & $N$ & Changed & Mean & Maximum & Native & 5k & $\Delta$ \\
    \midrule
    LoCoMo & 1,540 & 41 & 4,753.5 & 4,999 & \second{89.07} & \best{89.42} & $+0.35$ \\
    LongMemEval & 500 & 500 & 4,144.4 & 4,271 & \second{87.53} & \best{89.20} & $+1.67$ \\
    MEME & 1,188 & 1,188 & 4,173.1 & 4,273 & \best{39.30} & \second{38.21} & $-1.09$ \\
    \bottomrule
  \end{tabular}
\end{table}

\paragraph{Findings and limitations.}
Table~\ref{tab:rag-other-5k-results} shows higher rerun scores on
LoCoMo and LongMemEval and a 1.09-point decrease on MEME.
Relative to Table~\ref{tab:raw-reader-lengths}, mean evidence length
falls by 63.5\% on LongMemEval and 65.2\% on MEME.
Category-level changes are reported in Table~\ref{tab:rag-5k-task-results}.
Direct verification of all rerun reader inputs confirms agreement
with the selected display policy, the 5,000-token ceiling, and no
recorded answer-generation error responses.

Disabling turn expansion shortens the evidence without dropping or clipping
any selected chunk. All answers and judgments are regenerated, so the score
changes also include reader and judge variability. On LoCoMo, the 41
modified contexts contribute $+0.065$ points, while the 1,499 unchanged
contexts contribute $+0.281$ points. Most of its observed gain therefore
comes from questions whose evidence did not change. These RAG-only reruns
describe performance under the smaller budget; significance and equivalence
have not been tested. Comparisons with other systems use their original
scores and evidence budgets.

\paragraph{Baseline comparison.}
Table~\ref{tab:rag-5k-task-results} gives the native and capped
\rawrag{} category scores; baseline scores remain in
Tables~\ref{tab:main} and~\ref{tab:main-other}.
The 5k Overall scores exceed the strongest non-RAG memory-system
baseline by 1.30 points on LoCoMo (MemMachine, 88.12), 1.80 points
on LongMemEval (Memora, 87.40), and 1.15 points on MEME
(MemMachine, 37.06), using the original baseline scores.
The category-level results also remain mixed: for example,
\rawrag{} (5k) trails EMem on LongMemEval multi-session questions
and MEME aggregation, deletion, and cascades, despite its higher
Overall scores.

\begin{table}[!htbp]
  \caption{\rawrag{} accuracy (\%) by task with native evidence and a
  5,000-token cap (fixed). For LoCoMo, Single, Multi, Temp., and Open denote
  single-hop, multi-hop, temporal, and open-domain questions.
  For LongMemEval, S-U., S-A., and S-P.\ denote single-session user,
  assistant, and preference questions; M-S., multi-session; KU., knowledge
  update; TR., temporal reasoning. For MEME, ER denotes exact recall;
  Agg, aggregation; Tr, tracking; Del, deletion; Cas, cascade; Abs,
  invalidation requiring abstention. Bold marks the higher score in each
  column, including ties.}
  \label{tab:rag-5k-task-results}
  \centering\small
  \setlength{\tabcolsep}{4pt}
  \renewcommand{\arraystretch}{1.1}
  \begin{tabular}{lccccccc}
    \toprule
    \multicolumn{8}{>{\columncolor{gray!15}}c}{\modelname{LoCoMo: GPT-4o-mini}} \\
    \methodrule
    Condition & Single & Multi & Temp. & Open & & & Overall \\
    \midrule
    Native & \second{94.57} & \second{85.93} & \best{82.04} & \second{73.61} & & & \second{89.07} \\
    fixed & \best{95.05} & \best{88.06} & \second{80.17} & \best{75.00} & & & \best{89.42} \\
    \midrule
    \multicolumn{8}{>{\columncolor{gray!15}}c}{\modelname{LongMemEval: GPT-4.1-mini}} \\
    \methodrule
    Condition & S-U. & S-A. & S-P. & M-S. & KU. & TR. & Overall \\
    \midrule
    Native & \best{100.00} & \best{98.21} & \best{100.00} & \best{77.69} & \second{91.03} & \second{81.45} & \second{87.53} \\
    fixed & \second{98.57} & \best{98.21} & \second{96.67} & \second{74.94} & \best{95.30} & \best{89.47} & \best{89.20} \\
    \midrule
    \multicolumn{8}{>{\columncolor{gray!15}}c}{\modelname{MEME: GPT-4o-mini}} \\
    \methodrule
    Condition & ER & Agg & Tr & Del & Cas & Abs & Overall \\
    \midrule
    Native & \best{97.00} & \best{19.00} & \second{78.00} & \best{26.00} & \best{4.27} & \second{11.54} & \best{39.30} \\
    fixed & \second{93.00} & \second{13.00} & \best{81.00} & \second{24.00} & \second{3.66} & \best{14.62} & \second{38.21} \\
    \bottomrule
  \end{tabular}
\end{table}
\subsection{Diagnostic analyses}
\label{app:diagnostic-results}

The diagnostics address different questions: whether a particular mutation
policy helps long-term conversation understanding, whether readers can reconstruct state from supplied source
evidence, whether retrieved contexts contain sufficient evidence and paths,
and what changes under privileged access to stored representations.
Each diagnostic has its own evidence access and reader conditions,
defined with the statistical methods in
Appendix~\ref{app:diagnostic-methods}.

The main text reports the mutation audit
(Figure~\ref{fig:mem0-actions} and Table~\ref{tab:no-update}), source oracle
(Table~\ref{tab:meme-oracle}), and result/path assessment (Table~\ref{tab:path}).
We next report uncertainty and complete-store results.

\subsubsection{Coverage and utilization}
\label{app:coverage-significance}

Tables~\ref{tab:mab-audit} and~\ref{tab:meme-audit} give the main estimates.
The following tables supply omnibus tests, confidence intervals, and all
pairwise comparisons for the same evidence assessments. Coverage and utilization
are defined in Appendix~\ref{app:coverage-definitions}; resampling units,
test choices, and multiplicity correction are specified in
Appendix~\ref{app:statistical-methods}.
The omnibus comparison uses Cochran's $Q$ for MemoryAgentBench coverage
and the centered max-range bootstrap for the other three assessments.
Pairwise tests use exact McNemar tests for MemoryAgentBench coverage and
two-sided bootstrap sign tests for utilization; both MEME metrics use
episode-clustered two-sided bootstrap sign tests.

\begin{table*}[!htbp]
  \caption{Tests of cross-system differences in evidence coverage and
  utilization across seven systems. MemoryAgentBench coverage uses
  Cochran's $Q$ with six degrees of freedom; the remaining tests use a
  centered bootstrap of the maximum-minus-minimum range, measured in
  percentage points (pp). Holm sig.\ counts significant contrasts among
  the 21 system pairs after Holm correction at $\alpha=0.05$.}
  \label{tab:coverage-omnibus}
  \centering\small
  \setlength{\tabcolsep}{5pt}
  \renewcommand{\arraystretch}{1.15}
  \begin{tabular}{llccc}
    \toprule
    Benchmark & Metric & Statistic/range & $p$ & Holm sig. \\
    \midrule
    MemoryAgentBench & Coverage & $Q(6)=897.802$ & $1.12\!\times\!10^{-190}$ & 20/21 \\
    MemoryAgentBench & Utilization & 9.71 pp & $6.00\!\times\!10^{-5}$ & 7/21 \\
    MEME & Coverage & 40.66 pp & $2.00\!\times\!10^{-5}$ & 19/21 \\
    MEME & Utilization & 21.04 pp & $2.00\!\times\!10^{-5}$ & 17/21 \\
    \bottomrule
  \end{tabular}
\end{table*}
All four omnibus nulls are rejected. More pairwise contrasts are significant
for coverage than utilization on both benchmarks (20 versus 7 on
MemoryAgentBench and 19 versus 17 on MEME), and the observed coverage range is
larger in each case. The uncertainty analysis supports coverage as the
broader cross-system bottleneck, with significant variation in utilization
as well.

\begin{table*}[!htbp]
  \caption{Evidence coverage and conditional reader accuracy
  (utilization), with paired-bootstrap 95\% confidence intervals (CIs).
  Separate columns give each percentage, its supporting count $n/N$,
  and its CI.
  For coverage, $n/N$ counts sufficient contexts among all questions;
  for utilization, correct answers among sufficient contexts.
  Bold marks the highest estimate per benchmark and metric.}
  \label{tab:coverage-system-cis}
  \centering\small
  \setlength{\tabcolsep}{4pt}
  \renewcommand{\arraystretch}{1.1}
  \begin{tabular*}{\linewidth}{@{\extracolsep{\fill}}l*{6}{c}@{}}
    \toprule
    & \multicolumn{3}{c}{Coverage} & \multicolumn{3}{c}{Utilization} \\
    \cmidrule(lr){2-4}\cmidrule(lr){5-7}
    System & \% & $n/N$ & 95\% CI & \% & $n/N$ & 95\% CI \\
    \midrule
    \multicolumn{7}{>{\columncolor{gray!15}}c}{\textit{MemoryAgentBench}} \\
    \methodrule
    Mem0~\citep{chhikara2025mem0} & \second{40.03} & 829/2071 & [38.29, 41.77] & 82.39 & 683/829 & [79.84, 84.86] \\
    EMem~\citep{zhou2025simple} & 34.72 & 719/2071 & [33.17, 36.26] & \best{89.71} & 645/719 & [87.46, 91.87] \\
    MemOS~\citep{li2025memos} & 34.48 & 714/2071 & [32.83, 36.17] & 85.85 & 613/714 & [83.33, 88.30] \\
    EverMemOS~\citep{hu2026evermemos} & 25.11 & 520/2071 & [23.56, 26.65] & 80.00 & 416/520 & [76.60, 83.30] \\
    MemMachine~\citep{wang2026memmachine} & 21.54 & 446/2071 & [19.99, 23.08] & 84.30 & 376/446 & [80.89, 87.59] \\
    MemU~\citep{nevamind2025memu} & 19.07 & 395/2071 & [17.62, 20.57] & \second{89.37} & 353/395 & [86.24, 92.31] \\
    \methodrule
    \cellcolor{rawraghighlight}\textbf{\rawrag{}} & \best{45.29} & 938/2071 & [43.65, 46.93] & 84.43 & 792/938 & [82.15, 86.66] \\
    \midrule
    \multicolumn{7}{>{\columncolor{gray!15}}c}{\textit{MEME}} \\
    \methodrule
    MemMachine~\citep{wang2026memmachine} & \second{70.37} & 836/1188 & [68.47, 72.23] & 82.66 & 691/836 & [80.57, 84.75] \\
    Mem0~\citep{chhikara2025mem0} & 63.22 & 751/1188 & [60.41, 65.94] & 65.51 & 492/751 & [62.28, 68.83] \\
    MemOS~\citep{li2025memos} & 62.12 & 738/1188 & [59.59, 64.65] & 66.67 & 492/738 & [63.99, 69.37] \\
    EMem~\citep{zhou2025simple} & 50.08 & 595/1188 & [47.90, 52.31] & \best{86.55} & 515/595 & [83.89, 89.15] \\
    EverMemOS~\citep{hu2026evermemos} & 47.64 & 566/1188 & [44.87, 50.42] & 76.86 & 435/566 & [73.64, 79.96] \\
    MemU~\citep{nevamind2025memu} & 35.69 & 424/1188 & [33.33, 38.05] & \second{84.20} & 357/424 & [80.50, 87.78] \\
    \methodrule
    \cellcolor{rawraghighlight}\textbf{\rawrag{}} & \best{76.35} & 907/1188 & [74.32, 78.31] & 73.76 & 669/907 & [70.98, 76.44] \\
    \bottomrule
  \end{tabular*}
\end{table*}

Tables~\ref{tab:mab-pairwise} and~\ref{tab:meme-pairwise} report every
pairwise contrast, including nonsignificant ones, as system-by-system
matrices with a shared ordering. The upper triangle reports coverage and
the lower triangle utilization. Each cell gives the row-minus-column
difference in percentage points, its paired-bootstrap 95\% percentile
interval, and the Holm-adjusted $p_{\mathrm H}$, from top to bottom.
Bold differences and stars mark $p_{\mathrm H}<0.05$.

\begin{table*}[!htbp]
  \caption{Pairwise differences in MemoryAgentBench evidence coverage
  (upper triangle, blue) and utilization (lower triangle, gray).
  Each cell lists the row-minus-column difference in percentage points
  (pp), its paired-bootstrap 95\% confidence interval (CI), and the
  Holm-adjusted $p$-value $p_{\mathrm H}$.
  Bold and $^*$ mark $p_{\mathrm H}<0.05$, with correction over 21 pairs
  per metric. MemMach.~\citep{wang2026memmachine}\ denotes MemMachine~\citep{wang2026memmachine}.}
  \label{tab:mab-pairwise}
  \centering\footnotesize
  \setlength{\tabcolsep}{2pt}
  \renewcommand{\arraystretch}{1.15}
  \begin{tabular*}{\linewidth}{@{\extracolsep{\fill}}l*{7}{c}}
    \toprule
    System & \shortstack{Mem0\\\citep{chhikara2025mem0}} & \shortstack{EMem\\\citep{zhou2025simple}} & \shortstack{MemOS\\\citep{li2025memos}} & \shortstack{EverMemOS\\\citep{hu2026evermemos}} & \shortstack{MemMach.\\\citep{wang2026memmachine}} & \shortstack{MemU\\\citep{nevamind2025memu}} & \textbf{\rawrag{}} \\
    \midrule
    Mem0~\citep{chhikara2025mem0}
      & \cellcolor{gray!18}---
      & \cellcolor{blue!6}\paircell{\textbf{+5.31}\textsuperscript{*}}{[+3.43,+7.19]}{4.96\!\times\!10^{-7}}
      & \cellcolor{blue!6}\paircell{\textbf{+5.55}\textsuperscript{*}}{[+3.77,+7.34]}{2.91\!\times\!10^{-7}}
      & \cellcolor{blue!6}\paircell{\textbf{+14.92}\textsuperscript{*}}{[+12.99,+16.85]}{5.02\!\times\!10^{-46}}
      & \cellcolor{blue!6}\paircell{\textbf{+18.49}\textsuperscript{*}}{[+16.37,+20.57]}{3.12\!\times\!10^{-54}}
      & \cellcolor{blue!6}\paircell{\textbf{+20.96}\textsuperscript{*}}{[+18.98,+22.89]}{3.47\!\times\!10^{-67}}
      & \cellcolor{blue!6}\paircell{\textbf{-5.26}\textsuperscript{*}}{[-7.39,-3.14]}{9.84\!\times\!10^{-6}} \\
    \addlinespace[2pt]
    EMem~\citep{zhou2025simple}
      & \cellcolor{gray!8}\paircell{\textbf{+7.32}\textsuperscript{*}}{[+4.43,+10.22]}{0.00084}
      & \cellcolor{gray!18}---
      & \cellcolor{blue!6}\paircell{+0.24}{[-1.64,+2.12]}{0.853}
      & \cellcolor{blue!6}\paircell{\textbf{+9.61}\textsuperscript{*}}{[+7.63,+11.59]}{3.27\!\times\!10^{-18}}
      & \cellcolor{blue!6}\paircell{\textbf{+13.18}\textsuperscript{*}}{[+11.25,+15.11]}{1.74\!\times\!10^{-33}}
      & \cellcolor{blue!6}\paircell{\textbf{+15.64}\textsuperscript{*}}{[+13.71,+17.58]}{1.40\!\times\!10^{-37}}
      & \cellcolor{blue!6}\paircell{\textbf{-10.57}\textsuperscript{*}}{[-12.51,-8.64]}{4.84\!\times\!10^{-23}} \\
    \addlinespace[2pt]
    MemOS~\citep{li2025memos}
      & \cellcolor{gray!8}\paircell{+3.47}{[+0.54,+6.41]}{0.225}
      & \cellcolor{gray!8}\paircell{-3.85}{[-6.76,-0.94]}{0.124}
      & \cellcolor{gray!18}---
      & \cellcolor{blue!6}\paircell{\textbf{+9.37}\textsuperscript{*}}{[+7.44,+11.35]}{6.64\!\times\!10^{-18}}
      & \cellcolor{blue!6}\paircell{\textbf{+12.94}\textsuperscript{*}}{[+10.91,+14.97]}{6.02\!\times\!10^{-27}}
      & \cellcolor{blue!6}\paircell{\textbf{+15.40}\textsuperscript{*}}{[+13.42,+17.38]}{7.51\!\times\!10^{-37}}
      & \cellcolor{blue!6}\paircell{\textbf{-10.82}\textsuperscript{*}}{[-12.80,-8.84]}{1.26\!\times\!10^{-19}} \\
    \addlinespace[2pt]
    EverMemOS~\citep{hu2026evermemos}
      & \cellcolor{gray!8}\paircell{-2.39}{[-6.01,+1.20]}{1.000}
      & \cellcolor{gray!8}\paircell{\textbf{-9.71}\textsuperscript{*}}{[-13.23,-6.22]}{0.00084}
      & \cellcolor{gray!8}\paircell{\textbf{-5.85}\textsuperscript{*}}{[-9.63,-2.10]}{0.0352}
      & \cellcolor{gray!18}---
      & \cellcolor{blue!6}\paircell{\textbf{+3.57}\textsuperscript{*}}{[+1.69,+5.46]}{0.00235}
      & \cellcolor{blue!6}\paircell{\textbf{+6.04}\textsuperscript{*}}{[+4.01,+8.06]}{1.76\!\times\!10^{-6}}
      & \cellcolor{blue!6}\paircell{\textbf{-20.18}\textsuperscript{*}}{[-22.11,-18.25]}{3.86\!\times\!10^{-73}} \\
    \addlinespace[2pt]
    MemMachine~\citep{wang2026memmachine}
      & \cellcolor{gray!8}\paircell{+1.92}{[-1.89,+5.71]}{1.000}
      & \cellcolor{gray!8}\paircell{\textbf{-5.40}\textsuperscript{*}}{[-8.94,-1.88]}{0.0408}
      & \cellcolor{gray!8}\paircell{-1.55}{[-5.28,+2.09]}{1.000}
      & \cellcolor{gray!8}\paircell{+4.30}{[+0.22,+8.42]}{0.355}
      & \cellcolor{gray!18}---
      & \cellcolor{blue!6}\paircell{\textbf{+2.46}\textsuperscript{*}}{[+0.48,+4.44]}{0.0445}
      & \cellcolor{blue!6}\paircell{\textbf{-23.76}\textsuperscript{*}}{[-25.64,-21.87]}{3.18\!\times\!10^{-107}} \\
    \addlinespace[2pt]
    MemU~\citep{nevamind2025memu}
      & \cellcolor{gray!8}\paircell{\textbf{+6.98}\textsuperscript{*}}{[+3.36,+10.46]}{0.00408}
      & \cellcolor{gray!8}\paircell{-0.34}{[-3.88,+3.11]}{1.000}
      & \cellcolor{gray!8}\paircell{+3.51}{[-0.16,+7.12]}{0.492}
      & \cellcolor{gray!8}\paircell{\textbf{+9.37}\textsuperscript{*}}{[+5.07,+13.61]}{0.00084}
      & \cellcolor{gray!8}\paircell{+5.06}{[+0.77,+9.30]}{0.225}
      & \cellcolor{gray!18}---
      & \cellcolor{blue!6}\paircell{\textbf{-26.22}\textsuperscript{*}}{[-28.25,-24.19]}{1.22\!\times\!10^{-95}} \\
    \addlinespace[2pt]
    \textbf{\rawrag{}}
      & \cellcolor{gray!8}\paircell{+2.05}{[-0.95,+5.06]}{1.000}
      & \cellcolor{gray!8}\paircell{\textbf{-5.27}\textsuperscript{*}}{[-7.92,-2.62]}{0.00216}
      & \cellcolor{gray!8}\paircell{-1.42}{[-4.34,+1.50]}{1.000}
      & \cellcolor{gray!8}\paircell{+4.43}{[+0.95,+7.98]}{0.163}
      & \cellcolor{gray!8}\paircell{+0.13}{[-3.33,+3.59]}{1.000}
      & \cellcolor{gray!8}\paircell{-4.93}{[-8.31,-1.46]}{0.0689}
      & \cellcolor{gray!18}--- \\
    \bottomrule
  \end{tabular*}
\end{table*}

\begin{table*}[!htbp]
  \caption{Pairwise differences in MEME evidence coverage
  (upper triangle, blue) and utilization (lower triangle, gray).
  Each cell lists the row-minus-column difference in percentage points
  (pp), its paired episode-bootstrap 95\% confidence interval (CI), and
  the Holm-adjusted $p$-value $p_{\mathrm H}$.
  Bold and $^*$ mark $p_{\mathrm H}<0.05$, with correction over 21 pairs
  per metric. MemMach.~\citep{wang2026memmachine}\ denotes MemMachine~\citep{wang2026memmachine}.}
  \label{tab:meme-pairwise}
  \centering\footnotesize
  \setlength{\tabcolsep}{1.5pt}
  \renewcommand{\arraystretch}{1.15}
  \begin{tabular*}{\linewidth}{@{\extracolsep{\fill}}l*{7}{c}}
    \toprule
    System & \shortstack{Mem0\\\citep{chhikara2025mem0}} & \shortstack{EMem\\\citep{zhou2025simple}} & \shortstack{MemOS\\\citep{li2025memos}} & \shortstack{EverMemOS\\\citep{hu2026evermemos}} & \shortstack{MemMach.\\\citep{wang2026memmachine}} & \shortstack{MemU\\\citep{nevamind2025memu}} & \textbf{\rawrag{}} \\
    \midrule
    Mem0~\citep{chhikara2025mem0}
      & \cellcolor{gray!18}---
      & \cellcolor{blue!6}\paircell{\textbf{+13.13}\textsuperscript{*}}{[+9.98,+16.22]}{0.00084}
      & \cellcolor{blue!6}\paircell{+1.09}{[-1.51,+3.66]}{0.432}
      & \cellcolor{blue!6}\paircell{\textbf{+15.57}\textsuperscript{*}}{[+12.23,+18.92]}{0.00084}
      & \cellcolor{blue!6}\paircell{\textbf{-7.15}\textsuperscript{*}}{[-10.03,-4.28]}{0.00084}
      & \cellcolor{blue!6}\paircell{\textbf{+27.53}\textsuperscript{*}}{[+24.06,+30.96]}{0.00084}
      & \cellcolor{blue!6}\paircell{\textbf{-13.13}\textsuperscript{*}}{[-15.95,-10.35]}{0.00084} \\
    \addlinespace[2pt]
    EMem~\citep{zhou2025simple}
      & \cellcolor{gray!8}\paircell{\textbf{+21.04}\textsuperscript{*}}{[+16.94,+25.08]}{0.00084}
      & \cellcolor{gray!18}---
      & \cellcolor{blue!6}\paircell{\textbf{-12.04}\textsuperscript{*}}{[-14.80,-9.26]}{0.00084}
      & \cellcolor{blue!6}\paircell{+2.44}{[-0.82,+5.68]}{0.294}
      & \cellcolor{blue!6}\paircell{\textbf{-20.29}\textsuperscript{*}}{[-22.85,-17.63]}{0.00084}
      & \cellcolor{blue!6}\paircell{\textbf{+14.39}\textsuperscript{*}}{[+11.66,+17.06]}{0.00084}
      & \cellcolor{blue!6}\paircell{\textbf{-26.26}\textsuperscript{*}}{[-28.78,-23.69]}{0.00084} \\
    \addlinespace[2pt]
    MemOS~\citep{li2025memos}
      & \cellcolor{gray!8}\paircell{+1.15}{[-2.93,+5.25]}{0.890}
      & \cellcolor{gray!8}\paircell{\textbf{-19.89}\textsuperscript{*}}{[-23.18,-16.55]}{0.00084}
      & \cellcolor{gray!18}---
      & \cellcolor{blue!6}\paircell{\textbf{+14.48}\textsuperscript{*}}{[+11.39,+17.64]}{0.00084}
      & \cellcolor{blue!6}\paircell{\textbf{-8.25}\textsuperscript{*}}{[-11.34,-5.16]}{0.00084}
      & \cellcolor{blue!6}\paircell{\textbf{+26.43}\textsuperscript{*}}{[+23.40,+29.43]}{0.00084}
      & \cellcolor{blue!6}\paircell{\textbf{-14.23}\textsuperscript{*}}{[-17.15,-11.24]}{0.00084} \\
    \addlinespace[2pt]
    EverMemOS~\citep{hu2026evermemos}
      & \cellcolor{gray!8}\paircell{\textbf{+11.34}\textsuperscript{*}}{[+6.41,+16.11]}{0.00084}
      & \cellcolor{gray!8}\paircell{\textbf{-9.70}\textsuperscript{*}}{[-13.57,-5.91]}{0.00084}
      & \cellcolor{gray!8}\paircell{\textbf{+10.19}\textsuperscript{*}}{[+6.14,+14.14]}{0.00084}
      & \cellcolor{gray!18}---
      & \cellcolor{blue!6}\paircell{\textbf{-22.73}\textsuperscript{*}}{[-25.54,-19.93]}{0.00084}
      & \cellcolor{blue!6}\paircell{\textbf{+11.95}\textsuperscript{*}}{[+8.53,+15.38]}{0.00084}
      & \cellcolor{blue!6}\paircell{\textbf{-28.70}\textsuperscript{*}}{[-31.71,-25.67]}{0.00084} \\
    \addlinespace[2pt]
    MemMachine~\citep{wang2026memmachine}
      & \cellcolor{gray!8}\paircell{\textbf{+17.14}\textsuperscript{*}}{[+13.10,+21.11]}{0.00084}
      & \cellcolor{gray!8}\paircell{\textbf{-3.90}\textsuperscript{*}}{[-6.86,-0.95]}{0.0474}
      & \cellcolor{gray!8}\paircell{\textbf{+15.99}\textsuperscript{*}}{[+12.57,+19.34]}{0.00084}
      & \cellcolor{gray!8}\paircell{\textbf{+5.80}\textsuperscript{*}}{[+2.37,+9.36]}{0.00512}
      & \cellcolor{gray!18}---
      & \cellcolor{blue!6}\paircell{\textbf{+34.68}\textsuperscript{*}}{[+31.42,+37.94]}{0.00084}
      & \cellcolor{blue!6}\paircell{\textbf{-5.98}\textsuperscript{*}}{[-7.65,-4.34]}{0.00084} \\
    \addlinespace[2pt]
    MemU~\citep{nevamind2025memu}
      & \cellcolor{gray!8}\paircell{\textbf{+18.69}\textsuperscript{*}}{[+13.87,+23.40]}{0.00084}
      & \cellcolor{gray!8}\paircell{-2.36}{[-6.42,+1.80]}{0.789}
      & \cellcolor{gray!8}\paircell{\textbf{+17.53}\textsuperscript{*}}{[+13.71,+21.42]}{0.00084}
      & \cellcolor{gray!8}\paircell{\textbf{+7.34}\textsuperscript{*}}{[+2.92,+11.90]}{0.00744}
      & \cellcolor{gray!8}\paircell{+1.54}{[-2.48,+5.53]}{0.890}
      & \cellcolor{gray!18}---
      & \cellcolor{blue!6}\paircell{\textbf{-40.66}\textsuperscript{*}}{[-43.70,-37.60]}{0.00084} \\
    \addlinespace[2pt]
    \textbf{\rawrag{}}
      & \cellcolor{gray!8}\paircell{\textbf{+8.25}\textsuperscript{*}}{[+3.52,+12.90]}{0.00512}
      & \cellcolor{gray!8}\paircell{\textbf{-12.79}\textsuperscript{*}}{[-16.33,-9.15]}{0.00084}
      & \cellcolor{gray!8}\paircell{\textbf{+7.09}\textsuperscript{*}}{[+3.71,+10.47]}{0.00084}
      & \cellcolor{gray!8}\paircell{-3.10}{[-7.01,+0.89]}{0.504}
      & \cellcolor{gray!8}\paircell{\textbf{-8.90}\textsuperscript{*}}{[-11.74,-6.10]}{0.00084}
      & \cellcolor{gray!8}\paircell{\textbf{-10.44}\textsuperscript{*}}{[-14.78,-6.12]}{0.00084}
      & \cellcolor{gray!18}--- \\
    \bottomrule
  \end{tabular*}
\end{table*}

\subsubsection{Complete-store evaluation}
\label{app:complete-store-results}

\paragraph{Evaluation scope.}
The diagnostic covers the same 100 MEME episodes with 32k distractor tokens
and 1,188 question IDs in every condition: 494 before-phase and
694 after-phase questions. Native ($N$) uses the designated system's
ordinary retrieval and reader path. Complete-store ($C$) replaces
native retrieval with privileged selection over the exported persisted
representations, subject to the phase/provenance rules in
Appendix~\ref{app:complete-store}, and adds a task-aware reader policy.
Raw evidence ($R$) selects original
support sessions with the same gold signals and task policy.
All three columns use GPT-4o-mini answers; Overall is the macro-average
of the six after-phase task accuracies, including the benchmark's
before-pass gate for Del, Cas, and Abs.
The repeated $R$ column reports one shared raw-evidence run.

\begin{table}[htbp]
  \centering
  \caption{MEME complete-store diagnostic with GPT-4o-mini, reported as
  six-task macro-accuracy (\%). $N$ uses native retrieval; $C$ uses
  gold-informed selection from the complete store; $R$ is the shared
  raw-evidence control. Both $C$ and $R$ use a 5,000-token budget and
  task-aware reader instructions. Differences and paired episode-bootstrap
  95\% confidence intervals (CIs) are in percentage points.
  Bold marks the highest score in $N$ and $C$.}
  \label{tab:complete-store-full}
  \small
  \setlength{\tabcolsep}{3pt}
  \renewcommand{\arraystretch}{1.2}
  \begin{tabular}{lccccc}
    \toprule
    System & $N$ & $C$ & $R$ & $C-N$ [95\% CI] & $R-C$ [95\% CI] \\
    \midrule
    \multicolumn{6}{>{\columncolor{gray!15}}c}{\modelname{GPT-4o-mini}} \\
    \methodrule
    MemMachine~\citep{wang2026memmachine} & \second{37.06} & \second{57.46} & 61.60 & +20.40 [17.30, 23.40] & +4.14 [0.23, 8.07] \\
    EverMemOS~\citep{hu2026evermemos} & 21.57 & 51.59 & 61.60 & +30.02 [26.53, 33.47] & +10.01 [6.17, 13.87] \\
    EMem~\citep{zhou2025simple} & 33.84 & 48.36 & 61.60 & +14.52 [10.58, 18.42] & +13.24 [9.27, 17.19] \\
    MemOS~\citep{li2025memos} & 30.24 & 45.12 & 61.60 & +14.88 [11.04, 18.57] & +16.48 [12.76, 20.19] \\
    Mem0~\citep{chhikara2025mem0} & 27.76 & 32.26 & 61.60 & +4.50 [1.75, 7.27] & +29.34 [25.47, 33.10] \\
    MemU~\citep{nevamind2025memu} & 19.87 & 31.07 & 61.60 & +11.20 [8.48, 13.93] & +30.54 [27.33, 33.73] \\
    \methodrule
    \cellcolor{rawraghighlight}\textbf{\rawrag{}} & \best{39.30} & \best{57.54} & 61.60 & +18.24 [14.78, 21.65] & +4.06 [0.74, 7.50] \\
    \bottomrule
  \end{tabular}
\end{table}

\paragraph{Packing-cap sensitivity.}
To test whether the nominal 5,000-token packing ceiling drives the
complete-store results, we repeat the diagnostic with that ceiling removed
while holding the exported candidate set, gold-informed selector, ordering,
task policy, reader, and scoring fixed.  The raw control is subjected to the
same intervention.  Table~\ref{tab:complete-store-uncapped} reports paired
results over the same 100 episodes and 1,188 question IDs.  ``Affected'' counts
questions whose assembled reader evidence changes; dropped tokens are counted
before the outer prompt and question are added.

\begin{table}[htbp]
  \centering
  \caption{Effect of removing the 5,000-token evidence cap in the MEME
  complete-store diagnostic with GPT-4o-mini. $C_{5k}$ and $C_{\infty}$
  are capped and uncapped six-task macro-accuracies (\%); $R$ is the
  shared raw control. Affected counts questions whose evidence changes;
  Dropped is the percentage of evidence tokens omitted by the cap.
  Differences and paired episode-bootstrap 95\% confidence intervals
  (CIs) are in percentage points. Bold marks score-column maxima
  excluding the raw control.}
  \label{tab:complete-store-uncapped}
  \footnotesize
  \setlength{\tabcolsep}{2.7pt}
  \renewcommand{\arraystretch}{1.15}
  \begin{tabular}{lcccccc}
    \toprule
    Condition & Affected & Dropped & $C_{5k}$ & $C_{\infty}$
      & $C_{\infty}-C_{5k}$ [95\% CI] & $R-C_{\infty}$ [95\% CI] \\
    \midrule
    \multicolumn{7}{>{\columncolor{gray!15}}c}{\modelname{GPT-4o-mini}} \\
    \methodrule
    \textit{Raw control} & 0 & 0.00\% & 61.60 & 61.60
      & +0.00 [+0.00, +0.00] & +0.00 [+0.00, +0.00] \\
    \methodrule
    MemMachine~\citep{wang2026memmachine} & 1,172 & 83.39\% & \second{57.46} & \best{54.02}
      & $-3.44$ [$-6.49$, $-0.37$] & +7.58 [+3.03, +12.15] \\
    Mem0~\citep{chhikara2025mem0} & 62 & 1.45\% & 32.26 & 32.22
      & $-0.04$ [$-1.04$, +0.96] & +29.38 [+25.58, +33.18] \\
    MemOS~\citep{li2025memos} & 408 & 12.79\% & 45.12 & 44.58
      & $-0.53$ [$-1.60$, +0.44] & +17.02 [+13.22, +20.80] \\
    MemU~\citep{nevamind2025memu} & 604 & 63.07\% & 31.07 & 32.55
      & +1.49 [$-0.33$, +3.28] & +29.05 [+25.77, +32.36] \\
    EMem~\citep{zhou2025simple} & 171 & 26.33\% & 48.36 & 48.19
      & $-0.17$ [$-0.83$, +0.50] & +13.41 [+9.53, +17.20] \\
    EverMemOS~\citep{hu2026evermemos} & 920 & 65.77\% & 51.59 & 51.19
      & $-0.40$ [$-2.67$, +1.82] & +10.42 [+6.54, +14.39] \\
    \methodrule
    \cellcolor{rawraghighlight}\textbf{\rawrag{}} & 1,188 & 84.59\% & \best{57.54} & \second{53.85}
      & $-3.69$ [$-7.92$, +0.55] & +7.76 [+3.12, +12.46] \\
    \bottomrule
  \end{tabular}
\end{table}

All seven stores and the raw control have complete results for all 1,188
questions. Removing the ceiling lowers MemMachine's Overall from 57.46\%
to 54.02\%, a change of $-3.44$ points [$-6.49$, $-0.37$]; the other
six stores' $C_{\infty}-C_{5k}$ intervals include zero. MemMachine's
mean assembled evidence grows from 4,982.8 to 29,459.7 tokens, with no
context-window failures. Its uncapped ER, Agg, Tr, Del, Cas, and Abs
scores are 95.00, 50.00, 55.00, 61.00, 46.95, and 16.15\%, respectively.
On the subset whose reader evidence changes, MemU improves by 2.70 points
[0.30, 5.23], whereas MemMachine declines by 3.46 points
[$-6.49$, $-0.40$]; the other subset intervals include zero.
These subset results describe questions with changed inputs. Every
raw-minus-uncapped-store interval remains positive.  The residual gap in
Table~\ref{tab:complete-store-full} therefore cannot be attributed solely to
the nominal packing ceiling.

\paragraph{Content-filter sensitivity.}
The full results count nine content-filtered reader responses for
\rawrag{} and seven for MemMachine as empty, incorrect answers.
As a sensitivity check, we exclude the union of filtered questions and
their paired before/after questions used in gating from every condition: 12
questions in total, leaving 687 after-phase questions. On this common
subset, MemMachine changes from 57.46\% to 54.85\%, or $-2.61$ points
[$-5.56$, +0.36], and \rawrag{} changes from 57.54\% to 54.68\%,
or $-2.86$ points [$-6.92$, +1.28]; the raw control is 61.53\%.
All seven cap-removal intervals now include zero. Thus MemMachine's
negative full-set interval is sensitive to provider filtering.

\paragraph{Condition alignment.}
Table~\ref{tab:complete-store-full} uses the same Native conditions as
Figure~\ref{fig:complete-store-loss}. RAG Native averages three judge
decisions per answer (39.30\%); the first decision alone gives 39.80\%.
EMem, EverMemOS, and MemMachine Native answers
are scored with GPT-4o. Each $C-N$ comparison pairs question-level
scores from the stated Native condition with its complete-store
condition. Mem0 uses the local memory system (27.76\%); its cloud
counterpart scores 29.12\%. MemOS uses fine-grained extraction.
MemU uses local item-only memory (Appendix~\ref{app:baseline-details}).

The paired episode-bootstrap procedure and its uncertainty scope are
specified in Appendix~\ref{app:statistical-methods}.

\paragraph{Source-oracle comparisons.}
Table~\ref{tab:oracle-protocols} compares the two GPT-4o-mini raw-source
conditions defined in Appendix~\ref{app:complete-store}. Their selection,
packing, and reader instructions differ, and their score difference
reflects these combined changes.

\begin{table}[htbp]
  \centering
  \caption{MEME accuracy (\%) with GPT-4o-mini under two raw-source
  oracle configurations that differ in evidence selection, context assembly,
  and reader instructions. ER denotes exact recall; Agg, aggregation;
  Tr, tracking; Del, deletion; Cas, cascade; Abs, invalidation requiring
  abstention. Overall is the six-task macro-average.
  Bold marks the higher score in each column.}
  \label{tab:oracle-protocols}
  \small
  \setlength{\tabcolsep}{4pt}
  \begin{tabular}{lccccccc}
    \toprule
    Raw-source setting & ER & Agg & Tr & Del & Cas & Abs & Overall \\
    \midrule
    \multicolumn{8}{>{\columncolor{gray!15}}c}{\modelname{GPT-4o-mini}} \\
    \methodrule
    Answer-location oracle & \best{98} & \second{47} & \best{81} & \second{39} & \second{1.83} & \best{36.15} & \second{50.50} \\
    Gold-informed raw control & \second{94} & \best{52} & \second{69} & \best{92} & \best{34.15} & \second{28.46} & \best{61.60} \\
    \bottomrule
  \end{tabular}
\end{table}

All $R-C$ intervals in Table~\ref{tab:complete-store-full} exclude zero;
Appendix~\ref{app:complete-store-boundary} discusses limitations.

\subsection{Efficiency and scalability}
\label{app:efficiency-results}

We distinguish system-level latency, scaling with history length, and the
cost of additional memory-management operations. Table~\ref{tab:latency} reports
Add-to-ready and Search for the same 10-conversation, 110-question MEME
slice. Figure~\ref{fig:teaser-results} reports cumulative sequential
ingestion over one 1,710-message BEAM-1M conversation. Both use one
outer worker and allow each system's native internal parallelism;
neither includes answer generation or evaluation. Timing boundaries,
readiness rules, and execution environments are detailed in
Appendices~\ref{app:timing-protocol} and~\ref{app:environment}.

Table~\ref{tab:reverse-ablation} separately reports the computation and
latency of the memory-management additions on the 11 commonly measured
non-conflict MemoryAgentBench sources. The timing totals cover Add and Search
in this deployment; Table~\ref{tab:ablation} compares accuracy across all 14 sources.

\begin{table}[!htbp]
  \caption{LLM usage and elapsed time for \rawrag{} and its
  memory-management variants on the 11 MemoryAgentBench sources outside
  Conflict Resolution. Calls counts LLM requests; Prompt, Compl.\
  (completion), and Tokens report input, output, and total tokens in
  millions (M). Add and Search measure ingestion and retrieval time;
  Total is their sum. All times are in seconds (s).
  Measurements exclude answer generation and evaluation.
  Bold marks column minima.}
  \label{tab:reverse-ablation}
  \centering\scriptsize
  \setlength{\tabcolsep}{3pt}
  \resizebox{\columnwidth}{!}{%
  \begin{tabular}{lccccccc}
    \toprule
    Configuration & Calls & Prompt (M) & Compl. (M) & Tokens (M) & Add (s) & Search (s) & Total (s) \\
    \midrule
    Atomic facts
    & 40{,}248 & 24.0 & 22.3 & 46.3 & 28{,}837.0 & 2{,}836.8 & 31{,}673.8 \\
    Raw summary
    & \second{32{,}144} & \second{14.8} & 6.8 & \second{21.6} & \second{8{,}804.4} & \second{1{,}523.1} & \second{10{,}327.5} \\
    Graph-expanded
    & 33{,}748 & 105.7 & \second{3.7} & 109.3 & 11{,}506.3 & 4{,}093.3 & 15{,}599.6 \\
    \methodrule
    \cellcolor{rawraghighlight}\textbf{\rawrag}
    & \best{0} & \best{0} & \best{0} & \best{0} & \best{272.6} & \best{445.8} & \best{718.4} \\
    \bottomrule
  \end{tabular}}
\end{table}

\section{Raw-RAG: Method Details}
\label{app:method}

This appendix details the \rawrag{} pipeline, from source-preserving
indexing and retrieval to evidence assembly and reader input construction.
We describe query construction, benchmark-specific adaptations, and
the distinction between retrieval selection and reader-visible evidence.
We also specify native context-length and overflow policies, separating
them from the diagnostic and context-budget interventions.

\subsection{Indexing and evidence assembly}

\paragraph{Source-preserving representation.}
Let $H=(m_1,\ldots,m_T)$ denote the available history. Each message is
represented by its date, speaker, and content; missing dates are marked
as unknown. Supplied image captions enter as text.
``Raw'' denotes preservation of this source text. Token windows have length 256 and stride
$256-32=224$, measured with the GPT-4 tokenizer. Each window remains
linked to its original messages so that complete turns can be recovered
when assembling evidence for the reader.

\paragraph{Retrieval and evidence assembly.}
For question $q$, an embedding encoder $e$ selects a candidate set
$D_q$, and a reranker $r$ selects the reader evidence $S_q$:
\begin{align}
  D_q &= \operatorname{Top}_{128}
    \left\{\cos(e(q),e(c)):c\in\operatorname{chunks}(H)\right\},\\
  S_q &= \operatorname{Top}_{15}
    \left\{r(q,c):c\in D_q\right\}.
\end{align}
The index computes cosine similarity within the relevant
conversation. The index is fixed after ingestion, and selection uses
a fixed top-$k$ budget.
Where turn expansion is enabled,
each selected window is rendered as the complete source turns that
overlap it. Expansion changes only the reader-visible text; selection
scores remain unchanged.
Repeated turns across selected windows are included in the default
configuration. The turn-deduplication condition is a separate ablation.
Headers, overlapping content, and turn expansion determine the actual
reader context length for the 15 selected windows.

\paragraph{Benchmark adaptations.}
MEME presents the selected chunks in source order after relevance-based
selection to preserve temporal progression. We use the condition with
32k tokens of distractor content;
before/after questions use the evidence available at the corresponding
phase. MemoryAgentBench disables display-time turn expansion. For its
two fact-consolidation sources, each fact is treated as one message,
and the index uses one complete message per chunk with no overlap.
These atomic facts preserve their serial-number cues for conflict
resolution. Other MemoryAgentBench sources use the token-window
configuration.

\subsection{Query construction}
\label{app:query-construction}

\paragraph{Retrieval query.}
For \rawrag{}, the retrieval query is the original question or user request supplied by the
benchmark: the same question text is sent to the embedding encoder and paired
with each candidate passage for reranking.

\paragraph{Task conditioning.}
LoCoMo, LongMemEval, MEME, and PersonaMem-v2 use the original question text.
When a question date or evaluation phase is available, it determines
evidence visibility. MemoryAgentBench likewise retrieves with the original
task question; the instructions for each source are added only
when constructing the reader input.

\paragraph{Reranker formulation.}
The Qwen3 reranker inserts the unchanged query and candidate passage
into a relevance-classification template:
``Given a question and a passage, determine if the passage contains
information relevant to answering the question.'' A system instruction
restricts its relevance decision to \texttt{yes} or \texttt{no}.
See the full template in Appendix~\ref{app:prompt-qwen-reranker}.

\paragraph{Reader input.}
Retrieval queries are distinct from the final answer-generation prompts.
After evidence selection, the reader receives the retrieved context and
the benchmark-specific question instructions: a question-date reminder
where applicable, the appropriate MemoryAgentBench task template, or
the PersonaMem-v2 personalization reminder followed by lettered response
options.
Appendix~\ref{app:prompts} reproduces the reader templates and
defines all per-example substitutions.

\subsection{Context-length policies}
\label{app:native-context-policy}

The following rules distinguish ordinary reader-input assembly from the
explicit context-budget interventions. Actual evidence-length distributions
and the resulting capped scores are reported in
Appendix~\ref{app:reader-context-budget}.

\paragraph{Truncation and overflow.}
The main \rawrag{} conditions have no additional evidence-token cap
and no automatic head--tail clipping after turn expansion. In all
seven examined benchmark--model conditions, the saved reader evidence
is identical to the assembled retrieval evidence, with no recorded
answer-generation error responses. The reader receives the evidence plus its task prompt and
question, and the configured output allowance is separate. The
pipeline submits the full input with that allowance. If a provider rejects
an oversized request, the default three-attempt retry policy resends the
same input. A request that exhausts these attempts is scored as an
answer-generation failure.

Other evaluated configurations have different, explicit policies.
The 100,000-token evidence cap used for SimpleMem and Memora keeps
the beginning of the assembled context in its existing order, clips
at the token boundary, and omits a truncation marker. It uses the reader's recognized
\texttt{tiktoken} encoding, falling back to \texttt{o200k\_base};
the evidence cap excludes the question and fixed prompt.
The local Full-context condition instead budgets the assembled
request within 128,000 tokens, reserving 4,096 output tokens and a
256-token margin. When needed, it keeps approximately equal
character-length prefixes and suffixes of the history, inserts an
explicit truncation marker, and checks the resulting request length;
it fails if the prompt and question alone cannot fit.
The MEME complete-store packing rules are specified separately in
Appendix~\ref{app:complete-store}.

\FloatBarrier

\section{Diagnostic Procedures and Statistical Methods}
\label{app:diagnostic-methods}

This appendix defines the diagnostic procedures and statistical methods
used in our analyses. We formalize evidence coverage, utilization, and
executable-path retention, and describe memory-management ablations,
write-policy and context-budget interventions, and source-oracle and
complete-store controls. We also clarify
the limits of their interpretation and specify the paired statistical tests,
bootstrap confidence intervals, and multiple-comparison corrections.

\subsection{Coverage, utilization, and path retention}
\label{app:coverage-definitions}

For question $i$ and system $s$, let $C_{is}\in\{0,1\}$ indicate
assessed evidence sufficiency and let $Y_{is}\in[0,1]$ be its evaluated
answer score. The question-weighted summaries are
\begin{equation}
  \widehat{C}_s=\frac{1}{N}\sum_i C_{is},
  \qquad
  \widehat{U}_s=\frac{\sum_i C_{is}Y_{is}}{\sum_i C_{is}}.
\end{equation}
Utilization is undefined if no question is judged sufficient.
The product $\widehat{C}_s\widehat{U}_s$ is the score contributed by
questions with sufficient evidence. Total accuracy additionally includes
$N^{-1}\sum_i(1-C_{is})Y_{is}$.
Correct answers outside the sufficient set may reflect sufficiency-label errors,
partial evidence, prior knowledge, or guessing. Official macro-averages
use different weights.

MemoryAgentBench uses semantic sufficiency judgments, whereas the
MEME task-specific assessment checks required values, temporal cues,
deletion evidence, and dependency triggers. Deletion coverage requires
an explicit deletion event.
For chain-sensitive after-state questions, result coverage and
executable-path coverage are judged separately. Conditional path
retention is the fraction of result-covered questions whose context
also contains the required path. Task-specific
sufficiency rules are given below. Paired uncertainty estimates appear in
Appendix~\ref{app:coverage-significance}, with statistical procedures in
Appendix~\ref{app:statistical-methods}.

\paragraph{Task-specific sufficiency.}
ER requires the target value; Agg all aggregated values; Tr also requires
chronology; Del-after an explicit deletion event; Cas-after the new target,
dependency rule, and upstream trigger; and Abs-after the old target,
dependency, and invalidating trigger.

\subsection{Write-policy and evidence controls}

\paragraph{Mem0 mutation audit.}
This analysis distinguishes emitted decisions, returned actions, and
persisted history records. Proposed mutation rates count
\texttt{UPDATE} and \texttt{DELETE} decisions relative to all emitted
decisions, including \texttt{NONE}. An effective update must have a
persisted record with available old and new text that differ; a
persisted deletion also counts as effective mutation. Effectiveness is
defined by persisted text changes. In the ADD/NONE-only intervention, proposed
updates become additions and proposed deletions become no-ops.
Figure~\ref{fig:mem0-actions} orders all 500 LongMemEval conversations by
proposed mutation rate, pairing each proposed rate (orange) with its
effective rate (blue).
The experiment tests the usefulness of that mutation policy for
downstream QA.

\paragraph{Evidence interventions.}
The candidate sets, selection rules, packing budgets, and reader instructions
for the answer-location source oracle and gold-informed complete-store
diagnostic are specified below.

\subsection{Memory-management ablations}
\label{app:agentic-mutation-ablations}

The three memory-management variants in Table~\ref{tab:ablation} reuse the
\rawrag{} retrieval and reader stack, changing the representation built
at ingestion or the candidate set passed to reranking. Unless stated
otherwise, all conditions tokenize each conversation into 256-token
windows with 32-token overlap, embed the resulting items with
\texttt{text-embedding-3-small}, retrieve the 128 highest cosine-similarity
candidates, rerank those candidates with the same cross-encoder, and pass
the final 15 items to the reader. All LLM calls for these operations use
\texttt{gpt-4o-mini-2024-07-18}, temperature zero, and a JSON object
containing only the requested fields; malformed or empty outputs are
retried, and repeated failure aborts the build.

\paragraph{Atomic facts.}
For each raw window, the extractor returns at most six standalone facts.
Each fact contains the fact text, speaker, time, within-window order,
serial/newness marker, up to five entities, and a short supporting quote.
Facts are restricted to information explicitly supported by the window;
names, dates, speaker labels, event order, and serial numbers are preserved.
Each fact is capped at 28 words and each source quote at 12 words.
The raw window is then replaced in the searchable memory index by one
retrieval item per extracted fact. Each item retains the source-window
identifier and token span, and is rendered with the fact fields and
provenance before embedding. Thus, the atomic-fact condition does not
present the original window as reader evidence.

\paragraph{Raw summary.}
For each raw window, a companion summary is generated with five fields:
a dense factual summary, ordered key events, important entities, temporal
or serial-order cues, and conflicts or updates. The searchable item
contains both the complete raw window and this structured summary, so the
summary augments rather than replaces the source text. The item retains
the source-window identifier and token span. This condition therefore
changes the indexed text and its embedding while preserving the raw
evidence in each retrieved item.

\paragraph{Graph-expanded raw.}
The graph nodes are the original raw windows. During ingestion, each new
window is compared by embedding similarity with up to its ten nearest
earlier windows. An LLM then selects any genuinely related candidates and
returns a short reason for each accepted link. Links are accepted only
from this candidate set and are stored as undirected edges, even though
they are created from an earlier window to the new window. At query time,
the question embedding first selects 32 seed windows. The candidate set is
expanded by adding every one-hop graph neighbor of each seed; no positive
candidate cap is applied in the evaluated configuration. The expanded raw
windows, rather than graph descriptions, are then reranked and the top 15
are shown to the reader. Graph cues are disabled in the reader context.
Thus, graph expansion changes candidate recall but preserves reader-visible source text.

\paragraph{Shared configuration.}
The three variants use separate local, conversation-scoped indices.
Ingestion-time memory management is concurrent, with configured maxima of 80, 16, and
32 concurrent calls for atomic facts, raw summaries, and graph-expanded
raw, respectively. Embeddings are requested in batches of 128. The
MemoryAgentBench condition disables display-time turn expansion and treats
each prepared fact as a complete message with no overlap; other benchmark
conditions retain the ordinary display-time policy. The exact prompts and
JSON schemas are reproduced in the prompt appendix,
Appendix~\ref{app:prompt-agentic-mutation-ablations}.

\subsection{Context-budget interventions}
\label{app:budget-interventions}

The two 5k interventions start from the original retrieved items and
alter only their display and packing.
The outer prompt and question are outside this budget.
Appendix~\ref{app:reader-context-budget} reports the results for each
display policy and set of systems.

\paragraph{Prefix-based truncation.}
This separate control freezes the retrieved evidence and applies a
5,000-token ceiling to the formatted evidence, including its headers
and separators, using \texttt{o200k\_base}. It preserves the longest
\emph{prefix of complete ranked entries} that fits. At the first
non-fitting entry it stops.
If the first entry alone is too long, it keeps that entry's first
5,000 tokens. This is the sole partial-entry case. The resulting prefix
is passed to the reader without a truncation marker, with any remaining
budget unused.

\paragraph{Expansion-aware budgeting.}
We additionally freeze the selected top-15 chunks and their ordering
for the main GPT-4o-mini LoCoMo and MEME conditions and the
GPT-4.1-mini LongMemEval condition. Evidence already within 5,000
\texttt{o200k\_base} tokens is left verbatim. For each over-budget
question, we first disable turn expansion for \emph{all} selected
chunks and display their original token-window text, including
headers and separators. If this still exceeds the ceiling, the
fallback is the longest prefix of complete entries in the existing
display order, with token-prefix clipping only when the first entry
alone exceeds the budget. This fallback uses the same prefix rule as
the six-system intervention, including omission of a truncation marker.
MEME uses chronological order; LoCoMo and LongMemEval use
relevance order. No deduplication or task-specific readout instruction
is added, and the ceiling excludes the outer prompt and question.

\subsection{Complete-store and oracle evaluations}
\label{app:complete-store}

The following diagnostic conditions provide source-oracle evidence or gold-informed
access to the complete store and its raw control.

\paragraph{Answer-location source oracle.}
The source oracle in Table~\ref{tab:meme-oracle} supplies all annotated
supporting sessions at or before the question's cutoff, in chronological
order. Selection uses neither the question, gold answer, nor entity values
to choose among those sessions, and applies no token cap or truncation.
Non-Abs tasks use the standard reader prompt; Abs alone uses the value-free
dependency-validity instruction in Appendix~\ref{app:prompt-abs-validity}.

\paragraph{Complete-store access.}
We denote native retrieval by $N$, the complete-store condition by $C$,
and its raw-evidence control by $R$. The complete-store condition replaces
native retrieval with gold-informed selection over exported persisted
representations. Its raw control filters
phase-visible annotated source sessions using the same gold fields.
Both pack evidence within a nominal 5,000-token context budget and append
the task policy in Appendix~\ref{app:prompt-complete-store}.
That policy supplies phase, task type, target identifiers, and transition
semantics to every task. In particular, its after-phase Abs branch states
that the upstream entity changed without a replacement target.
The policy is value-free and supplies this task/status information explicitly.
Compared with the answer-location oracle, selection, context assembly,
budget, and readout instructions all change.

\paragraph{Gold-informed selection.}
The selector matches each question to its annotated task and collects
target-entity before values, plus after values and history for after-phase
questions; it also collects before/after values of a cascade source
when present, the question's annotated entity values, the expected
answer, and target/cascade entity names with underscores replaced
by spaces. Strings are lowercased, normalized to word sequences,
and deduplicated. For a stored object $m$, let $H(m)$ be the set
of normalized phrases occurring as substrings and let $Q$ be the
question's word set. The memory-object ranking is
\begin{equation}
 s(m)=\sum_{h\in H(m)}\left(4+\min(4,|h|/3)\right)
      +2\frac{|\operatorname{words}(m)\cap Q|}{\max(1,|Q|)},
\end{equation}
where $|h|$ counts words. No retrieval-similarity term contributes
to this complete-store ranking. Objects without a phrase hit or
query-word overlap are not selected; ties in descending score are
resolved by the order in which objects were enumerated from the store.
Gold values and answers guide this offline selector only. The reader's
context contains the selected memory text and the task policy.

\paragraph{Temporal eligibility and provenance.}
``Complete'' refers to enumerating persisted reader-visible
representations. Table~\ref{tab:complete-store-exports} specifies each export.
For provenance-filtered stores, a candidate must have a nonempty
mapped source-session set whose maximum index does not exceed the
question cutoff. Normalized duplicate texts are removed, keeping the
first exported occurrence. EMem instead uses its actual before/after
snapshot and deduplicates within that snapshot. Raw provenance text
is used for provenance checks or session mapping, never substituted for an
extracted memory. Exported model-control-token literals are escaped as text.

\begin{table}[htbp]
  \centering
  \caption{Stored evidence and temporal eligibility rules for the seven
  systems in the MEME complete-store diagnostic. The complete store
  comprises the listed reader-visible representations; source provenance
  or before/after snapshots determine eligibility at the question's
  temporal cutoff. EDUs are elementary discourse units.}
  \label{tab:complete-store-exports}
  \footnotesize
  \setlength{\tabcolsep}{3pt}
  \renewcommand{\arraystretch}{1.15}
  \begin{tabular}{@{}>{\raggedright\arraybackslash}p{0.13\linewidth}
      >{\raggedright\arraybackslash}p{0.40\linewidth}
      >{\raggedright\arraybackslash}p{0.43\linewidth}@{}}
    \toprule
    System & Exported reader evidence & Provenance and scope \\
    \midrule
    Mem0\newline \citep{chhikara2025mem0} &
    All stored textual memories, excluding the auxiliary entity
    graph. &
    Match each memory to its conversation and map its creation time
    (or recorded timestamp if unavailable) to the source session. This is the
    final local mutable store. \\
    \addlinespace
    MemOS\newline \citep{li2025memos} &
    All memories in the fine-grained stores used by the Native
    condition, including preference, tool, and skill memories. &
    Map source-message timestamps to source sessions.
    Source-message content is excluded from reader text. \\
    \addlinespace
    MemU\newline \citep{nevamind2025memu} &
    All item summaries and any nonempty linked category summaries
    in the local stores; no original resource text. &
    Resource timestamps map items to sessions, including
    segmented resources. Category-item links supply the union of source
    sessions; a category is visible only after all mapped sources.
    Empty category summaries are checked but excluded. \\
    \addlinespace
    EMem\newline \citep{zhou2025simple} &
    All elementary discourse units (EDUs) in the 200 stored
    before/after snapshots. &
    Use the largest completed snapshot when multiple copies exist.
    Snapshot boundaries determine visibility; graph attributes map EDU
    IDs to source sessions for display order. No source transcript is
    appended. Task-specific packing is described below. \\
    \addlinespace
    EverMemOS\newline \citep{hu2026evermemos} &
    Every episode text and atomic-fact text in the per-conversation
    persisted vector-index files. &
    Map original-message timestamps to source sessions;
    original-message content is excluded. Historical versions
    are unavailable in this export. \\
    \addlinespace
    MemMachine\newline \citep{wang2026memmachine} &
    All episodic and semantic memories,
    plus persisted working-memory summaries. Render episodic text with
    timestamp/speaker and semantic text as category/name/value. &
    Use episodic source metadata and database citation episodes.
    Uncited semantic features and current summaries conservatively
    depend on all conversation sessions. Citations serve only to locate
    source episodes. \\
    \methodrule
    \textbf{\rawrag{}} &
    All stored 256-token chunks with 32-token overlap;
    embeddings are excluded from reader evidence. &
    Chunk turn ranges map through turn timestamps to MEME sessions.
    Exclude chunks crossing the cutoff; never crop a chunk or expand it
    into unselected source turns. \\
    \bottomrule
  \end{tabular}
\end{table}

\paragraph{Snapshot limitation.}
Except for EMem's explicit phase snapshots, exports reflect the final
store and are filtered using source metadata. Overwritten or deleted
objects are unavailable, and surviving objects may contain later revisions.
Mem0 is particularly subject to this limitation because it maps object
creation time.
Missing or unmappable required provenance invalidates an export
unless an explicit conservative rule applies; the
MemMachine rule above is an explicit exception for uncited
consolidated objects. ``Phase-visible'' refers to visibility under these
export and filtering rules.

\paragraph{Ordering and truncation.}
The diagnostic uses \texttt{o200k\_base} token counting and reserves
the task-policy tokens before evidence packing. The nominal 5,000-token
budget covers formatted evidence and policy; the outer reader prompt and
question are excluded, and the joining newline is added afterward. Token
counts refer to the formatted context actually supplied to the reader.
The following packing rules are part of the reported conditions:
\begin{enumerate}
  \item \textbf{Ordinary memory objects.} Mem0, MemOS, MemU,
  EverMemOS, and MemMachine present separately delimited memory blocks in
  gold-relevance order. If the first block
  exceeds the available evidence budget, retain its token prefix;
  subsequently skip any non-fitting block and continue to later
  candidates. Headers count toward the budget.
  \item \textbf{RAG whole chunks.} Add one point to $s(m)$ when a
  chunk overlaps an annotated phase-visible evidence session, and
  admit location-matching chunks even without lexical hits.
  Greedily select fitting chunks by new phrase coverage, then score,
  then enumeration order; finally display them in source-text order.
  Skip oversized chunks without truncating them.
  \item \textbf{EMem.} Agg and Del use the ordinary relevance-order
  packer. ER keeps the full ranked list but packs whole EDUs and then
  restores chronology. Tr, Cas, and Abs first select a compact set
  covering each available gold-value phrase once, add at most two
  target/cascade-label EDUs, and fall back to the first three ranked
  EDUs if none were selected. Pack fitting whole EDUs, then sort by
  earliest source session and enumeration order; EDUs without
  session provenance appear last. This chronological packer skips
  non-fitting EDUs rather than truncating them.
  \item \textbf{Raw control.} Among annotated sessions at or before
  the cutoff, retain sessions with at least one gold-phrase hit;
  if none match, fall back to all visible annotated sessions.
  Display selected sessions chronologically. Truncate an oversized
  first session to the evidence budget; skip later non-fitting
  sessions and continue. This differs both from untruncated
  answer-location readout and from RAG's indivisible chunk packing.
\end{enumerate}

\paragraph{Budget relaxation.}
The no-cap intervention removes only the 5,000-token stopping condition from
the packers above.  It preserves each condition's exported candidate set,
gold-informed relevance scores, admission rules, display order, task policy,
reader model, and scoring procedure.  Thus every admitted candidate appears
in the same final display order; the raw control is unchanged because all of
its selected evidence already fits.  Reader inputs that are byte-identical to
the 5,000-token condition reuse the corresponding answer and judgment; changed
inputs are answered with the same GPT-4o-mini snapshot, and identical answer
strings reuse judgments.  Before-phase gates are recomputed over the full
paired set before after-phase task scores and affected-subset scores are
formed.  The resulting contrasts are reported in
Table~\ref{tab:complete-store-uncapped}.
All accepted new reader responses identify the GPT-4o-mini-2024-07-18
snapshot; responses identifying a different model are discarded and
the original requests are retried. Each accepted request is checked
against its prepared input fingerprint, and its server-reported prompt
token count must match the full prepared prompt. No condition exceeds
the context window. For MemMachine, 16 unchanged inputs reuse historical
results and 1,172 inputs receive new reader responses. Intervals use
20,000 paired episode-bootstrap resamples, conditional on the observed
outputs.

\paragraph{Interpretive limitations.}
\label{app:complete-store-boundary}
The raw-to-store gap reflects both stored information and the process used
to present it to the reader. Export coverage and temporal provenance define
the candidate evidence; selection, ordering, packing, and reading determine
how that evidence is used. The comparison therefore does not isolate
information loss during ingestion. Native-to-complete gains likewise combine
broader access to stored evidence with changes to task information and
readout policy. Relations missed by native retrieval may remain in the store.

Task scores in Appendix~\ref{app:complete-store-results} show where the
aggregate comparison needs care: raw evidence scores below 100\%,
and system rankings vary by task. On MEME Abs, missing evidence
can elicit an ``uncertain'' answer that receives credit, so a correct answer
does not necessarily indicate successful dependency reconstruction.

\subsection{Statistical inference}
\label{app:statistical-methods}

All procedures below operate on paired outcomes from the stated completed
conditions. The resampling unit follows the benchmark structure and keeps
repeated judgments of each answer together. Reported intervals cover
question/episode sampling uncertainty conditional on the observed reader
and semantic-judge outputs, store builds, and sufficiency labels, including those
from the deterministic MEME verifier.

\paragraph{Coverage and utilization.}

Table~\ref{tab:coverage-omnibus} gives the four omnibus tests underlying
Section~\ref{sec:mechanism}.  MemoryAgentBench coverage uses Cochran's $Q$ on
the seven paired binary sufficiency outcomes, followed by exact paired McNemar
tests. Its intervals use 50,000 source-stratified paired bootstrap samples.
Because utilization has a system-specific denominator, its omnibus and
pairwise tests instead use a centered max-range bootstrap and two-sided
bootstrap sign tests. The MemoryAgentBench resampling shares question indices
across systems within each of the 14 sources. The MEME analyses resample 100
whole episodes with shared indices across systems, including all questions and
before/after pairs from each sampled episode. They use 50,000 replicates. In
every case, Holm adjustment is applied over the 21 pairwise comparisons within
one benchmark--metric family.

\paragraph{Retrieval-design ablations.}
The MemoryAgentBench design ablations use 20,000 paired,
source-stratified bootstrap resamples and Holm correction, as reported
in Section~\ref{sec:ablation}. Source scores are recomputed under the
benchmark's capability-weighting scheme rather than pooling all questions.
Final-$k$ intervals compare each condition with the default $k=15$;
capability-specific contrasts use their stated task family.

\paragraph{Complete-store contrasts.}

For each contrast, 20,000 paired bootstrap replicates sample 100
episodes with replacement, keeping all questions and both phases
of an episode together. Each replicate recomputes the six-task
macro-average in both conditions and then their difference; the
2.5th and 97.5th percentiles give the interval. The $R-C$ intervals
use the same after-phase scores as Figure~\ref{fig:complete-store-loss}.
The $C-N$ intervals
use those same complete-store scores and the aligned Native scores.
These are pointwise intervals, without multiplicity correction,
conditional on the observed reader and judge
outputs and store builds. RAG Native aggregates three judge calls,
while its complete-store and raw controls use one.

Question identities and after-phase scores are aligned across the
compared conditions. The original random seeds for the $R-C$
bootstrap are unavailable, so new resamples may yield slightly different
interval endpoints.

\FloatBarrier

\section{Experimental Implementation and Reproducibility}
\label{app:reproducibility}

This appendix specifies the experimental settings needed to reproduce
our evaluations, including benchmark inputs and task definitions,
baseline memory construction and retrieval configurations, and shared
evaluation procedures. We also document scoring and averaging rules,
timing and readiness criteria, and the software and hardware environment.

\subsection{Benchmark descriptions}
\label{app:benchmark-details}

Alongside the four main benchmarks, we describe PersonaMem-v2 for the
supplementary accuracy comparison and BEAM for ingestion timing only.

\subsubsection{LoCoMo}

\paragraph{Data and tasks.}
LoCoMo~\citep{maharana2024locomo} supplies long, multi-session
conversations between two named speakers, with session dates, dialogue
identifiers, and, for some turns, images and textual captions. A question
refers to the history of one conversation and requests a free-form answer.
We use the 10-conversation release, with 1,540 questions after
excluding adversarial category 5. The remaining types are single-hop
(category 4, 841 questions), multi-hop (category 1, 282), temporal
(category 2, 321), and open-domain (category 3, 96).

\paragraph{Capability requirements.}
Single-hop questions test precise recovery of an entity or fact from a
particular interaction. Multi-hop questions require combining facts
distributed across turns or sessions, including resolving who said or
experienced what. Temporal questions require anchoring relative dates
to the conversation timeline and reasoning about event order.
Open-domain questions connect conversational evidence with broader
knowledge or reasonable inference. The shared difficulty is finding
sparse evidence in a long, personally grounded history while preserving
speaker attribution, names, and time references.

\paragraph{Evaluation setup.}
Each dialogue history forms one memory store shared by its questions.
Session boundaries and speaker attribution are preserved.
Images are represented by the supplied caption in message
text. All included questions reuse their
conversation's memory store; question category and reference evidence
are reserved for evaluation. The
different published judge and lexical-metric conventions remain separated
in Appendix~\ref{app:results}.

\subsubsection{LongMemEval-S}

\paragraph{Data and tasks.}
Each LongMemEval-S example~\citep{wu2025longmemeval} contains a
user--assistant question, its question date, and a dated collection of
historical chat sessions containing both relevant and distractor
interactions. Our standalone evaluation uses 500 examples. Its six
reported question types are \emph{single-session-user} (recall a
user-provided fact), \emph{single-session-assistant} (recall an
assistant-provided fact), \emph{single-session-preference} (use a
preference expressed within a session), \emph{multi-session}
(combine evidence across sessions), \emph{knowledge-update}
(select the valid information after a change), and \emph{temporal-reasoning}
(reason about dates, intervals, or order). Unanswerable/abstention
examples form an additional evaluation condition across these types.

\paragraph{Capability requirements.}
Unlike LoCoMo, where multiple questions share a small fixed set of histories,
LongMemEval supplies a separate conversation history for each example. The benchmark tests
whether a system can distinguish user from assistant knowledge, recover
preferences, integrate dispersed evidence, resolve superseded facts,
and avoid answering when the history lacks support. The question date
is the reference time for phrases such as ``current'' or ``last month''.

\paragraph{Evaluation setup.}
Each example has an isolated memory store that preserves session dates
and speaker roles. Annotated answer sessions are used only for
evaluation. The reader receives the question-date instruction
shown in Appendix~\ref{app:prompt-longmemeval-reader}. Published
non-abstention scores use their specified question subsets;
our full-set scores use all 500 examples.

\subsubsection{MEME}

\paragraph{Data and tasks.}
MEME~\citep{jung2026meme} supplies chronological episodes in personal-life
and software-project domains. Each contains evidence sessions,
distractor sessions, entity values, and explicit dependency rules,
with questions before and after a designated change. We use the
condition with 32k tokens of distractor content: 100 episodes and 1,188 phase-specific
questions. Its six types distinguish different operations:
\begin{itemize}
  \item \textbf{Exact Recall (ER):} reproduce a stable entity value.
  \item \textbf{Aggregation (Agg):} collect several entities' required
  values from distributed evidence.
  \item \textbf{Tracking (Tr):} reconstruct an entity's sequence of
  historical values in chronological order.
  \item \textbf{Deletion (Del):} acknowledge removal without revealing
  the deleted value.
  \item \textbf{Cascade (Cas):} apply a stated conditional rule after
  an upstream change to derive a dependent entity's new value,
  including one- and two-hop dependencies.
  \item \textbf{Absence (Abs):} recognize that a dependent old value
  is no longer trustworthy when an upstream condition changes and
  no applicable replacement value is supplied.
\end{itemize}

\paragraph{Capability requirements.}
Fictitious values reduce the usefulness of parametric factual knowledge.
The task crosses single- versus multi-entity reasoning with static
versus evolving information. In particular, Cas requires executing an
update rule, whereas Abs requires withholding an unsupported current
value. Del evaluates whether the answer respects a removal request;
physical erasure from storage is outside its scoring criteria.

\paragraph{Evaluation setup.}
Questions carry their episode, before/after phase, target entities,
dependency depth, and temporal cutoff. All systems receive the official
short-answer prompt. Scoring uses task-specific judgments:
Tr checks the ordered history, Agg checks the required
values, and Cas, Abs, and Del have different correctness conditions.
Task scores use after-phase answers. For Cas, Abs, and Del, an
after-phase pass receives credit only if the corresponding before-phase
answer was also correct, preventing a system that never knew a value
from receiving credit merely for failing to report it. Overall is the
six-task macro-average.
Phase visibility is a requirement of each run; retrieval-time timestamp
filtering and the explicit phase-scoped oracle controls are distinguished
below and in Appendix~\ref{app:method}.

\subsubsection{MemoryAgentBench}

\paragraph{Data and tasks.}
MemoryAgentBench~\citep{hu2026memoryagentbench} presents a long source
context as a sequence of memory-ingestion units, followed by multiple
questions over the same context. The evaluated suite has 2,071
questions from 14 sources, organized into four capabilities:
\begin{itemize}
  \item \textbf{Accurate Retrieval (AR):} LongMemEval chat-history
  questions, EventQA next-event questions over narrative evidence,
  and the two RULER QA sources over long document collections.
  These test locating the correct conversational fact, event, or
  document-supported answer amid distractors.
  \item \textbf{Test-Time Learning (TTL):} five in-context classification
  sources (Banking77, CLINC150, NLU, TREC-coarse, and TREC-fine)
  supply labeled demonstrations and ask for a numerical label for a
  new input. ReDial supplies recommendation dialogues and asks for
  movie recommendations. These test remembering and applying examples
  or preferences acquired during ingestion, with model weights fixed.
  \item \textbf{Long-Range Understanding (LRU):} DetectiveQA asks for
  answers requiring narrative-level reasoning, while InfBench asks
  for book summarization. These test integrating globally distributed
  clues or salient content.
  \item \textbf{Conflict Resolution (CR):} single-hop and multi-hop
  FactConsolidation supply numbered, potentially conflicting facts.
  A larger serial number denotes newer information. Questions require
  selecting the latest valid fact and, for multi-hop cases, composing
  relations under that rule instead of relying on real-world knowledge.
\end{itemize}

\paragraph{Capability requirements.}
The ``ingest once, query many'' structure measures how a reusable
memory representation supports heterogeneous downstream tasks. Exact
example-to-label associations matter for TTL; narrative coverage
matters for LRU; provenance and precedence matter for CR. Compression
that benefits one capability can remove precisely the information
needed by another.

\paragraph{Evaluation setup.}
Each source context forms one shared memory store, using that source's
ingestion units and question wording. Splitting the source into ingestion
units is distinct from the subsequent retrieval-window construction
in \rawrag{}. Original LongMemEval session dates
are preserved where available. For otherwise undated sources,
assigned timestamps encode source order;
FactConsolidation precedence comes from its original serial numbers.
Scoring follows each source's task: substring matching, numerical-label
or choice accuracy, recommendation recall, or LLM-based judging and
summary scoring. The aggregate combines these native task metrics using
the nested capability weights specified below. The embedded LongMemEval
source uses separate evaluation settings from the standalone
500-example benchmark.

\subsubsection{PersonaMem-v2}

\paragraph{Data and tasks.}
Our supplementary evaluation uses the PersonaMem-v2 text
benchmark~\citep{jiang2025personamemv2} and 32k chat
histories, with 5,000 multiple-choice questions. Each input pairs a
personalized user request with candidate responses. The nine scenario
types are consultation, personal email, translation, professional email,
creative writing, professional writing, knowledge query, social media,
and chat.

\paragraph{Capability requirements.}
The task asks whether a response fits the user's preferences and
circumstances from past interactions. Preference attribution and updates
can determine which plausible response is appropriate,
testing personalization grounded in conversational evidence.

\paragraph{Evaluation setup.}
Questions referring to the same 32k history share a memory store.
System-role persona messages are excluded from the ingested dialogue;
user and assistant turns are included.
The persona name can appear as a speaker label; full persona
profiles are excluded from \rawrag{}'s memory evidence.
The reader receives the question and lettered response options;
accuracy is exact agreement between the selected and reference letters.
Preference annotations and reference snippets are reserved for evaluation.
Option order and the corresponding answer letters must remain fixed
across compared runs. Table~\ref{tab:personamem-v2} reports the results
under this multiple-choice evaluation condition.

\subsubsection{BEAM ingestion workload}

BEAM~\citep{tavakoli2026beam} supplies the long conversation used for
the scaling study. Our input is one ordered, 1,710-message conversation
approaching one million tokens. Each message is added before
the next timed Add operation begins. We use BEAM solely to measure ingestion
throughput, readiness cost, and persistence under a growing history.

\subsection{Evaluation workflow}
\label{app:integration-contract}

Each experiment separates four operations: ingesting the available
history, retrieving evidence for a question, generating an answer,
and scoring that answer. Histories preserve speaker attribution and
available dates. Questions and reference answers remain separate from
memory construction; references are used only for evaluation and the
explicitly identified diagnostic controls. Each system uses its own
memory construction and retrieval procedure. The selected evidence is
then presented to a reader under the stated prompt condition
and scored according to the benchmark.

The configurations below specify system-specific memory units, speaker
scope, and retrieval budgets; comparison scope is summarized in
Appendix~\ref{app:results}.

\paragraph{Temporal visibility.}
Where supported, retrieval uses the question date to exclude later
memories. Several systems first retrieve a larger candidate set, filter
it by timestamp, and then apply the final evidence budget; memories
with unknown timestamps are kept by default.
A transformed item can combine evidence from multiple times.
EMem has explicit question-date-scoped indexing for applicable
datasets, while oracle analyses use their specified phase-visible
sources or snapshots. Timestamp filtering alone does not establish temporal
validity for revised memories or undated semantic features; their visibility
also depends on how each system constructs and updates its store.

\subsection{Baseline configurations}
\label{app:baseline-details}

The configurations below specify each agentic memory system's
memory-management operations and ordinary execution settings, alongside
the \rawrag{} and Full-context reference conditions.
Concurrency denotes a configured upper bound. Conversation-level
ingestion, concurrent queries, and internal model calls are different
levels of parallelism. Benchmark-specific conditions are identified
separately; in particular, the timing experiments use the single-worker
setup described below. The shared answer stage defaults to 50
concurrent requests unless a method or run overrides it.

\subsubsection{\rawrag{}}

The source-preserving representation, retrieval and reranking procedure,
benchmark-specific evidence assembly, and query construction are defined
in Appendix~\ref{app:method}. Ablation conditions change only their stated
operation while keeping the corresponding benchmark inputs and scoring
procedure. Here we specify the execution configuration.

\paragraph{Execution settings.}
The default limits are two concurrent
conversations during indexing, eight search requests, and 50 answer
requests. Embeddings are requested in batches of 128. The default
store is a local index maintained separately for each conversation.

\subsubsection{EverMemOS}

\paragraph{Memory construction.}
EverMemOS constructs organized episodic memories and associated
structured information~\citep{hu2026evermemos}. Histories are segmented
into coherent memory units called MemCells, from which episodic
representations and atomic facts are extracted. Lexical and embedding
indices support retrieval. We also enable incremental profile extraction.

\paragraph{Retrieval and readout.}
Retrieval combines lexical and semantic search with query refinement
and checks for evidence sufficiency. By default, 10 results are
returned and presented as subject--episode blocks.
Answer generation follows the system's own procedure,
with benchmark-specific prompts where specified.

\paragraph{Execution settings.}
EverMemOS runs locally.
Its extraction concurrency is 20, with a separate
atomic-fact concurrency limit of 20. Index construction permits five
concurrent embedding batches. Agentic retrieval
limits concurrent questions to 20, and its native response stage
allows 50 concurrent requests.

\subsubsection{MemOS}

\paragraph{Memory construction.}
MemOS manages extracted text memories and preference information
within its memory operating-system abstraction~\citep{li2025memos}.
We use its fine-grained extraction setting and wait for each ingestion
step to finish before proceeding. Memories are extracted and stored
by MemOS itself.

\paragraph{Retrieval and readout.}
Retrieval requests the top 20 text memories and preferences using
the \texttt{fine} search mode, reranking, and maximal-marginal-relevance
deduplication. We choose \texttt{fine} rather than the default
\texttt{fast} mode to evaluate MemOS's memory-management capabilities.
For two named conversational speakers, the system writes
and searches separate speaker perspectives and formats both sets of
memories, including returned preferences, for the reader. In
MemoryAgentBench it instead uses a context-scoped identity. For its
ICL sources, the reader additionally receives up to five original
label-mapping chunks selected by lexical overlap with the question.
These chunks supply raw labeled demonstrations from the ingested context.

\paragraph{Execution settings.}
We use locally hosted MemOS~2.0.23. Ordinary
evaluation allows 10 conversation workers and 10 concurrent searches.
Add batches contain 10 messages, reduced to six for LongMemEval and
two for MemoryAgentBench, with a 0.1-second request interval.
Storage uses Qdrant~1.15.3 and Neo4j, with version~5.26.6 specified
and version~5.26.4 permitted as a fallback.

\subsubsection{EMem and EMem-G}

\paragraph{Memory construction.}
EMem represents conversations using extracted elementary discourse
units (EDUs); EMem-G additionally uses graph-based
retrieval~\citep{zhou2025simple}. We use contextual extraction of EDUs
without a further context-generation step. Session and turn structure
is preserved. Where question dates define different visible histories,
separate stores can be built from the corresponding history prefixes.

\paragraph{Retrieval and readout.}
EMem retrieves without personalized PageRank, whereas EMem-G uses it
to propagate relevance through the graph; the two variants are distinct.
The default QA budget is 10, overridden to 20 for MEME and
MemoryAgentBench. Retrieved EDUs, or session documents when EDUs are
unavailable, provide the reader evidence.
For ICL it also prepends up to five lexical-overlap-selected raw
demonstration chunks, as in MemOS. MEME and MemoryAgentBench can use
the aligned benchmark reader over this evidence; other conditions
use EMem's own answer-generation procedure. These answer conditions
are distinguished in the experiment records.

\paragraph{Execution settings.}
EMem processes conversations and date-scoped histories sequentially
with internal batched model calls; it allows up to
50 in-flight requests. Embedding batches contain up to 256 items,
and answer generation uses up to 32 workers.
The random seed is 42. Extracted memories and indices are reused
only when inputs and experimental settings match.

\subsubsection{SimpleMem}

\paragraph{Memory construction.}
SimpleMem converts dialogue windows into compact, structured memory
entries and combines several retrieval channels~\citep{liu2026simplemem}.
Our direct evaluation covers LoCoMo. Each conversation has a separate
memory store, built from timestamped windows of 40 messages with a
two-message overlap. The final partial window is also included.

\paragraph{Retrieval and readout.}
Retrieval combines semantic, keyword, and
structured retrieval, with configured channel budgets of 25, 5, and
5, planning enabled, and up to two reflection rounds. We preserve
the merged native order and pass at most 20 entries to the reader,
using each entry's factual restatement and timestamp. Embeddings
follow the experiment's stated backbone. Final answering uses the
shared benchmark reader and prompt. The published SimpleMem
LongMemEval result is attributed to its source separately.

\paragraph{Execution settings.}
Two conversations may be ingested concurrently, and search uses four
query workers. Within a query, parallel retrieval is enabled with
up to three channel workers. Each conversation has its own LanceDB
store. Search has a 600-second timeout and two configured attempts;
the shared reader has a separate 100,000-token context cap.
These settings apply to our LoCoMo evaluation.

\subsubsection{MemU}

\paragraph{Memory construction.}
MemU extracts profile, event, knowledge, and behavior items from
interaction resources~\citep{nevamind2025memu}. Each conversation is
treated as a separate user's history, and its messages are ingested
in order.

\paragraph{Retrieval and readout.}
We use the default configuration of MemU~1.5.1.
Retrieval stays within the same conversation and includes intention
routing and evidence-sufficiency checks. Retrieved items are supplied to
the shared reader. The final result cap is 20 by default and 10 for
MemoryAgentBench; internal candidate retrieval has a separate budget.

\paragraph{Execution settings.}
We use MemU~1.5.1. Ingestion permits five
concurrent conversations but preserves the order of writes within
each conversation. Ingestion batches contain up to 128 messages, with
one message per MemU resource. Persistence uses PostgreSQL with
pgvector. Search batches up to eight questions at a time and reuses
a conversation-scoped GPU item index for exact cosine similarity
on the GPU; index loading fetches 512 items per batch.
GPU similarity computation is separate from API-based embeddings
and LLM routing and sufficiency checks.

\subsubsection{MemMachine}

\paragraph{Memory construction.}
MemMachine combines source-grounded episodic memory with extracted
semantic memory~\citep{wang2026memmachine}.
Each conversation has an isolated store with both memory types
enabled. Semantic extraction may continue after message ingestion,
so ingestion completion and readiness for retrieval are distinct
timing boundaries.

\paragraph{Retrieval and readout.}
We use the default configuration of MemMachine~0.3.4.
Episodes, episode summaries, and semantic features appear in separate
context sections. Episodic and semantic lists have
independent top-20 budgets and keep their separate rankings. Episodic
candidates are date-filtered, but undated semantic features bypass the
filter. The date cutoff therefore does not establish temporal validity
for semantic evidence. The formatted sections are sent to the
shared reader for final answer generation.

\paragraph{Execution settings.}
We use MemMachine~0.3.4 with model access through OpenRouter.
Storage uses PostgreSQL~16 with pgvector and
Neo4j~5.23 Community. Ordinary evaluation configures 10 ingestion
workers and 10 search workers, with 10-message Add batches
(two for MemoryAgentBench). Ordinary evaluation uses a
1,800-second post-Add wait. Timing runs instead wait for semantic
ingestion to finish.

\subsubsection{Mem0}

\paragraph{Memory management.}
Classic Mem0 extracts candidate facts,
retrieves related existing memories, and chooses ADD, UPDATE, DELETE,
or NONE operations~\citep{chhikara2025mem0}. Message batches are ingested
chronologically into separate conversation stores, with memory changes
recorded for the mutation audit. This also defines the unmodified-policy control.

\paragraph{Retrieval and readout.}
We use the default configuration of each evaluated Mem0 version.
For two named speakers, the system maintains separate
perspectives and queries both; MemoryAgentBench instead uses one
context-scoped identity. Native memory search supplies the text
facts, with a configured top-20 budget per searched identity.
Retrieved facts undergo temporal filtering
before being supplied to the shared reader. The extraction model
is selected separately from the answer model; the default reader
uses the structured reasoning prompt in Appendix~\ref{app:prompts}.
The ADD/NONE-only policy converts proposed UPDATE to ADD and DELETE to NONE.

\paragraph{Execution settings.}
The general local evaluation uses Mem0~2.0.11, including the
MEME store used in the evidence analyses, whereas the classic
mutation audit and ADD/NONE intervention use Mem0~1.0.11.
Ordinary evaluation permits five
conversation workers and five search workers, with five-message
Add batches (two for MemoryAgentBench) and an 8,000-character
per-message limit. Writes remain sequential within a conversation;
the sharded configuration uses separate on-disk Qdrant and SQLite
files per conversation to avoid concurrent-writer conflicts.
Model requests have a 60-second timeout and no HTTP-client retries.
Failed ingestion stages can be retried, and timeout-triggered
splitting can further subdivide a message batch.

\subsubsection{Memora}

\paragraph{Memory construction.}
Memora separates stored memory values from indexed primary abstractions
and cue anchors, linking compact retrieval representations to more
specific factual and episodic content~\citep{xia2026memora}.
Each conversation has an isolated store. Messages are grouped by
session and segmented before memory extraction, with source
attribution preserved. MemoryAgentBench skips this extra segmentation
and ingests its prepared source chunks directly.

\paragraph{Retrieval and readout.}
Search uses a prompted-policy retriever with at most five steps.
The evaluated configuration
enables cue indexing, BM25 hybrid search, and episodic memory, with
top-20 output before applicable temporal processing. For each
returned factual memory, the reader receives its value and history
together with linked episodic memories, supplying both specific
conversation evidence and factual details to the common reader.
The answer context has a separate 100,000-token cap. This condition
combines Memora's memory-management pipeline with a shared reader;
the published policy-retriever scores are attributed separately.

\paragraph{Execution settings.}
Ordinary evaluation permits 10 ingestion workers and 10 search workers,
with a separate limit of two segmentation workers; resumed runs can
reduce conversation concurrency. Each conversation uses an isolated
Chroma persistent directory. LLM-based segmentation determines the
ordinary ingestion units; the configured two-message batch size
applies to non-LLM segmentation. MemoryAgentBench disables the extra
segmentation and ingests one prepared source chunk at a time.
Extraction inputs are bounded to 10,000 characters and can be
recursively split down to 1,000 characters after output-length
failures. The model-request timeout
is 600 seconds with three retries; extraction uses seed 42
and zero temperature.

\subsubsection{Full-context}

Full-context omits both memory transformation and retrieval: the
reader receives the visible chronological history directly, filtered
by question date when supplied, under the bounded-history policy in
Appendix~\ref{app:native-context-policy}.
Published Full-context results retain their original settings.

\paragraph{Execution settings.}
No memory transformation or vector index is used.
History preparation uses four workers
and 10 answer workers. These limits and the context truncation rule
apply only to our direct evaluation.

\subsection{Experimental controls}

We save the reader context, answer, and score for each question to
distinguish retrieval, generation, and judging changes under the
conditions defined in Appendix~\ref{app:results}.

\paragraph{Evaluation comparability.}
A complete evaluation requires the expected question count and a unique
identifier linking each question to its context, answer, and judgments
within a source. For MEME, evidence must also respect the question's phase.
Two implementation choices affect which condition is being evaluated:
falling back to dense ordering omits reranking, and changing chunking or
embeddings requires a rebuilt index. Interpreting a score also requires
the dataset version, model identities, experimental settings,
and source-specific exclusions.

\subsection{Scoring and benchmark-level summaries}
\label{app:metric-aggregation}

\paragraph{Benchmark weights.}
LoCoMo averages correctness over the 1,540 included questions and
the specified judge repetitions. LongMemEval averages over its 500
examples under the stated judging procedure. MEME reports the macro-average
of the six task scores. PersonaMem-v2 uses question-weighted
multiple-choice accuracy over 5,000 questions.
MemoryAgentBench first computes source scores and then the four
capability scores. In Test-Time Learning, the five ICL source scores
are averaged before being combined with ReDial:
\begin{equation}
  \mathrm{TTL}=\frac{1}{2}
  \left(\mathrm{ReDial}+\frac{1}{5}\sum_{j=1}^{5}\mathrm{ICL}_j\right),
  \qquad
  \mathrm{Overall}=\frac{\mathrm{AR}+\mathrm{TTL}+\mathrm{LRU}+\mathrm{CR}}{4}.
\end{equation}
Accurate Retrieval, Long-Range Understanding, and Conflict Resolution
average their constituent source scores. A flat average over all
questions or over the six TTL sources would assign different weights.
Full-benchmark results require complete source coverage.

\subsection{Latency measurement}
\label{app:timing-protocol}
Sequential-ingestion measurements record the duration of each message's
ingestion. The BEAM comparison includes complete runs
over the same 1,710-message conversation, and cumulative Add time is
the sum of the recorded message-level durations. Both conversation-level
and outer ingestion concurrency are set to one:
each single-message Add finishes before the next starts, with native
internal parallelism unchanged. For MemMachine,
the timer also includes a semantic-ingestion readiness barrier,
polled every second with a one-hour timeout; its sequential-run
configuration triggers semantic ingestion after one message rather
than waiting for a multi-message batch.

The separate MEME latency experiment measures Add-to-ready and Search
over 10 conversations and 110 questions, with both outer conversation
and search concurrency set to one. MemOS uses synchronous Add;
MemMachine readiness is determined by an empty pending-ingestion
queue rather than the ordinary fixed post-Add wait.
Total latency sums the unrounded Add-to-ready and Search times, excluding
overhead outside these stages; the displayed entries are rounded independently.
Neither timing experiment includes reader generation or evaluation.
Both run on the host described in Appendix~\ref{app:environment}.
Within the measured stages, wall-clock time includes local computation
and storage, API-service latency, network communication, and retries.
The comparison is influenced by each system's readiness rules and internal
parallelism, as well as the available CPU, memory, disk, container
resources, and endpoint load.

\subsection{Computational environment}
\label{app:environment}

\paragraph{Deployment configuration.}
All experiments conducted in this study, including the latency and
sequential-ingestion experiments, were orchestrated on the same local
machine. Evaluation processes, local memory stores, and the self-hosted
MemOS and MemMachine services ran on that machine. LLM, embedding, and
reranking inference used API endpoints. Published comparison scores were
obtained in the environments described by their respective studies.

\paragraph{Hardware.}
The host has an Intel Core i9-13900HX CPU with 24 physical cores and
32 logical processors, 16~GiB of installed RAM, an NVIDIA GeForce
RTX~5070~Ti Laptop GPU with approximately 12~GB of dedicated memory,
and a 1~TB YMTC PC411 NVMe SSD. The recorded host software is 64-bit
Windows~11 Home, build 26200, with NVIDIA driver 581.29.
The GPU accelerates enabled local operations, notably
MemU's exact item-similarity search.

\paragraph{Software environment.}
Methods with incompatible dependencies run in separate environments.
The general evaluation environment uses Python~3.12 and PyTorch~2.9.1
with CUDA~12.8. MemU uses Python~3.13, MemU~1.5.1, PostgreSQL with
pgvector, and PyTorch-based exact similarity search rather than
FAISS-GPU. SimpleMem uses Python~3.12, OpenAI client~2.3.0,
LanceDB~0.25.3, PyArrow~22.0.0, and pylance~0.39.0; its embeddings
are obtained through model APIs. Memora uses Chroma for persistent
storage and BM25 for lexical retrieval.

The Mem0 mutation audit and ADD/NONE intervention use Mem0~1.0.11,
whereas the MEME store used for evidence analysis uses Mem0~2.0.11. MemOS and MemMachine
run as locally hosted Docker services. Software versions and operating
conditions are given where recorded; some historical dependency versions
are unavailable.

\FloatBarrier

\section{Review of Prior RAG Evaluations}
\label{app:prior-rag-audit}

This appendix reviews raw and naive RAG baselines in prior studies.
Weak baseline configurations may partly explain the reported poor performance of RAG. These observations motivate stronger retrieval controls.

\subsection{Review scope}

We examine retrieval-design limitations that may explain weak historical RAG
baselines relative to \rawrag{}: overly large chunks, overly low top-$k$, and
absent reranking. We examine studies that independently evaluated raw or
naive RAG, along with the public RAG configuration accompanying Memobase's
LoCoMo evaluation. For Memobase, we examine only the configuration.
The published reference scores use different readers, prompts, question
sets, budgets, and scoring across papers.

\subsection{Comparative analysis}

Table~\ref{tab:prior-rag-audit} names the configuration limitations directly.
``Too large'' and ``too low'' are our diagnoses of coarse retrieval granularity
and restricted evidence coverage, respectively.
``Not reported'' denotes a reporting gap. Unless specified otherwise,
scores are LLM-judge accuracies on a 0--100 scale.

{\small
\setlength{\tabcolsep}{4pt}
\renewcommand{\arraystretch}{1.08}
\setlength{\LTcapwidth}{\linewidth}
\begin{longtable}{@{}>{\raggedright\arraybackslash}p{0.18\linewidth}
                    >{\raggedright\arraybackslash}p{0.38\linewidth}
                    >{\raggedright\arraybackslash}p{0.40\linewidth}@{}}
\caption{Review of raw-history retrieval-augmented generation (RAG)
baselines in 14 prior evaluations. $k$ is the number of retrieved units;
listed scores are on a 0--100 scale. Retrieval limitations are our
assessments of the reported configurations, not controlled estimates of
their effects. Scores are not directly comparable across studies.
For Memobase, only the public configuration is reviewed.}
\label{tab:prior-rag-audit}\\
\toprule
Work and locator & Reported RAG configuration and reference score & Missing or under-tuned design choice \\
\midrule
\endfirsthead
\multicolumn{3}{l}{\textit{Table~\ref{tab:prior-rag-audit} continued}}\\
\toprule
Work and locator & Reported RAG configuration and reference score & Missing or under-tuned design choice \\
\midrule
\endhead
\midrule
\multicolumn{3}{r}{\textit{Continued on next page}}\\
\endfoot
\bottomrule
\endlastfoot
Mem0 \citep{chhikara2025mem0}\newline Sec.~3.3; Tab.~2
& LoCoMo RAG (256/2) reaches 60.97 with GPT-4o-mini. The paper searches
seven chunk sizes, 128--8192 tokens, but always uses $k\in\{1,2\}$ and
embedding similarity only.
& Top-$k$ is too low ($k\leq 2$): even the small-chunk configurations
retrieve at most two evidence locations across the chunk-size search.\\
\addlinespace
Memobase \citep{memobase2025locomo}\newline LoCoMo evaluation
& The public raw-RAG configuration uses 500-token chunks and $k=1$,
ranked by embedding cosine similarity. No independent RAG score is
reported; the comparison scores for other methods are imported from Mem0. 
& Top-$k$ is too low ($k=1$): only one 500-token chunk reaches the reader,
so evidence distributed across multiple chunks cannot be combined.
\\
\addlinespace
Nemori \citep{nan2025nemori}\newline Sec.~4
& RAG-4096 uses 4096-token chunks; its LoCoMo RAG reference is 30.20
with GPT-4o-mini and 32.90 with GPT-4.1-mini.
& Chunk size is too large (4096 tokens): each retrieval unit bundles
many dialogue details, reducing the specificity of dense retrieval and
spending context on unrelated material.\\
\addlinespace
RecMem \citep{dai2026recmem}\newline App.~E.1, E.5
& Message-preserving $\sim$2048-token chunks; $k=3/5$ on
LoCoMo/LongMemEval-S. Scores: LoCoMo 49.68/54.42; LongMemEval-S
59.40/67.00 (GPT-4o-mini/GPT-4.1-mini).
& Chunk size is too large ($\sim$2048 tokens), and top-$k$ is too low
($k=3/5$). The matched budgets still use coarse retrieval units and restrict
evidence to a few locations.\\
\addlinespace
Memora \citep{xia2026memora}\newline Sec.~5.1
& 500-token chunks, $k=3$; LoCoMo 63.30 with GPT-4.1-mini.
The reported 65.60 LongMemEval result belongs to Full-context and is
excluded here.
& Top-$k$ is too low ($k=3$): only three chunks, totaling at most
1500 source tokens, can supply evidence from the conversation.\\
\addlinespace
xMemory \citep{hu2026xmemory}\newline Secs.~4.1, 5.3
& Top-20 original-message chunks by vector similarity; LoCoMo
27.95 BLEU-1 / 36.48 F1 with Qwen3-8B.
& Cross-encoder reranking is absent from the reported raw baseline.
LLMLingua-2 is evaluated separately as a pruning control.\\
\addlinespace
BEAM / LIGHT \citep{tavakoli2026beam}\newline Sec.~4.1; App.~C.3
& On BEAM, RAG indexes each user--assistant pair and returns the five most
similar pairs; BGE-small and BGE-large are compared.
& Top-$k$ is too low ($k=5$): only five user--assistant pairs can
cover evidence dispersed across a long conversation.\\
\addlinespace
PREMem \citep{kim2025premem}\newline Secs.~4.1, 5.3
& Turn/session retrieval with Stella-400M; context-budget experiments
are included.
& Short-turn retrieval can separate evidence from adjacent context;
whole-session retrieval can bundle unrelated material. These structural
units vary in token length.\\
\addlinespace
LightMem \citep{fang2025lightmem}\newline Tab.~1; App.~D.1--D.2
& NaiveRAG scores 61.00 on LongMemEval-S with GPT-4o-mini.
& Chunk size, final $k$, and candidate/reranking policy are
underspecified; an independently tuned raw baseline is not documented.\\
\addlinespace
LoCoMo \citep{maharana2024locomo}\newline Secs.~5--6; App.~C.1
& DRAGON retrieval with GPT-3.5-Turbo-16k; dialogue units, observations,
summaries, and retrieval counts are compared.
& No separate cross-encoder reranker is reported for the raw-dialogue
baseline.\\
\addlinespace
LongMemEval \citep{wu2025longmemeval}\newline Sec.~5; App.~E
& Studies session/round units, fact-expanded keys, temporal filtering,
budgets, and reading policies.
& Extensive retrieval tuning is already present; the remaining omission
is cross-encoder reranking over a larger candidate pool.\\
\addlinespace
SeCom \citep{pan2025secom}\newline Tab.~1; App.~A.5
& MPNet turn retrieval: 57.99 GPT4Score on author-generated LoCoMo QA;
54.77 turns occupy 3288 tokens on average under a 4k-token budget.
& Turns average only about 60 tokens, so evidence can be separated from
adjacent context. A token-window raw alternative could preserve this local
context within each retrieval unit.\\
\addlinespace
SGMem \citep{wu2025sgmem}\newline Sec.~4.3; Tab.~1
& Tests turn/round/session units at $k=5,10$. Top-10 turn/session RAG:
54.4/57.6 on LongMemEval; 33.0/40.8 on a 500-question LoCoMo sample.
& Turn units can isolate evidence from adjacent context, whereas session
units can bundle unrelated details. Retrieved token volume and granularity
vary across these units at a shared $k$.\\
\addlinespace
MemoryAgent\-Bench \citep{hu2026memoryagentbench}\newline Secs.~4.1, 4.3; App.~E, F
& The benchmark evaluates 512- and 4096-token input units, sweeps
$k\in\{2,5,10\}$, and reports additional budget analyses.
& Chunk-size/top-$k$ tuning is already present; the missing component is
cross-encoder reranking of a broad dense candidate pool.\\
\end{longtable}
}

\subsection{Synthesis}

Several reviewed baselines have concrete configuration bottlenecks:
Nemori and RecMem use overly large chunks, while Mem0, Memobase's public
RAG configuration, RecMem, Memora, and BEAM restrict retrieval to too few
evidence locations. Addressing these concerns requires changing chunk size
and retrieval depth alongside ranking. Other evaluations already explore
granularity and depth more
extensively, making reranking a distinct remaining control.
These diagnoses motivate joint chunk-size and retrieval-depth tuning under
matched budgets. Quantifying their contribution to published score gaps
requires controlled reruns.

\begin{findingbox}{Review Finding}
Several prior RAG baselines use chunks that are too large or top-$k$ that is
too low, limiting retrieval specificity or evidence coverage. Establishing
a need for agentic memory systems requires comparison with \rawrag{} that
addresses these basic design constraints.
\end{findingbox}
\FloatBarrier

\section{Extended Discussion and Limitations}
\label{app:additional-limitations}

\paragraph{Operational memory management.}
The evaluated tasks ask for answers from recorded histories. Operational
memory systems may also need to expose authoritative state to other tools, trigger
actions after updates, or erase stored information. These requirements
introduce memory-management objectives beyond answer accuracy.

\paragraph{Reader dependence.}
The large, task-dependent reader differences in the phase-visible MEME
oracle (Table~\ref{tab:meme-oracle}) show that state reconstruction depends
on reader capability. Cross-system utilization estimates also depend on
which questions each system supplies with sufficient evidence.
Whether state construction during ingestion helps weaker readers or reduces
repeated inference remains an open question. Retrieval should be evaluated
with the intended reader.

\paragraph{Demonstration structure.}
MemoryAgentBench's test-time-learning split combines few-shot
classification with recommendation. For classification, usable
demonstrations require factual content, recognizable
input--label pairings, example boundaries, and presentation format.
Atomic-fact extraction or summarization can disrupt that format even
when much of the underlying content survives. The drop from 47.68\%
to 15.37\% under atomic-fact extraction may reflect reduced evidence
volume at fixed item count, disrupted demonstration structure, and
reader--format mismatch, alongside factual loss. Both token volume and
demonstration format vary in this ablation; the separate six-system 5k
control in Appendix~\ref{app:reader-context-budget} leaves these factors
coupled. The result shows that the tested representation and reader
combination performs poorly on TTL.

\paragraph{Scalability and generalizability.}
Raw storage grows with history length, and a fixed candidate and
reader budget can miss dispersed or low-similarity evidence. Turn
expansion and overlapping windows may also spend context on repeated
text. Retrieval quality across history lengths and the cost of repeated
query-time reasoning remain to be measured alongside ingestion.
Hosted model and service behavior can change, and outputs can vary
across temperature-zero requests and repeated judgments.
Longer interactive workloads, bounded-storage settings, and broader
multimodal evidence may favor different memory-management trade-offs.

\FloatBarrier

\section{Prompt Templates}
\label{app:prompts}

We reproduce memory-management, reranker, reader, and judge templates, with long lines
wrapped for typesetting. In the LoCoMo judge prompt, a
non-grading aside is omitted; all grading criteria and output
requirements are unchanged.
Braced fields denote per-example substitutions. In reader prompts,
\texttt{\{context\}} is the retrieved
evidence supplied to the reader, and \texttt{\{question\}} is the question
after any benchmark-specific formatting described below. Gold answers
and hidden-gold evidence requirements are judge-side inputs. In the
complete-store control, gold fields also guide
evidence selection, while the reader-visible policy supplies task metadata
and transition semantics but no gold answer or entity values.

\subsection{Qwen3 reranker}
\label{app:prompt-qwen-reranker}

For Qwen3-Reranker-4B, we use the chat template from the official model
card accompanying \citet{zhang2025qwen3embedding}, with the
question--passage relevance instruction shown below.
Here \texttt{\{retrieval\_query\}} is the question
defined in Appendix~\ref{app:query-construction}, and
\texttt{\{passage\}} is one candidate retrieval window, before
display-time expansion to complete turns.

\begin{promptbox}{Qwen3 reranker: query--passage relevance template}
<|im_start|>system
Judge whether the Document meets the requirements based on the Query and the Instruct provided. Note that the answer can only be "yes" or "no".<|im_end|>
<|im_start|>user
<Instruct>: Given a question and a passage, determine if the passage contains information relevant to answering the question.
<Query>: {retrieval_query}
<Document>: {passage}<|im_end|>
<|im_start|>assistant
<think>

</think>

\end{promptbox}

\subsection{Memory-management variants}
\label{app:prompt-agentic-mutation-ablations}

The three memory-management variants share the following system message:

\begin{promptbox}{Ingestion-time memory management: system message}
You are a careful memory extraction engine. Return one valid JSON object and no prose.
\end{promptbox}

The following user prompts are instantiated once per raw 256-token window.
The fields \texttt{\{conversation\_id\}}, \texttt{\{source\_chunk\_id\}},
\texttt{\{source\_chunk\_index\}}, and \texttt{\{source\_memory\_chunk\}}
are filled from the current conversation and window. The numeric limits
in the atomic-fact prompt are the evaluated limits.

\begin{promptbox}{Atomic-fact extraction}
Extract explicit atomic facts from the source memory chunk.

Rules:
- Use only facts directly supported by the text. Do not infer hidden causes.
- Preserve names, titles, locations, dates, event order, speaker labels, and serial numbers exactly.
- For conflict-resolution facts, keep the original serial number. Larger serial numbers mean newer facts; do not drop this ordering clue.
- Return at most 6 high-salience facts for retrieval and question answering.
- Prefer concrete events, state changes, decisions, locations, dates, names, and conflict/newness cues.
- Omit generic commentary, style descriptions, duplicate details, and low-value micro-actions when the chunk has many candidates.
- Keep each fact under 28 words.
- Keep each source quote under 12 words.
- Keep entities to at most 5 names.
- Ignore wrapper instructions such as "I have memorized"; extract from the actual remembered content.
- Return JSON only.

Required JSON schema:
{
  "facts": [
    {
      "fact": "one concise standalone fact",
      "speaker": "speaker if present, else empty string",
      "time": "time/date if present, else empty string",
      "order": 0,
      "serial_number": "exact serial/newness marker if present, else empty string",
      "entities": ["up to 5 entity names"],
      "source_quote": "short exact quote supporting the fact"
    }
  ]
}

Conversation ID: {conversation_id}
Source chunk id: {source_chunk_id}
Source chunk index: {source_chunk_index}

SOURCE MEMORY CHUNK:
{source_memory_chunk}
\end{promptbox}

\begin{promptbox}{Raw-summary extraction}
Summarize this source memory chunk for retrieval and question answering.

Rules:
- Use only source-supported information.
- Preserve event order, entities, dates/times, speaker names, and serial numbers.
- For conflict-resolution facts, explicitly mention serial numbers and newer/older ordering.
- Do not replace the raw text; produce a compact structured companion summary.
- Return JSON only.

Required JSON schema:
{
  "summary": "dense factual summary",
  "key_events": ["ordered event or state update"],
  "entities": ["important entities"],
  "time_order": ["date/time/order/serial cues"],
  "conflicts_or_updates": ["conflicts, revisions, newest facts, if any"]
}

Conversation ID: {conversation_id}
Source chunk id: {source_chunk_id}
Source chunk index: {source_chunk_index}

SOURCE MEMORY CHUNK:
{source_memory_chunk}
\end{promptbox}

\begin{promptbox}{Graph-link judgment}
Judge which existing chunks are genuinely related to the new chunk and should
be connected by an undirected graph edge.

Link chunks only when their connection would be useful for multi-hop retrieval
or joint question answering: for example, they describe the same entity,
event, relationship, evolving state, cause/effect chain, or a concrete follow-up.
Topical or stylistic similarity alone is not enough. Use only the supplied text.
It is valid to return no related chunks. Return JSON only.

Required JSON schema:
{
  "related_chunks": [
    {"chunk_index": 0, "reason": "brief source-supported connection"}
  ]
}

Conversation ID: {conversation_id}
New chunk index: {new_chunk_index}

NEW CHUNK:
{new_chunk}

CANDIDATE EXISTING CHUNKS (embedding Top-10):
{candidate_existing_chunks}
\end{promptbox}

\subsection{LoCoMo LLM judge}
\label{app:prompt-locomo-judge}

The evaluator sends the following system and user messages. The fields
\texttt{\{golden\_answer\}} and \texttt{\{generated\_answer\}} are
replaced by the reference answer and the evaluated model answer.
The reasoning request and final JSON-label instruction are both reproduced
from the Mem0 and EverMemOS grading prompts
\citep{chhikara2025mem0,hu2026evermemos}. This preserves the evaluated
wording; response parsing and repetition counts are specified below.

\begin{promptbox}{LoCoMo judge: system message}
You are an expert grader that determines if answers to questions match a gold standard answer
\end{promptbox}

\begin{promptbox}{LoCoMo judge: user message}
Your task is to label an answer to a question as 'CORRECT' or 'WRONG'. You will be given the following data:
    (1) a question (posed by one user to another user),
    (2) a 'gold' (ground truth) answer,
    (3) a generated answer
which you will score as CORRECT/WRONG.

The point of the question is to ask about something one user should know about the other user based on their prior conversations.
The gold answer will usually be a concise and short answer that includes the referenced topic, for example:
Question: Do you remember what I got the last time I went to Hawaii?
Gold answer: A shell necklace
The generated answer might be much longer, but you should be generous with your grading - as long as it touches on the same topic as the gold answer, it should be counted as CORRECT.

For time related questions, the gold answer will be a specific date, month, year, etc. The generated answer might be much longer or use relative time references (like "last Tuesday" or "next month"), but you should be generous with your grading - as long as it refers to the same date or time period as the gold answer, it should be counted as CORRECT. Even if the format differs (e.g., "May 7th" vs "7 May"), consider it CORRECT if it's the same date.

Now it's time for the real question:
Question: {question}
Gold answer: {golden_answer}
Generated answer: {generated_answer}

First, provide a short (one sentence) explanation of your reasoning, then finish with CORRECT or WRONG.
Do NOT include both CORRECT and WRONG in your response.

Just return the label CORRECT or WRONG in a json format with the key as "label".
\end{promptbox}

\paragraph{Response parsing and scoring.}
Our judge requests use temperature zero without a JSON-only response
constraint. The parser first selects a brace-delimited object in the
first matching Markdown code block (with an optional \texttt{json}
language tag). If none is found, it selects the first non-nested
brace-delimited object containing a quoted \texttt{label} key and
string value, including when that object is embedded in explanatory
text. Otherwise, it attempts to parse the entire whitespace-trimmed
response as JSON. The selected text must decode to an object; the
\texttt{label} value is stripped of surrounding whitespace and
converted to uppercase, and only an exact \texttt{CORRECT} scores 1.
Explanatory text around a valid JSON label is accepted; bare
\texttt{CORRECT} or \texttt{WRONG} text is rejected.

A \texttt{WRONG} label scores 0. Empty responses, malformed JSON,
missing or unusable labels, and other labels also score 0, as do
request exceptions that reach the evaluator. All outcomes remain in
the denominator with their assigned scores.
There is no format-repair retry at the evaluator level. Three
independent judgments are requested per fixed answer, and the reported
accuracy is their question-weighted mean,
$100(3N)^{-1}\sum_{i=1}^{N}\sum_{r=1}^{3} y_{ir}$, with
$N=1{,}540$ and $y_{ir}\in\{0,1\}$.

\paragraph{Failure attribution.}
The historical evaluation records contain binary scores. Raw judge
responses and per-judgment parse and request status are unavailable.
A zero may represent a valid \texttt{WRONG} judgment, a formatting
failure, or a failed request, leaving the exact format-failure count
unknown. For the four \rawrag{} conditions in
Tables~\ref{tab:locomo-openai-protocol}
and~\ref{tab:locomo-qwen-protocol}, we verify all 4,620 expected binary
judgments per condition (18,480 in total), with zero missing judgment
slots. Their zero-score counts are 505 and 291 for the
\texttt{text-embedding-3-small} conditions, and 502 and 252 for the
Qwen3 conditions, in GPT-4o-mini/GPT-4.1-mini reader order.
For each condition, the number of format failures is bounded between
zero and its zero-score count. Scores are reported unchanged; recovering
the historical failure rate would require the original responses and
diagnostic status.

\subsection{LoCoMo reader}
\label{app:prompt-locomo-reader}

The standard \rawrag{} reader receives the following structured
reasoning prompt as one user message, without a separate system message.
Evaluation uses the text following \texttt{FINAL ANSWER:}, or the full
response if that marker is absent. The minimal-answer variant is reported
separately in Appendix~\ref{app:results}.

\begin{promptbox}{LoCoMo reader: user message (also shared by LongMemEval)}
You are an intelligent memory assistant tasked with retrieving accurate information from episodic memories.

# CONTEXT:
You have access to episodic memories from conversations between two speakers. These memories contain
timestamped information that may be relevant to answering the question.

# INSTRUCTIONS:
Your goal is to synthesize information from all relevant memories to provide a comprehensive and accurate answer.
You MUST follow a structured Chain-of-Thought process to ensure no details are missed.
Actively look for connections between people, places, and events to build a complete picture. Synthesize information from different memories to answer the user's question.
It is CRITICAL that you move beyond simple fact extraction and perform logical inference. When the evidence strongly suggests a connection, you must state that connection. Do not dismiss reasonable inferences as "speculation." Your task is to provide the most complete answer supported by the available evidence.

# CRITICAL REQUIREMENTS:
1. NEVER omit specific names - use "Amy's colleague Rob" not "a colleague"
2. ALWAYS include exact numbers, amounts, prices, percentages, dates, times
3. PRESERVE frequencies exactly - "every Tuesday and Thursday" not "twice a week"
4. MAINTAIN all proper nouns and entities as they appear

# RESPONSE FORMAT (You MUST follow this structure):

## STEP 1: RELEVANT MEMORIES EXTRACTION
[List each memory that relates to the question, with its timestamp]
- Memory 1: [timestamp] - [content]
- Memory 2: [timestamp] - [content]
...

## STEP 2: KEY INFORMATION IDENTIFICATION
[Extract ALL specific details from the memories]
- Names mentioned: [list all person names, place names, company names]
- Numbers/Quantities: [list all amounts, prices, percentages]
- Dates/Times: [list all temporal information]
- Frequencies: [list any recurring patterns]
- Other entities: [list brands, products, etc.]

## STEP 3: CROSS-MEMORY LINKING
[Identify entities that appear in multiple memories and link related information. Make reasonable inferences when entities are strongly connected.]
- Shared entities: [list people, places, events mentioned across different memories]
- Connections found: [e.g., "Memory 1 mentions A moved from hometown → Memory 2 mentions A's hometown is LA → Therefore A moved from LA"]
- Inferred facts: [list any facts that require combining information from multiple memories]

## STEP 4: TIME REFERENCE CALCULATION
[If applicable, convert relative time references]
- Original reference: [e.g., "last year" from May 2022]
- Calculated actual time: [e.g., "2021"]

## STEP 5: CONTRADICTION CHECK
[If multiple memories contain different information]
- Conflicting information: [describe]
- Resolution: [explain which is most recent/reliable]

## STEP 6: DETAIL VERIFICATION CHECKLIST
- [ ] All person names included: [list them]
- [ ] All locations included: [list them]
- [ ] All numbers exact: [list them]
- [ ] All frequencies specific: [list them]
- [ ] All dates/times precise: [list them]
- [ ] All proper nouns preserved: [list them]

## STEP 7: ANSWER FORMULATION
[Explain how you're combining the information to answer the question]

## FINAL ANSWER:
[Provide the concise answer with ALL specific details preserved]

---

{context}

Question: {question}

Now, follow the Chain-of-Thought process above to answer the question:
\end{promptbox}

\paragraph{Minimal-answer CoT variant.}
The minimal-answer condition in Appendix~\ref{app:results} keeps the
seven-step CoT structure, context and question placeholders, single-user-message
format, and final-answer extraction described above. It shortens the final
answer by replacing the instruction under \texttt{FINAL ANSWER:} with:

\begin{promptbox}{LoCoMo minimal-answer CoT: replacement final-answer instruction}
## FINAL ANSWER:
[Output only the shortest answer span that directly answers the question. Prefer exactly the entity, date, number, or list requested. Do not explain, restate the question, or add background, evidence, introductory phrases, or concluding remarks. For list questions, output only a comma-separated list. Preserve only the specific details required for correctness.]
\end{promptbox}

The template also makes three minor wording or notation
changes: in Step~2, \emph{list any recurring patterns} becomes
\emph{list all recurring patterns}; in Step~3, \emph{mentioned across
different memories} becomes \emph{mentioned in multiple memories}, and
the arrows in the \emph{Connections found} example are written as
\texttt{->}. All other prompt text is unchanged.

\subsection{LongMemEval reader}
\label{app:prompt-longmemeval-reader}

The standalone LongMemEval benchmark uses the same
structured reasoning prompt reproduced in
Appendix~\ref{app:prompt-locomo-reader}; there is no separate system
message. For examples with an annotated question date, the
reader receives the following date-aware query in place of
\texttt{\{question\}} in that shared template.
\texttt{\{raw\_question\}} below denotes the original benchmark question.
If no question date is present, the original question is used directly.
This standalone configuration differs from MemoryAgentBench's
LongMemEval source (prompt in Appendix~\ref{app:prompt-mab-reader}).

\begin{promptbox}{LongMemEval: date-aware replacement for \texttt{\{question\}}}
Question date: {question_date}
Treat this as the current time when resolving relative phrases such as "recently", "last month", "previous", "current", or "now". Only use memories dated at or before this question date.
Question: {raw_question}
\end{promptbox}

\subsection{MEME reader}
\label{app:prompt-meme-reader}

The main MEME evaluation uses the following official short-answer
prompt, sent as one user message without
a separate system message. The context contains the evidence available
at the corresponding evaluation phase. This is the standard answer
prompt; the value-free dependency-validity control for Abs is a
separate oracle condition (Appendix~\ref{app:prompt-abs-validity}).
The complete-store control instead appends the task policy in
Appendix~\ref{app:prompt-complete-store} to its selected context.

\begin{promptbox}{MEME reader: user message}
Answer the user's question based ONLY on the context provided below.

If the information is not in the context, say you don't have that information.

Answer with ONLY the value. Do not explain or add context.

Context:

{context}

Question: {question}
\end{promptbox}

\subsection{MemoryAgentBench reader}
\label{app:prompt-mab-reader}

The MemoryAgentBench reader receives the following system and user
messages.
\texttt{\{task\_query\}} is one of the eight RAG query templates
reproduced below, selected by source. The appended
\texttt{\{current\_time\}} is the local date and time at evaluation,
formatted as \texttt{YYYY-MM-DD HH:MM:SS}.

\begin{promptbox}{MemoryAgentBench reader: system message}
You are a helpful AI. Answer the question based on query and memories.
{context}
\end{promptbox}

\begin{promptbox}{MemoryAgentBench reader: user message}
{task_query}

Current Time: {current_time}
\end{promptbox}

\paragraph{Accurate Retrieval.}
The source selects the LongMemEval, EventQA, or RULER QA query template.
In all task templates below, \texttt{\{question\}} denotes
the original source question.

\begin{promptbox}{MemoryAgentBench: LongMemEval query}
The history chats are between you and a user. Based on the relevant chat history, answer the question as concisely as you can, using a single phrase if possible.

 {question} 

 Answer:
\end{promptbox}

\begin{promptbox}{MemoryAgentBench: EventQA query}
Based on the context you memorized, complete the task below:

{question}

 The event that happens next is:
\end{promptbox}

\begin{promptbox}{MemoryAgentBench: RULER QA query}
Answer the question based on the memorized documents. Only give me the answer and do not output any other words. 

 Now Answer the Question: {question}
\end{promptbox}

\paragraph{Test-Time Learning.}
The five ICL sources share the numerical-label template.
The string \texttt{\{label\}} in that template is sent literally.
ReDial uses the recommendation template.

\begin{promptbox}{MemoryAgentBench: in-context learning query}
Use the provided mapping from the context to numerical label to assign a numerical label to the context. Only output "label: {label}" and nothing else. 

Question:{question} 

 label:
\end{promptbox}

\begin{promptbox}{MemoryAgentBench: ReDial query}
Pretend you are a movie recommender system. You need to recommend movies based on the dialogues you have memorized. Now I will give you a new conversation between a user and you (a recommender system). Based on the conversation, you reply me with 20 recommendations without extra sentences. 

For Example:

[Conversation]

The recommendations are: 
1.movie1
2.movie2
...

 Here is the conversation: {question} 

 The recommendations are: 
\end{promptbox}

\paragraph{Long-Range Understanding.}
DetectiveQA and InfBench use the respective question-answering and
summarization templates below.

\begin{promptbox}{MemoryAgentBench: DetectiveQA query}
Based on the context you memorized, answer the question below. You are required to answer the question based on the strict output format.

 {question} 

\end{promptbox}

\begin{promptbox}{MemoryAgentBench: InfBench summarization query}
You are given a book above and you are tasked to summarize it. 

{question} 

 Now summarize the book.
\end{promptbox}

\paragraph{Conflict Resolution.}
The single-hop and multi-hop fact-consolidation sources share the
following template. Its illustrative answer is part of the original
synthetic-knowledge instruction.

\begin{promptbox}{MemoryAgentBench: fact-consolidation query}
Pretend you are a knowledge management system. Each fact in the knowledge pool is provided with a serial number at the beginning, and the newer fact has larger serial number. 
 You need to solve the conflicts of facts in the knowledge pool by finding the newest fact with larger serial number. You need to answer a question based on this rule. You should give a very concise answer without saying other words for the question **only** from the knowledge pool you have memorized rather than the real facts in real world. 

For example:

 [Knowledge Pool] 

 Question: Based on the provided Knowledge Pool, what is the name of the current president of Russia? 
Answer: Donald Trump 

 Now Answer the Question: Based on the provided Knowledge Pool, {question} 
Answer:
\end{promptbox}

\subsection{PersonaMem-v2 reader}
\label{app:prompt-personamem-reader}

For the multiple-choice results in Appendix~\ref{app:personamem-v2},
the reader receives the evidence, original query, personalization reminder,
response options,
and selection instruction in the single user message shown below;
no system message is added.
\texttt{\{raw\_question\}} denotes the original user request, before
the personalization reminder is appended; only this original request
is used for retrieval in the reported run.
\texttt{\{options\_text\}} contains all options in dataset order, one
per line in the form \texttt{A. option text}.

\begin{promptbox}{PersonaMem-v2 reader: fully assembled user message}
{context}

{raw_question} Please recall my related preferences from our conversation history to give personalized responses.

Please choose the best answer from the following options:

{options_text}

Think step by step about which answer best fits the user's query and conversation context. Provide your reasoning first, then give your final answer as 'Final Answer: [Letter]'
\end{promptbox}

\subsection{Semantic sufficiency judge}
\label{app:prompt-semantic-judge}

The MemoryAgentBench coverage/utilization analysis uses the following
system and user messages. The same template supplies the
original MEME semantic-answerability assessment. The full evaluations use
DeepSeek-V4-Flash, temperature zero, and JSON-object response
format. Neither system identity nor the evaluated reader answer is
included in these messages. The boolean \texttt{answerable} is the
sufficiency label; \texttt{support} and \texttt{notes} provide diagnostic
information.

\begin{promptbox}{Semantic sufficiency judge: system message}
You are a strict evidence answerability judge.
\end{promptbox}

\begin{promptbox}{Semantic sufficiency judge: user message}
You are judging retrieval answerability.

Return JSON only with:
{"answerable": true/false, "support": "sufficient|partial|none|contradictory", "notes": "short reason"}

Mark answerable=true only if the retrieved context contains enough information to derive the gold answer without relying on world knowledge or the model's prior memory. Exact wording is not required.

Question:
{question}

Gold answer:
{golden_answer}

Retrieved context excerpt:
{excerpt}
\end{promptbox}

\paragraph{Evidence selection.}
The judge receives a selected excerpt of the reader context. Selection
uses the ordered list of reference-answer values, ignores
values shorter than three characters, and finds the first
case-insensitive literal match of
each remaining value in the formatted context. Each match contributes
up to 900 characters before and after the matched span. Snippets are
concatenated in value-list order with \texttt{[...]} separators, without
deduplication, and clipped to the 9,000-character limit. If no value
matches, the first 9,000 characters of context are used. The resulting
judgment measures the sufficiency of this excerpt for the reference answer.

\subsection{Result and path assessment}
\label{app:prompt-result-path}

The task-aware assessment judges result and path coverage jointly.
It reassesses Tr in both phases and Del, Cas, and Abs after the update;
other task/phase rows keep the original semantic label. The
judge uses DeepSeek-V4-Flash, temperature zero, JSON-object
response format, and complete formatted contexts without a character
limit or the gold-localized excerpt above.

\begin{promptbox}{MEME result/path judge: system message}
You audit complete reasoning-chain evidence and return JSON only.
\end{promptbox}

\begin{promptbox}{MEME result/path judge: user message}
You are a strict MEME retrieval-evidence auditor.

Judge whether the RETRIEVED CONTEXT satisfies the task-specific evidence contract. The hidden-gold
contract tells you what to look for, but it is NOT evidence. Only statements in RETRIEVED CONTEXT
count. Do not use the question, contract, world knowledge, or assumptions to fill a missing link.
For cascade/invalidation, isolated mentions are insufficient: the dependency rule, trigger state,
and target state must be related. For deletion, absence of an old value is never proof of deletion.

Return one JSON object with exactly these keys:
{"sufficient": true/false, "value_complete": true/false,
 "temporal_chain": true/false, "deletion_targeted": true/false,
 "dependency_rule": true/false, "upstream_trigger_state": true/false,
 "linked": true/false, "notes": "brief evidence-based reason"}
Use false for component fields that do not apply.

QUESTION:
{question_text}

HIDDEN-GOLD EVIDENCE CONTRACT (requirements only; not evidence):
{contract_json}

RETRIEVED CONTEXT:
{context}
\end{promptbox}

\paragraph{Evidence requirements.}
The placeholder \texttt{\{contract\_json\}} is a JSON object serialized
with two-space indentation. Its fields are shown below; braced
placeholders are replaced by JSON strings or arrays.
Targets and the upstream entity come from the matching annotated MEME
task. History values come from the question's annotated entity values;
old/new target values and upstream before/after values come from episode
entity metadata. Value lists are flattened and deduplicated in order.
These evidence criteria are supplied only to the judge.

\begin{promptbox}{MEME result/path judge: evidence-requirement fields}
{
  "task_type": "{task_type}",
  "phase": "{phase}",
  "target_entities": {target_entities_json_array},
  "history_values": {history_values_json_array},
  "old_target_values": {old_target_values_json_array},
  "new_target_values": {new_target_values_json_array},
  "upstream_entity": "{upstream_entity}",
  "upstream_before_values": {upstream_before_values_json_array},
  "upstream_after_values": {upstream_after_values_json_array},
  "sufficiency_rule": "{task_specific_rule}"
}
\end{promptbox}

The task type selects exactly one of these four strings for
\texttt{sufficiency\_rule}:

\begin{promptbox}{MEME result/path criteria: Tr rule}
Every history value must appear, and the context must establish their chronological order using dates or explicit temporal language such as then/later/after.
\end{promptbox}

\begin{promptbox}{MEME result/path criteria: Del rule}
An explicit deletion instruction/event must point to the target entity or old value. Merely failing to retrieve the old value is not deletion evidence.
\end{promptbox}

\begin{promptbox}{MEME result/path criteria: Cas rule}
The new target value, the dependency rule, and the upstream post-trigger state must all appear and be linked into one executable causal chain.
\end{promptbox}

\begin{promptbox}{MEME result/path criteria: Abs rule}
The old target value, its dependency rule, and the upstream state that invalidates it must all appear and be linked into one executable causal chain.
\end{promptbox}

\paragraph{Coverage labels.}
For the Tr-after and Cas-after comparison, \texttt{value\_complete} is
result coverage and \texttt{sufficient} is path coverage. The analysis checks that
no path-positive row is result-negative before computing
$P(\mathrm{path}\mid\mathrm{result})$. It uses the judge's sufficiency
label directly. Tr requires the complete history plus ordering; Cas requires the
new target plus a linked dependency and trigger. Del and Abs are excluded
from this nested result/path comparison because a deleted or invalidated
old value is invalid as a current result. They remain part of the broader
task-aware assessment. Both labels in the nested comparison come from the
joint full-context judgment.

\subsection{Dependency-validity control}
\label{app:prompt-abs-validity}

The source-oracle Abs control adds the dependency-validity instruction
reproduced below.

\begin{promptbox}{Abs source-oracle control: prefix to reader context}
<DEPENDENCY_VALIDITY_RULE>
For a current-answer question, a fact explicitly stated to depend on another
fact becomes stale if the context later says that dependency changed. Do not
report that earlier dependent value as the current answer. If no explicit new
value for the dependent fact is stated after the change, answer exactly: I
don't have that information.
</DEPENDENCY_VALIDITY_RULE>
\end{promptbox}

In Table~\ref{tab:meme-oracle}, Abs uses this condition; Overall therefore combines those
task-specific conditions. ``Value-free'' means that the added instruction
does not disclose the answer value.

\subsection{Complete-store task policy}
\label{app:prompt-complete-store}

In the complete-store diagnostic, the reader receives
the selected evidence followed by a newline and the policy block below,
inside \texttt{\{context\}} of the standard MEME reader prompt.
The phase, task type, and comma-separated target-entity identifiers are
filled from the question specification. Exactly one branch instruction
replaces \texttt{\{task\_rule\}}. The policy's token count is reserved
before evidence packing within the 5,000-token context budget.

\begin{promptbox}{Complete-store readout: policy appended to selected evidence}
<MEME_READOUT_POLICY>
Question phase: {phase}.
Task type: {task_type}.
Target entities only: {target_entities}.
Treat the evidence as a chronological state log. Later applicable state updates override earlier values; do not combine superseded and current values.
{task_rule}
Use only the evidence below. Do not infer a value that the evidence and the stated transition rule do not support.
</MEME_READOUT_POLICY>
\end{promptbox}

The wording ``evidence below'' is reproduced exactly as presented
to the reader, although the policy follows the evidence. The branch rules
are selected in the following order, with the first matching condition
used:

\begin{promptbox}{Complete-store policy: before-phase Cas or Abs}
Before-phase rule: the upstream change has not happened yet. Any conditional replacement is hypothetical and must not be activated; return the current pre-change target value.
\end{promptbox}

\begin{promptbox}{Complete-store policy: after-phase Cas}
Cascade rule: find the stated dependency between the target and its upstream entity, then apply it when the later upstream change occurs. Return only the resulting current target value.
\end{promptbox}

\begin{promptbox}{Complete-store policy: after-phase Abs}
Absence rule: the upstream entity changed but no replacement target value was declared. State that the current target is uncertain; a previous value may be identified only as previous, never as current.
\end{promptbox}

\begin{promptbox}{Complete-store policy: after-phase Del}
Deletion rule: a request to remove the target makes it unavailable as current memory. Say that it was explicitly removed and do not reveal the deleted value.
\end{promptbox}

\begin{promptbox}{Complete-store policy: Tr (either phase)}
Tracking rule: list every target value in chronological order, earliest to latest, without adding unrelated items.
\end{promptbox}

\begin{promptbox}{Complete-store policy: Agg (either phase)}
Aggregation rule: return one value for every listed target entity and exclude facts belonging to other entities.
\end{promptbox}

\begin{promptbox}{Complete-store policy: ER (either phase)}
Exact-recall rule: reproduce the target value verbatim, preserving its wording and order.
\end{promptbox}

\begin{promptbox}{Complete-store policy: fallback (including before-phase Del)}
Return the latest phase-visible value for the target entity.
\end{promptbox}

The raw-evidence control uses the same policy. Its information scope
and distinction from the source oracle are defined in
Appendix~\ref{app:complete-store}.

\end{document}